%% file: paper.tex
\documentclass[]{memtensor}

\usepackage[toc,page,header]{appendix}
\usepackage[utf8]{inputenc} % allow utf-8 input
\usepackage[T1]{fontenc}    % use 8-bit T1 fonts
\usepackage{hyperref}       % hyperlinks
\usepackage{url}            % simple URL typesetting
\usepackage{booktabs}       % professional-quality tables
\usepackage{amsfonts}       % blackboard math symbols
\usepackage{nicefrac}       % compact symbols for 1/2, etc.
\usepackage{microtype}      % microtypography
\usepackage{amsmath} 
\usepackage{etoolbox}
\usepackage{minitoc}
\usepackage{tablefootnote}  % footnote in table
\usepackage{threeparttable}
\usepackage{listings}
\usepackage{tabularx}
\usepackage{algorithm}
\usepackage{algorithmic}
\usepackage{tcolorbox}
\usepackage{array}
\usepackage{lipsum} % 用于生成示例文本，可以根据需要去掉
\usepackage{arydshln}
\usepackage{xspace}
\usepackage{float}
\usepackage{wasysym}
\newcommand{\method}{PERMA\xspace}
\usepackage{graphicx}
\usepackage{makecell}
\usepackage{pifont}
\newcommand{\cmark}{\textcolor{green!70!black}{\ding{51}}} % 绿色打钩
\newcommand{\xmark}{\textcolor{red!70!black}{\ding{55}}}   % 红色打叉
\newcommand{\pmark}{\textcolor{orange!80!black}{\LEFTcircle}} % 橙色/黄色符号（表示部分支持或有区别）
\usepackage{wrapfig}

\usepackage{amsthm} % 引入定理与定义宏包

\theoremstyle{definition}
\newtheorem{definition}{Definition}

\tcbuselibrary{skins} % prompt中的包

\renewcommand{\arraystretch}{1.5}
\usepackage{enumitem}
\usepackage{ulem}
\definecolor{lightgray}{gray}{0.9}
\usepackage{indentfirst}

\title{\method: Benchmarking \underline{Per}sonalized \underline{M}emory \underline{A}gents via Event-Driven Preference and Realistic Task Environments}

\author[1]{Shuochen Liu}
\author[2]{Junyi Zhu}
\author[1]{Long Shu}
\author[1]{Junda Lin}
\author[1]{Yuhao Chen}
\author[1]{Haotian Zhang}
\author[1,3]{Chao Zhang}
\author[1,3]{Derong Xu}
\author[4]{Jia Li}
\author[1,5\dagger]{Bo Tang}
\author[5]{Zhiyu Li}
\author[5]{Feiyu Xiong}
\author[1]{Enhong Chen}
\author[1\dagger]{Tong Xu}

\affiliation[1]{University of Science and Technology of China}
\affiliation[2]{ESAT-PSI, KU Leuven}
\affiliation[3]{City University of Hong Kong}
\affiliation[4]{Northeastern University}
\affiliation[5]{MemTensor (Shanghai) Technology Co., Ltd.}

\abstract{
Empowering large language models with long-term memory is crucial for building agents that adapt to users' evolving needs. Existing evaluations of this capability typically interleave preference-related dialogues with irrelevant conversations, reducing the task to needle-in-a-haystack retrieval while ignoring relationships between events driving user preference evolution. Such settings overlook a fundamental characteristic of real-world personalization: preferences emerge gradually and accumulate across interactions within noisy contexts. To bridge this gap, we introduce \textbf{\method}, a benchmark designed to evaluate persona consistency over time beyond static preference recall. Additionally, we incorporate \textit{(1) text variability} and \textit{(2) linguistic alignment} to simulate erratic user inputs and individual idiolects in real-world data. \method consists of temporally ordered interaction events spanning multiple sessions and domains, with preference-related queries inserted over time. We design both multiple-choice and interactive tasks to probe the model's understanding of persona along the interaction timeline. Experiments demonstrate that by linking related interactions, advanced memory systems extract precise preferences and reduce token consumption, outperforming traditional semantic retrieval of raw dialogues. Nevertheless, they still struggle to maintain a coherent persona across temporal depth and cross-domain interference, highlighting the need for more robust personalized memory management in agents.
}
\correspondence{tangboneu@foxmail.com, tongxu@ustc.edu.cn}
\checkdata[Author Legend]{\dag{}Corresponding author}
\checkdata[Code]{\url{https://github.com/PolarisLiu1/PERMA}}

\begin{document}
\maketitle

\section{Introduction}

\input{content/00_introduction}

\section{Related Work}
\input{content/01_relatedwork}

\section{Preliminaries}
\input{content/02_preliminary}

\section{Methodology}
\input{content/03_method}

\section{Experiments}
\input{content/04_experiments}

\section{Conclusion}
\input{content/05_conclusion}

\clearpage

\bibliographystyle{plainnat}
\bibliography{main}

\clearpage

\beginappendix
\input{content/06_appendix}
\end{document}

%% file: content/00_introduction.tex
\label{intro}
% The rapid advancement of large language models (LLMs) has catalyzed a transformative shift, evolving them from mere knowledge repositories into proactive agents capable of iterative interaction and continuous learning~\cite{zheng2025lifelonglearninglargelanguage}. Within this paradigm, the integration of robust memory architectures has become indispensable for enhancing performance in long-document comprehension and extended dialogue management. A reliable, memory-equipped agent is now regarded as a fundamental prerequisite for developing models that can dynamically adapt to the evolving needs of users~\cite{hu2025memoryageaiagents}. 

The rapid advancement of large language models (LLMs)~\cite{openai2024openaio1card, grattafiori2024llama3herdmodels, qwen2025qwen25technicalreport} has catalyzed a shift from static knowledge access paradigms to interactive agents capable of sustained engagement~\cite{zheng2025lifelonglearninglargelanguage}. In this new paradigm, long-term memory is no longer merely a tool for knowledge retrieval, but a prerequisite for maintaining coherent user representations across interactions. A reliable memory-equipped agent not only stores past information, but continuously constructs and updates a consistent persona state that reflects users’ constraints and behavioral tendencies over time~\cite{hu2025memoryageaiagents}.

Despite these requirements, inherent limitations in context window size and token costs present significant bottlenecks~\cite{tavakoli2026milliontokensbenchmarkingenhancing, shi2026lookreasonforwardrevisitable, zhang2025tearagtokenefficientagenticretrievalaugmented}. To mitigate these challenges, recent progress in retrieval-augmented generation (RAG)~\cite{gao2024retrievalaugmentedgenerationlargelanguage, lyu2024crudragcomprehensivechinesebenchmark, yue2025inferencescalinglongcontextretrieval, luo2025imagescopeunifyinglanguageguidedimage} has bolstered LLMs' capacity to extract relevant factual knowledge~\cite{zhu2024fastmemfastmemorizationprompt, chen-etal-2025-think, xu2025harnessinglargelanguagemodels, liu2025lookthinkunifyingreasoning}. To further refine retrieval quality, several approaches have introduced structured memory indices~\cite{xu2025singlemultigranularitylongtermmemory, xu2024multi, xu2026aligngraganchorrationaleguided}, such as tree-~\cite{rezazadeh2025isolatedconversationshierarchicalschemas, raptor} and graph-based~\cite{chhikara2025mem0buildingproductionreadyai, wang2025mirixmultiagentmemoryllmbased, ma2025thinkongraph20deepfaithful} representations. Complementing these retrieval techniques, reinforcement learning (RL) has been incorporated to optimize memory utilization in long-context settings~\cite{yu2025memagentreshapinglongcontextllm, yan2026memoryr1enhancinglargelanguage, mao2026collaborativemultiagentoptimizationpersonalized}. Beyond generic storage mechanisms, personalization-oriented agents leverage modular memory architectures and context-aware memory structures. These agents emphasize capturing user profiles and preferences to inform downstream conditioning for long-horizon planning and lightweight deployment~\cite{li2025memosmemoryosai, fang2025lightmemlightweightefficientmemoryaugmented}. While these architectural innovations endow agents with the potential for long-range memory, the method for quantifying their personalized performance within complex real-world scenarios remains an open challenge.

\begin{figure*}[!t]
    \centering
    \includegraphics[width=\textwidth]{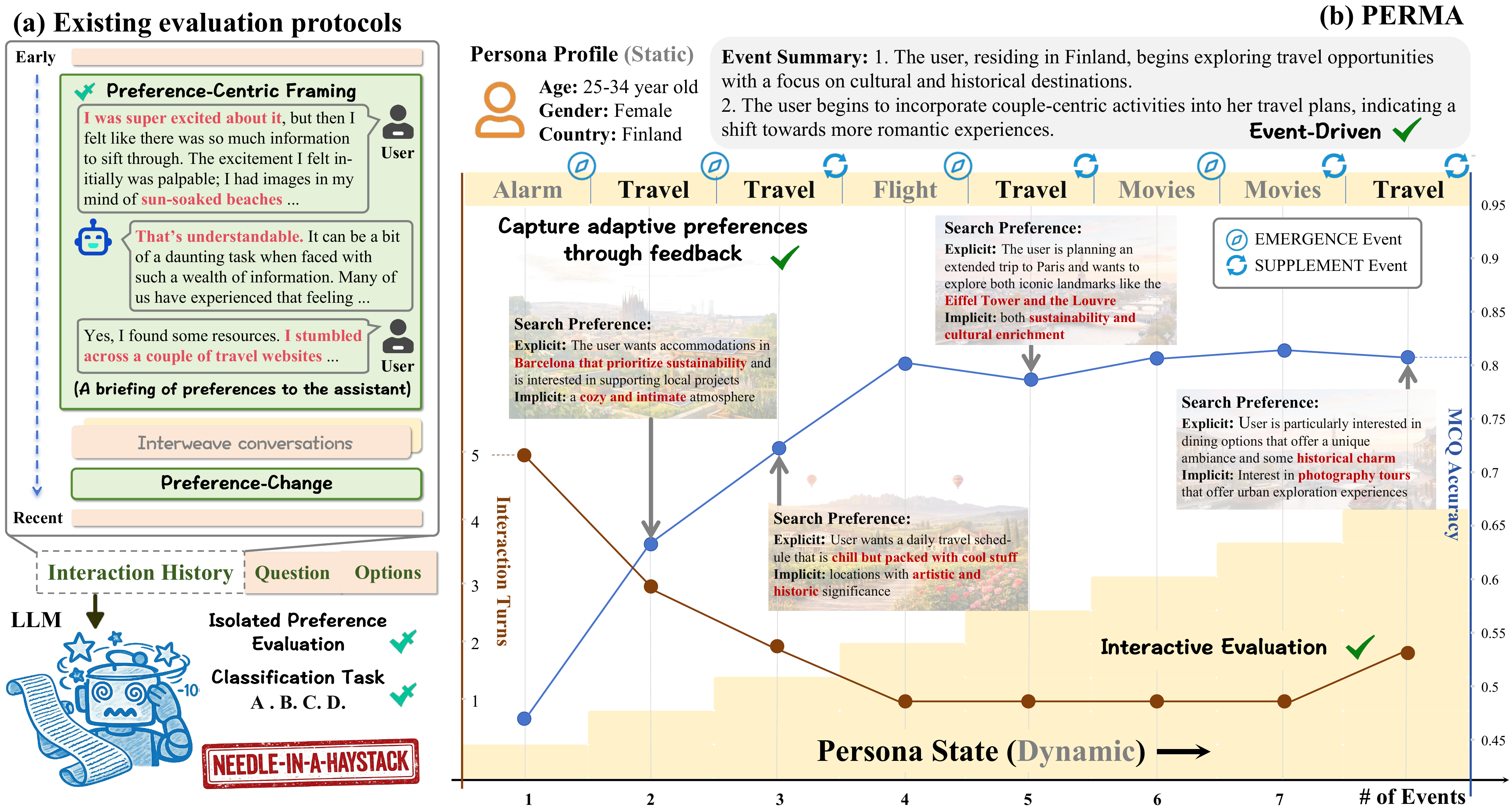}
    \caption{Comparison of context construction and evaluation. (\textbf{Left}) Existing benchmarks: Evaluate isolated preferences via sparse, ``Needle-in-a-Haystack'' retrieval. (\textbf{Right}) \textbf{\method}: 
    Implements an event-driven paradigm where preferences are integrated over time and across sessions to assess the capabilities of memory systems.}
    \label{fig:intro}
\end{figure*}

Along this line, empirical studies~\cite{jiang2025knowmerespondme, zhao2025llmsrecognizepreferencesevaluating} have underscored the inherent difficulties in verifying user-centric personalization, a task typically benchmarked by preference recall fidelity and robustness to token-level noise across extended interactions~\cite{maharana2024evaluatinglongtermconversationalmemory}. While current LLMs excel at factual retention~\cite{wang2025richrag}, effective personalization necessitates a paradigm shift~\cite{bian2026realmembenchmarkingllmsrealworld}: treating user preferences not as static attributes, but as evolving behavioral patterns that must be contextually inferred and extracted during ongoing task interactions~\cite {jiang2025personamemv2personalizedintelligencelearning, xu2026personalizedllmpoweredagentsfoundations}. Thus, the primary bottleneck is not simple memorization, but the \textbf{interpretation}, \textbf{integration}, and \textbf{retrieval} of preference signals that are often latent, incrementally expressed, or obscured by conversational noise. Unfortunately, as illustrated in Figure~\ref{fig:intro}~(a), existing evaluation frameworks remain insufficient for assessing the longitudinal consistency of persona understanding, due to three limitations: (1) \textbf{Preference-Centric Framing}: Existing protocols~\cite{zhao2025llmsrecognizepreferencesevaluating} construct the origins of preferences using static statements from a user perspective, rather than capturing the adaptive nature of preferences through iterative feedback. Such a formulation turns the conversation between user and assistant into a preference briefing, thereby overlooking how preferences manifest through event-driven dialogues. Consequently, the resulting evaluation ignores the interactive nuances of real-world deployment, failing to represent noise-rich scenarios such as ambiguous requests~\cite{li2025surveypersonalizationragagent}. (2) \textbf{Static User Modeling}: As preferences are predefined prior to dialogue, models are evaluated based on isolated user preferences~\cite{zhao2025llmsrecognizepreferencesevaluating, feng2026doespersonalizedmemoryshape}, treating them as static snapshots and neglecting cross-session dependencies that are essential for assessing agents' ability to reason over continuous interactions. It remains difficult to verify whether a model can achieve the cross-domain synthesis and retrieval required for complex queries~\cite{jiang2025knowmerespondme}, often leading retrieval-based approaches to appear effective under recall-oriented settings while their performance is overestimated~\cite{wu2026knowmebenchbenchmarkingpersonunderstanding}. (3) \textbf{Outcome Assessment}: Ideally, a memory-augmented agent should progressively integrate interaction histories and user preferences to enhance the efficiency of subsequent sessions, thereby reducing token overhead and minimizing the need for redundant repetitive user prompting. Current evaluation paradigms primarily focus on the capabilities of standalone LLMs and assess performance mainly based on the final LLM outputs, conflating memory retrieval with the generation process in memory systems rather than evaluating them separately. As a result, the quality and efficiency of the underlying memory systems, as well as memory utility in interactive scenarios, remain unexamined. These limitations highlight the need for alignment between benchmark construction and real-world interaction scenarios.
% These limitations highlight the need for evaluation frameworks that explicitly model temporally evolving user states and assess how agents utilize memory under realistic, noisy interaction settings.
% To address this limitation, evaluation frameworks should explicitly separate memory retrieval from response generation and measure whether agents can effectively utilize retrieved context during interaction.
% Without an assessment of both search results and the agents' ability to utilize retrieved context, it is challenging to discern whether breakdowns in persona consistency stem from memory retrieval or reasoning.

% Rather than focusing solely on long-context reading comprehension or treating preferences as isolated facts, it is increasingly critical to evaluate a model's capacity to adapt to users by learning from feedback accumulated during real-world interactions.
To this end, we introduce \textbf{\method}, a benchmark designed to evaluate the maintenance of persona states, defined as a combination of dynamic user preferences and episodic memory. \method models personalization as a temporally evolving state derived from interaction events, where each session addresses specific task requirements within a targeted topic. Through this process, fine-grained preference details are not explicitly given as standalone statements, but are inferred from how users articulate requests, provide feedback, and refine constraints across turns. These extracted preference signals, together with interaction history, are integrated into the evolving persona state, progressively enriching the user profile. As illustrated in Figure~\ref{fig:intro}~(b), we construct the interaction histories of representative users across \textbf{10} countries and diverse age demographics, spanning over \textbf{20} distinct topic domains and extending the scope to include \textbf{2,166} preference details. Table~\ref{tab:dataset} summarizes the resulting corpus, which contains upwards of \textbf{800} events across domains, with contexts totaling \textbf{1.8M} tokens. To ensure both structural coherence and linguistic realism, we adopt a two-stage generation process as illustrated in Figure~\ref{fig:pipeline}. A timeline generation agent first transforms these events into detailed descriptions and objectives necessary for dialogue reconstruction. Events are categorized into ``emergence'' and ``supplement'' types to distinguish between the emergence of new preferences and the refinement of existing ones across different topics. Subsequently, a dialogue generation agent is prompted to flesh out the full conversation based on the event descriptions and objectives, yielding a \textit{Clean} version with standard, non-rigid interactions. The agent also specifies the concrete preference being realized. To ensure coherence and consistency, we employ an LLM-based verification and human validation pipeline, triggering an iterative re-generation loop whenever preferences are not presented in the dialogue. Drawing on challenges from prior research, we incorporate two key factors to better approximate real-world interactions. First, we introduce \textit{text variability} through the injection of five types of human prompt biases~\cite{tian2025taxonomypromptdefectsllm} into each event as in-session noise, thereby capturing realistic within-session disruptions such as intent shifts, ambiguous expressions, and preference corrections. Second, we model \textit{linguistic alignment} with authentic data by aligning user queries with individual idiolects from WildChat~\cite{zhao2024wildchat1mchatgptinteraction}, fostering diverse phrasing patterns grounded in natural interactions. Finally, we construct three distinct context tiers for comparative analysis: \textit{Clean}, \textit{In-session Noise}, and \textit{Style-aligned Long-context}. Table~\ref{tab:query_taxonomy} provides illustrative examples of injected in-session noise types and style-aligned queries.

%  Building on the challenges identified in prior work, we address the lack of investigation into interaction noise by injecting five types of human prompt biases~\cite{tian2025taxonomypromptdefectsllm} into each event as in-session noise, capturing within-session disruptions such as intent shifts, ambiguous expressions, and preference corrections. In addition, we address the gap in style alignment with real-world data by aligning user queries with human linguistic styles from WildChat~\cite{zhao2024wildchat1mchatgptinteraction} and performing real-data injection. Finally, we produce three distinct context tiers for comparative analysis: \textit{Clean}, \textit{In-session Noise}, and \textit{Style-aligned Long-context}. Table~\ref{tab:query_taxonomy} provides illustrative examples of injected in-session noise types and style-aligned queries derived from real-world conversational styles.
%

\begin{table}[t]
\centering
\caption{Comprehensive statistics of the dataset across multiple dimensions. The table summarizes user demographics for profiles, followed by statistics for the base (comparing \textit{Clean} and \textit{Noise}) and the style-aligned long-context dataset.}
\label{tab:dataset}
\footnotesize
\setlength{\tabcolsep}{1mm}
\begin{tabular}{lccccccccccc}
\toprule
\textbf{Metric} & \textbf{P1} & \textbf{P2} & \textbf{P3} & \textbf{P4} & \textbf{P5} & \textbf{P6} & \textbf{P7} & \textbf{P8} & \textbf{P9} & \textbf{P10} & \textbf{Total} \\ \midrule
\rowcolor[gray]{0.95} \multicolumn{12}{l}{\textit{User Demographics}} \\
Age & 35-44 & 25-34 & 65+ & 55-64 & 35-44 & 18-24 & 25-34 & 35-44 & 25-34 & 25-34 & -- \\
Gender & M & M & F & M & M & M & M & M & F & M & -- \\
Education & Grad. & Univ. & -- & Univ. & -- & Grad. & Univ. & -- & Grad. & Univ. & -- \\
Country & CA & MX & FI & US & AU & UK & CH & IL & RU & BE & -- \\ \midrule

\rowcolor[gray]{0.95} \multicolumn{12}{l}{\textit{Interaction Statistics (Base Dataset: Clean / Noisy)}} \\
\# Interests & 17 & 16 & 15 & 15 & 17 & 13 & 13 & 17 & 16 & 12 & 151 \\
\# Queries & 63 & 62 & 60 & 59 & 63 & 51 & 50 & 63 & 60 & 49 & 580 \\
\# Events & 81 & 81 & 85 & 79 & 82 & 79 & 75 & 81 & 81 & 84 & 808 \\
\# Dialogs & 356/364 & 340/358 & 373/379 & 347/352 & 348/370 & 349/357 & 321/329 & 350/364 & 359/362 & 365/375 & 3.5k/3.6k \\
\# Tokens (k) & 34/34 & 32/33 & 33/36 & 31/31 & 33/33 & 32/32 & 31/31 & 32/36 & 34/39 & 33/34 & 324/331 \\ \midrule

\rowcolor[gray]{0.95} \multicolumn{12}{l}{\textit{Style-aligned Long-Context Dataset}} \\
\# Events & 156 & 141 & 156 & 139 & 137 & 107 & 176 & 94 & 155 & 127 & 1,388 \\
\# Dialogs & 1009 & 848 & 1084 & 876 & 888 & 802 & 1015 & 798 & 969 & 869 & 9,158 \\
\# Tokens (k) & 139.0 & 61.4 & 174.0 & 88.9 & 197.2 & 69.0 & 151.1 & 62.7 & 143.7 & 78.4 & 1,165.4 \\ 
\bottomrule
% \addlinespace
% \multicolumn{12}{p{\textwidth}}{\footnotesize \textit{Note: For User Demographics Profile (P), Gender: M (Male), F (Female); Education: Grad. (Graduate Degree), Univ. (University); Country codes follow ISO 3166-1 alpha-2 (e.g., CA: Canada, MX: Mexico). In Interaction Statistics, \# denotes the count, and (k) represents thousands of tokens.}} \\
\end{tabular}
\end{table}

\begin{table*}[t]
\centering
\small
\caption{Taxonomy of in-session noise types (1-5) and style-aligned queries (6). The structure highlights the agent's ability to resolve ambiguity (\colorbox{red!25}{noise}), maintain persona coherence (\colorbox{green!25}{memory}).}
\resizebox{\textwidth}{!}{
\begin{tabularx}{\textwidth}{>{\bfseries\color{blue}}l X}
\hline
\textbf{Type} & \textbf{Interaction Examples (U: User, A: Assistant)} \\ 
\hline

[1] Omitted Info. 
& \textit{\textbf{U:} Can you give me guidance on \colorbox{red!25}{that game}? $\rightarrow$ \textbf{A:} Could you specify which game? $\rightarrow$ \textbf{U:} \colorbox{green!25}{The Witcher 3}. $\rightarrow$ \textbf{A:} Certainly. I suggest optimizing the PC graphics and ...} \\ 

[2] Context Switch 
& \textit{\textbf{U:} \colorbox{red!25}{By the way}, do you remember my hockey tickets? $\rightarrow$ \textbf{A:} Yes, Maple Leafs tickets for Nov 5th... $\rightarrow$ \textbf{U:} Thanks. Now back to the wine tasting... $\rightarrow$ \textbf{A:} Understood. Regarding the \colorbox{green!25}{culinary events}...} \\ 

[3] Inconsistent Pref. 
& \textit{\textbf{U:} For this week, I want to see the \colorbox{red!25}{biggest} conference. $\rightarrow$ \textbf{A:} The Global Tech Summit (10k+ attendees) is happening now. $\rightarrow$ \textbf{U:} Those are too crowded. Maybe something \colorbox{green!25}{low-key} would be better.} \\ 
% $\rightarrow$ \textbf{A:} Tech Innovators Meetup would be a fit for \colorbox{green!25}{small-group networking}.

[4] Multi-lingual 
& \textit{\textbf{U:} ``Je voudrais explorer les dégustations de vins locaux.'' $\rightarrow$ \textbf{A:} ``Certainement. Préférez-vous des accords avec du fromage ou du chocolat ?'' $\rightarrow$ \textbf{U:} ``Avec du fromage, s'il vous plaît.'' $\rightarrow$ \textbf{A:} ``Parfait. J'ai trouvé plusieurs \colorbox{green!25}{événements gastronomiques} dans votre région pour ce week-end.''} \\

[5] Colloquial Exp. 
& \textit{\textbf{U:} I'm trying to \colorbox{red!25}{vibe} at \colorbox{red!25}{dope} tech events ASAP. $\rightarrow$ \textbf{A:} Understood. Are you looking for large summits or local meetups? $\rightarrow$ \textbf{U:} Keep it \colorbox{red!25}{chill and small}. $\rightarrow$ \textbf{A:} Here are a few \colorbox{green!25}{boutique tech gatherings ...}} \\ 
%  $\rightarrow$ \textbf{A:} Got it. Here are a few \colorbox{green!25}{boutique tech gatherings} that facilitate one-on-one professional connections.

[6] Style-aligned
& \textit{\textbf{U:} I need help organizing my calendar for a recurring event. I'd like to set up weekly family lunches on Sundays with my family. Can you assist me with that? $\rightarrow$ \textbf{(styled-aligned) U:} Trying to \colorbox{lightgray}{set up weekly family lunches} on \colorbox{lightgray}{Sundays} with the \colorbox{lightgray}{whole gang}.} \\ 

\hline
\end{tabularx}}
\label{tab:query_taxonomy}
\end{table*}
% \vspace{-10pt}

To conclude, \method enables us to investigate the impact of persona construction across multiple domains, natural prompting noise, and temporal evolution. To evaluate LLM agent behavior under these settings, we design multiple-choice questions (MCQs) from three perspectives, namely \textit{task completion}, \textit{preference consistency}, and \textit{information confidence}, to examine whether agents can retrieve and apply historical information in one shot. Meanwhile, to simulate real-world multi-turn interaction, we further implement an LLM-based user simulator that provides supplemental information when preference is unmet. This creates an iterative feedback loop between the simulator and the LLM agent, which continues until the generated response satisfies the underlying preference. We benchmark a diverse set of standalone LLMs and agents equipped with memory systems, analyzing their effectiveness in contextualizing user preferences and retrieving relevant memory segments. Furthermore, we conduct probing evaluations at various temporal intervals along the dialogue timeline to examine how performance evolves as persona states accumulate and potentially drift. This comprehensive analysis offers critical insights into the underlying mechanics of memory systems and their capacity to sustain coherent persona states under realistic, long-term interaction conditions. Our contributions are summarized below:

% Beyond evaluation, we further explore whether persona-consistent behavior can be improved through learning on our benchmark. By applying RL training on \method, we observe consistent gains not only on our test set but also across external preference benchmarks, demonstrating its effectiveness for generalizable personalization.
\begin{enumerate}[leftmargin=*, topsep=0pt, itemsep=0pt]

\item We introduce a benchmark \method for evaluating LLM agents that shifts the paradigm from static preference recall to preference synthesis based on event-driven personalized dialogue scenarios. \method incorporates realistic in-session noise and linguistic style to assess how well agents infer and extract preferences through iterative interactions.

\item We formulate a dynamic user modeling evaluation protocol that tests an agent's ability to maintain persona. By focusing on cross-domain synthesis, \method evaluates the agent's capacity to reason over multiple sessions. Importantly, the evaluation queries reflect natural user requests, enabling assessment of memory utilization in realistic application scenarios rather than recall-oriented settings.

\item We design a comprehensive assessment suite across multiple dimensions. By employing a user simulator to facilitate multi-turn interactions and incorporating temporal probing, we assess how agents synthesize and apply memory over evolving preferences, revealing their strengths and limitations under temporal drift and semantic interference.
% \noindent\textbf{4}) We conduct a systematic study under composite preference reasoning scenarios. Our analysis reveals the strengths and limitations of memory systems in sustaining coherent persona states under temporal drift and semantic interference.

% \noindent\textbf{4}) We further demonstrate that reinforcement learning on \method enhances persona consistency and generalization, yielding consistent improvements not only on our benchmark but also across external preference datasets.

\end{enumerate}

%% file: content/01_relatedwork.tex
\subsection{Evolution of Personalization Memory Benchmarks}
\subsubsection{Evaluating Long-Context Memory Capabilities of LLMs}
The evaluation of long-context capabilities in LLMs~\cite{wang2025mmlongbenchbenchmarkinglongcontextvisionlanguage, yang2025qwen3technicalreport, yen2025helmetevaluatelongcontextlanguage} has historically relied on ``Needle-in-a-Haystack'' style tests that measure a model’s ability to retrieve facts embedded within lengthy contexts~\cite{openai2024gpt4technicalreport, liu2023lostmiddlelanguagemodels}. While early benchmarks were limited to 10k-token interactions, recent efforts~\cite{maharana2024evaluatinglongtermconversationalmemory, tan2025personabench, tavakoli2026milliontokensbenchmarkingenhancing}, such as LongMemEval~\cite{wu2025longmemevalbenchmarkingchatassistants} extend evaluation to ultra-long multi-session dialogues ranging from 115k to over 1.5M tokens, incorporating tasks that probe cross-session memory and free-form question answering. Nevertheless, these benchmarks primarily evaluate whether models can access and reason over long interaction histories, rather than how user preferences are incrementally inferred and integrated during ongoing interactions for personalization. Table~\ref{tab:comparison} provides a comparison between existing memory and personalization evaluations across multiple dimensions.

\subsubsection{Towards Personalization in Memory}

At the core of personalized interaction and lifelong digital companions~\cite{wu2026knowmebenchbenchmarkingpersonunderstanding} lies the concept of user preferences~\cite{zhang2025proactivepersonalizationprofilecustomization, Ren_2019, du2024perltqapersonallongtermmemory}, defined as the longitudinal collection of a user's stable interests, stylistic inclinations, and task-specific constraints~\cite{he2026memoryarenabenchmarkingagentmemory}. For a personalized agent~\cite{li2025helloagainllmpoweredpersonalized, xu2026personalizedllmpoweredagentsfoundations, hu2026opbenchbenchmarkingoverpersonalizationmemoryaugmented}, these preferences constitute a latent state that must be captured from historical interactions to guide subsequent response generation. This ensures that agent behavior aligns with individual expectations without necessitating repetitive, explicit instructions. The fundamental necessity of preference modeling is to alleviate the ``interaction burden'' -- empowering agents to proactively anticipate user needs from historical context, thereby enhancing efficiency and user satisfaction in real-world applications.

Broadly, personalization benchmarks have evolved from (i) static preference conditioning to (ii) multi-session preference retention, yet few explicitly evaluate whether models can maintain a coherent persona state under evolving, information-dense interactions. Early frameworks, such as PrefEval~\cite{zhao2025llmsrecognizepreferencesevaluating} and PerLTQA~\cite{du2024perltqapersonallongtermmemory}, characterized personalization as conditional response generation, focusing on an LLM's capacity for long-context understanding, user modeling and adherence to a constrained set of stated preferences. These studies revealed that contemporary LLMs struggle to infer implicit preferences accurately and apply them consistently across conversational turns, particularly when intervening dialogues introduce unrelated topics as ``contextual token noise''~\cite{tan2025personabench}. Despite these advancements, it remains designed for ``Needle-in-a-Haystack'' tasks, in which isolated preferences are retrieved from vast volumes of unrelated context. To enhance the recognition of personalized intent, UserBench~\cite{qian2025userbenchinteractivegymenvironment} and PersonaLens~\cite{zhao2025personalensbenchmarkpersonalizationevaluation} introduced interactive, gym-like environments. These benchmarks advanced the field toward simulating real-world agent dynamics; however, they treat user profiles as fixed inputs rather than evolving constructs. A pivotal shift toward memory retention occurred with PersonaMem~\cite{jiang2025knowmerespondme}, which incorporated multi-session interaction histories. PersonaMem centers on preference-driven dialogue construction; consequently, its dialogues resemble subjective opinion sharing rather than objective problem solving. Subsequently, PersonaMem-v2~\cite{jiang2025personamemv2personalizedintelligencelearning} further advances this paradigm by focusing on implicitly expressed preferences and requiring LLMs to infer user traits from subtle conversational nuances. It assumes that implicit behavioral cues can be consistently abstracted into stable preference variables. However, in realistic interactions, many implicit signals are situational and do not necessarily correspond to enduring persona states. These frameworks still evaluate preference recognition and recall within constructed interaction traces, without explicitly probing the stability, cross-domain synthesis, and temporal consistency of the underlying persona state. Recent efforts, such as the KnowMe-Bench~\cite{wu2026knowmebenchbenchmarkingpersonunderstanding}, use long-form narratives as a data substrate to evaluate evidence-grounded inference about motivation, identity development, and psychological states. Yet KnowMe-Bench is better characterized as a reasoning benchmark rather than as an assessment of personalized conversational user-assistant interaction or dialogue management. 

To address these limitations, \method introduces a dynamic evaluation paradigm centered on active dialogue management. We shift the evaluative focus from single-point preference recall to composite synthesis, requiring memory systems to integrate interdependent traits across information-dense sessions to resolve complex queries. By employing an event-driven methodology, our framework captures the evolutionary nature of user profiles; specifically, the interplay between newly emerged and updated preferences, while moving beyond ``black-box'' long-context reasoning evaluation through a decoupled assessment of retrieved memory quality and generative reasoning. As a result, \method reframes personalization evaluation from preference identification to persona state maintenance across temporally ordered events with structured noise and cross-domain dependencies. Such requirements extend beyond traditional dialogue retrieval, motivating the development of dedicated memory systems and memory-enabled agents for persistent user modeling.

\begin{table*}[t]
\caption{Comparison of \method with existing benchmarks for persona modeling and memory evaluation. \textit{Event-driven Preference} refers to preference formation through interaction events; \textit{Context Noise} denotes the presence of irrelevant or distracting context; \textit{Temporal Probing} evaluates performance across multiple time points. \cmark, \pmark, and \xmark denote full support, partial support, and absence of the corresponding feature, respectively.}
\label{tab:comparison}
\centering
\small
\resizebox{\textwidth}{!}{
\begin{tabular}{lccccccccc}
\toprule
\textbf{Benchmark}
& \makecell{\textbf{Dynamic}\\\textbf{Preference}}
& \makecell{\textbf{Event-driven}\\\textbf{Preference}}
& \makecell{\textbf{Context}\\\textbf{Noise}}
& \makecell{\textbf{Implicit}\\\textbf{Preference}}
& \makecell{\textbf{Cross-domain}\\\textbf{Reasoning}}
& \makecell{\textbf{Memory}\\\textbf{Representation}}
& \makecell{\textbf{MCQ}\\\textbf{Evaluation}}
& \makecell{\textbf{Interactive}\\\textbf{Evaluation}}
& \makecell{\textbf{Temporal}\\\textbf{Probing}} \\
\midrule
\rowcolor[gray]{0.95} \multicolumn{10}{l}{\textit{Episodic Fact Memory}} \\

LoCoMo~\cite{maharana2024evaluatinglongtermconversationalmemory} 
& \xmark & \xmark 
& \xmark 
& \xmark 
& \cmark 
& Episodic Memory 
& \xmark 
& \xmark 
& \cmark \\

LongMemEval~\cite{wu2025longmemevalbenchmarkingchatassistants}
& \xmark & \xmark 
& \xmark 
& \xmark 
& \xmark 
& Episodic Memory  
& \xmark 
& \xmark 
& \cmark \\

MemoryBench~\cite{ai2025memorybenchbenchmarkmemorycontinual} 
& \cmark & \xmark 
& \xmark 
& \xmark 
& \xmark 
& Declarative Memory
& \xmark 
& \xmark 
& \cmark \\

\midrule
\rowcolor[gray]{0.95} \multicolumn{10}{l}{\textit{Personalized Agent}} \\

% PersonaChatGen~\cite{lee-etal-2022-personachatgen} 
% & \xmark & \xmark 
% & \xmark 
% & \xmark 
% & \xmark 
% & - 
% & \xmark 
% & \cmark 
% & \xmark \\

PersonaLens~\cite{zhao2025personalensbenchmarkpersonalizationevaluation} 
& \xmark & \xmark 
& \xmark 
& \cmark 
& \cmark 
& - 
& \xmark 
& \cmark 
& \xmark \\

UserBench~\cite{qian2025userbenchinteractivegymenvironment} 
& \cmark & \cmark 
& \xmark 
& \cmark 
& \xmark 
& - 
& \xmark 
& \cmark 
& \xmark \\

\midrule
\rowcolor[gray]{0.95} \multicolumn{10}{l}{\textit{Personalized Memory}} \\

PerLTQA~\cite{du2024perltqapersonallongtermmemory} 
& \xmark & \xmark 
& \xmark 
& \xmark 
& \cmark 
& Episodic Memory
& \xmark 
& \xmark 
& \xmark \\

PrefEval~\cite{zhao2025llmsrecognizepreferencesevaluating} 
& \xmark & \xmark 
& \pmark 
& \cmark 
& \xmark 
& Single-session 
& \cmark 
& \xmark 
& \cmark \\

PersonaMem~\cite{jiang2025knowmerespondme} 
& \cmark 
& \xmark 
& \pmark 
& \pmark 
& \xmark 
& Dynamic Memory 
& \cmark 
& \xmark 
& \cmark \\

PersonaMem-v2~\cite{jiang2025personamemv2personalizedintelligencelearning} 
& \cmark 
& \cmark 
& \xmark 
& \cmark 
& \xmark 
& Implicit Preference 
& \cmark 
& \xmark 
& \xmark \\

KnowMe-Bench~\cite{wu2026knowmebenchbenchmarkingpersonunderstanding} 
& \xmark 
& \xmark 
& \xmark 
& \cmark 
& \xmark 
& Autobiographical 
& \xmark 
& \xmark 
& \xmark \\

\midrule
\textbf{\method (Ours)} 
& \cmark 
& \cmark 
& \cmark 
& \cmark 
& \cmark 
& Persona State 
& \cmark 
& \cmark 
& \cmark \\

\bottomrule
\end{tabular}}
\end{table*}

\subsection{From Static Retrieval to Persistent Memory} 
Personalization in LLMs has undergone a fundamental architectural shift driven by the growing mismatch between transient context conditioning and the requirements of long-term user–assistant relationships~\cite{xu2026personalizedllmpoweredagentsfoundations, li2024personalllmagentsinsights}. Early approaches implicitly equated personalization with prompt augmentation, assuming that user history could be sufficiently represented within a fixed or expanded context window. However, this assumption fails at scale: attention-based architectures exhibit context dilution and the ``lost-in-the-middle'' phenomenon, rendering long-term user modeling inefficient~\cite{liu2023lostmiddlelanguagemodels}.

Retrieval-Augmented Generation (RAG)~\cite{gao2024retrievalaugmentedgenerationlargelanguage, lyu2024crudragcomprehensivechinesebenchmark} reframed this problem by externalizing memory into non-parametric storage. By enabling semantic access to repositories of user-related information, RAG alleviated context window constraints and improved factual grounding. Yet this architectural move introduced a new limitation: memory was reduced to static, document-like artifacts retrieved reactively based on semantic similarity~\cite{li2025surveypersonalizationragagent, memorag}. RAG systems remain inherently stateless, primarily serving as indexing mechanisms for user-assistant dialogue rather than modeling the temporal evolution and coherence of user personas~\cite{neelakanteswara-etal-2024-rags}.

Unlike passive retrieval in vanilla RAG, agentic frameworks utilize LLMs as reasoning engines to orchestrate tool use~\cite{AWM, yuan2025memsearchertrainingllmsreason} and multi-step planning~\cite{zhang2026memoryactionautonomouscontext}. However, for an agent to exhibit consistent personality and long-term adaptation, it requires a persistent memory system, which functions as the agent's long-term storage. Recent work~\cite{zhang2024surveymemorymechanismlarge, du2026memoryautonomousllmagentsmechanisms, xu2025amemagenticmemoryllm, wang2025mirixmultiagentmemoryllmbased} has begun to formalize this transition by treating memory as a structured, lifecycle-aware subsystem. Rather than storing immutable embeddings, memory architectures introduce mechanisms for consolidation, reinforcement, decay, and reorganization~\cite{hu2025memoryageaiagents, 10.1145/3774946}. This marks a shift from storage-centric memory to process-oriented memory, in which the relevance of stored information is dynamically updated over time. Memory systems that support fine-grained updates recognize that personalization must operate at multiple levels, including episodic experience, user preferences, and behavioral patterns. MemOS~\cite{li2025memosmemoryosai} was among the first to formalize memory as a core system component, introducing a lifecycle management mechanism (MemCube) that enables dynamic memory scheduling, aging out obsolete facts while reinforcing relevant ones over long horizons. Similarly, Mem0~\cite{chhikara2025mem0buildingproductionreadyai} is a scalable memory‑centric architecture that dynamically extracts, consolidates, and retrieves salient conversational information from ongoing interactions, supporting persistent, structured memory beyond raw document retrieval. Memobase focuses on extracting and maintaining structured user profiles and tagged memory records from chat interactions, serving as a scalable backend for personalized applications. Supermemory provides a unified memory API that builds structured relationships between facts to support persistent recall and relational context beyond simple vector proximity. Compared to existing benchmarks (Table~\ref{tab:comparison}), the differentiation of \method lies in the multi-dimensional decoupling of its evaluation. While LLMs and memory systems have improved in memory capacity, they still lack a fine-grained characterization of event-driven persona state evolution, which is essential for reflecting the mastery of user preferences. Furthermore, by integrating context noise and temporal probing, \method enables us to verify the coherence persona state within complex event streams, independent of mere interaction quality.

% \textbf{\# Tokens} & 34106/33968 & 32449/32975 & 33393/36547 & 30963/30917 & 32768/33273 & 32106/31864 & 30797/31180 & 32196/35928 & 34299/39283 & 33201/34439 & \textbf{324109/331393} \\ \midrule While these systems advance memory engineering, existing evaluations rarely isolate whether architectural improvements actually enhance persona state coherence over time. In contrast, our decoupled, state-aware evaluation measures this capability, highlighting the effect of memory design on maintaining coherent user states. We further complement this with interactive evaluations, which assess memory utilization in realistic, temporally structured interactions, moving beyond conventional recall- or classification-based metrics.

%% file: content/02_preliminary.tex
To formally characterize the construction of an event-driven dialogue dataset, we establish a mathematical framework defining the constituent elements of the user profile and the temporal interaction sequence.

\begin{definition}
\textbf{Data Specification.} Let $\mathcal{U}$ denote the set of users. For each user $u \in \mathcal{U}$, the user profile is defined as a triplet $\mathcal{P}_u = \{ \mathcal{D}_u, \mathcal{I}_u, \mathcal{E}_u \}$
where $\mathcal{D}_u$ is the demographic profile (e.g., age and gender). $\mathcal{I}_u$ denotes the set of user interests,
%%
% where $topic_k$ is a specific domain (e.g., Movie) and $w_k \in \{0, 1\}$ indicates interest presence in the domain. 
%%
$\mathcal{E}_u$ denotes the temporally ordered interaction events between the user and assistant.
\end{definition}

% The timestamp $ts_t$ follows a temporal order, $ts_t>ts_{t-1}$.

\begin{definition} \label{def:type}
\textbf{Interaction Event.} An interaction event in $\mathcal{E}_u$ at time step $t$ is defined as a tuple $e_t = (\tau_t, \text{dom}_t, \mathcal{C}_t, \Phi_t)$, where $\tau_t \in \{ \textsc{Emergence}, \textsc{Supplement}, \textsc{Task} \}$ denotes event type:
\begin{itemize}[topsep=0pt, itemsep=0pt]
    \item $\textsc{Emergence}$: The user's first discovery of a domain, establishing initial preferences.
    \item $\textsc{Supplement}$: A follow-up interaction where preferences are refined, modified, or deepened.
    \item $\textsc{Task}$: A task-oriented user query probing the model’s integration of prior preferences.
\end{itemize}

$\text{dom}_t \subseteq \mathcal{I}_u$ is the set of relevant domains involved in this event. $\mathcal{C}_t = [ (q_t^{(1)}, a_t^{(1)}), \dots, (q_t^{(N)}, a_t^{(N)}) ]$ represents the multi-turn dialogue content, consisting of $N$ turns of user queries $q_t^{(k)}$ and assistant responses $a_t^{(k)}$. $\Phi_t$ is the set of explicit or implicit preferences revealed during this event.
% Each interaction event induces a transition from $\mathcal{S}_{t-1}^u$ to $\mathcal{S}_t^u$, forming a temporally ordered state trajectory that constitutes the longitudinal personalization process.   = f(e_1, \dots, e_{t-1})

When $\tau_t = \textsc{Task}$, $\Phi_t$ is empty and $\mathcal{C}_t = [q_t^{(1)}]$. This event serves as a checkpoint to evaluate the model’s response to preference-relevant queries (i.e.\ $q_t^{(1)}$). A task can be \textbf{Single-Domain} if $|\text{dom}_t| = 1$ or \textbf{Multi-Domain} if $|\text{dom}_t| > 1$, with the latter requiring the joint synthesis of preferences across domains. Each task utilizes the current persona state as a user preference prior. Tasks are classified into three types based on their position in the interaction history:
\begin{itemize}[topsep=0pt, itemsep=0pt]
    \item \textbf{Type 1} (Zero-Memory): Evaluated at the onset of interaction before relevant preferences are established.
    \item \textbf{Type 2} (In-Time): Positioned immediately after all relevant sessions in $\text{dom}_t$ have occurred.
    \item \textbf{Type 3} (Post-Intervention): Positioned after a series of sessions containing unrelated topics.
\end{itemize}

\end{definition}

\begin{definition}
\textbf{Persona State.}
For a user $u$, we define the persona state at time $t$ as a text representation 
$\mathcal{S}_t^u$ that synthesizes integrated user preferences and  episodic memory derived from the concatenated dialogues 
$\mathcal{G}_{u, \leq t} = \{\mathcal{C}_1, \mathcal{C}_2, \dots, \mathcal{C}_t\}$ corresponding to historical events 
$\mathcal{E}_{u, \leq t} = (e_1, e_2, \dots, e_t)$:
\begin{equation}
\mathcal{S}_t^u = f(\{\mathcal{C}_1, \mathcal{C}_2, \dots, \mathcal{C}_t\}),
\end{equation}
where for memory systems, $f$ denotes an operator that aggregates the persona state by incorporating newly emerged or updated preferences from historical dialogues. In contrast, for standalone LLMs and vanilla RAG systems, $f$ reduces to an identity mapping over the raw interaction history. Unlike the static user profile $\mathcal{P}_u$, the persona state $\mathcal{S}_t^u$ evolves dynamically, reflecting incremental preference refinements and potential drift, thus enriching $\mathcal{P}_u$ with fine-grained preference and behavioral details.
\end{definition}

% where $f$ denotes a state transition operator that updates the persona state with newly emerged or refined preferences and cross-domain interactions. Each interaction event induces a transition from $\mathcal{S}_{t-1}^u$ to $\mathcal{S}_t^u$, forming a temporally ordered trajectory that captures the evolution of user-specific preferences. Unlike the static user profile $\mathcal{P}_u$, the persona state evolves over time, reflecting incremental refinements and potential drift while enriching $\mathcal{P}_u$ with fine-grained preference and behavioral details.

\begin{definition}
\textbf{Persona Consistency Objective.} 
Given a sequence of historical events 
$\mathcal{E}_{u, \leq t} = (e_1, \dots, e_{t})$, where $e_{t}$ is a task event, our benchmark requires the agent to generate a response $a_t$ 
to the query $q_t$ that maximizes alignment with the user's evolved persona state extracted from the interaction dialogues $\mathcal{G}_{u,\leq t}$. 
Formally, the model $P$ seeks to optimize the objective:
\begin{equation}
a = \arg\max_{a'} \; P\big(a' \mid q_t, \mathcal{S}_{t}^u \big) \quad \text{s.t.} \quad \mathcal{S}_{t}^u = f(\mathcal{G}_{u,\leq t}),
\end{equation}
where $a$ should satisfy the joint constraints of the demographic attributes 
in $\mathcal{D}_u$ and the preference updates encapsulated in 
$\Phi_{1:t}$. To this end, the agent must navigate multi-session and cross-domain histories to integrate relevant memory segments.
\end{definition}

\begin{definition}
\textbf{Memory System.}
Compared to providing raw historical interactions $\mathcal{G}_{u,\leq t}$ and user query $q_t$ to a standalone LLM to generate a response $a$, memory systems $\mathcal{M}$ implement a dual-process mechanism composed of an ingestion operation (\textsc{Add}) and a retrieval operation (\textsc{Search}). 
Given the completed interaction sessions $\mathcal{G}_{u,\leq t}$ of user $u$ observed at timestamp $t$, the system updates its internal persona state by \textsc{Add}:
\begin{equation}
\mathcal{S}^u_t \leftarrow \textsc{Add}(\mathcal{M}, \mathcal{G}_{u,\leq t}),
\end{equation}
where the ingestion process may involve indexing conversational history, extracting salient facts, or updating structured user persona state, such as a knowledge graph. 
Subsequently, for a user query $q_{t}$, the system retrieves a set of contextually relevant memory segments $\mathbf{s}_{t} \subseteq \mathcal{S}_t^u$ by \textsc{Search}:
\begin{equation}
\mathbf{s}_{t} \leftarrow \textsc{Search}(\mathcal{S}^u_t, \mathcal{M}, q_{t}, k),
\end{equation}
where $k$ denotes the number of retrieved items. The effectiveness of a memory system $\mathcal{M}$ is reflected by the utility of $\mathbf{s}_{t}$ in answering user query $q_t$.
\end{definition}

%% file: content/03_method.tex
\begin{figure*}[!t]
    \centering
    \includegraphics[width=\textwidth]{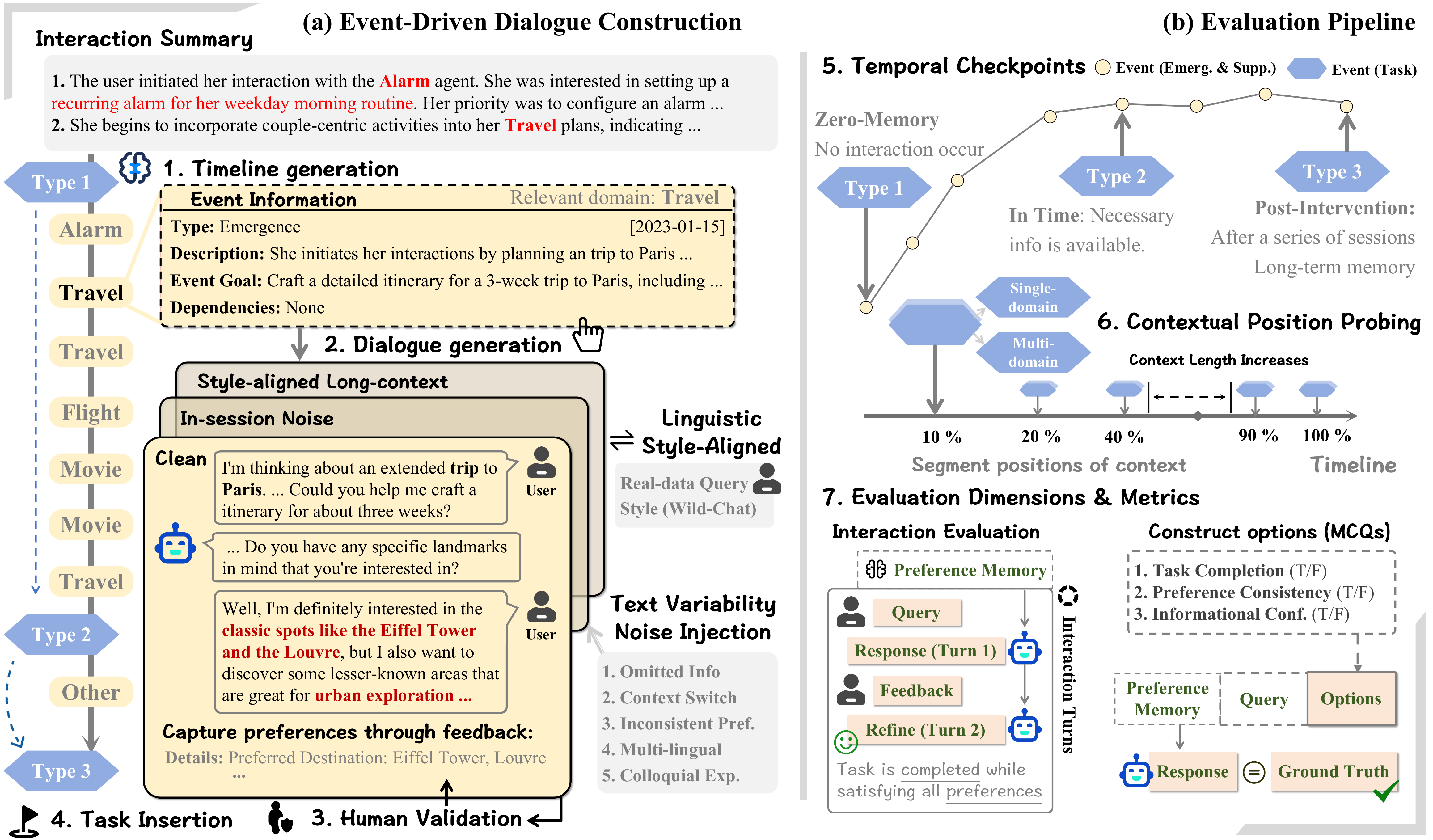}
    \caption{The \method pipeline for dialogue construction and evaluation. \textbf{Left}: The dialogue construction pipeline leverages User Profiles and domain-specific Interaction Summaries to generate a structured timeline. \textbf{Right}: Evaluation of \textsc{Task} Events is conducted through two protocols: (1) One-shot MCQ probing, which measures selection accuracy across three evaluation dimensions to assess zero-shot preference recall; and (2) Interactive evaluation, involving multi-turn dialogues where a user simulator assesses task completion and preference satisfaction, while providing corrective feedback for suboptimal responses. Both evaluation protocols are executed across varying temporal depths within the full dialogue history.}
    \label{fig:pipeline}
\end{figure*} 

In this section, we introduce \textbf{\method}. Rather than treating personalization as static preference recall, our framework models personalization as evolving preferences through an event-driven reconstruction of dialogue histories (Section~\ref{sec:event}). To better reflect real-world user-assistant interactions, Section~\ref{sec:noise} introduces two interaction variants by (1) injecting text variability based on common user prompt biases~\cite{tian2025taxonomypromptdefectsllm}, (2) aligning query linguistic styles with real-world data. We evaluate downstream queries that do not explicitly trigger preference retrieval, thereby testing whether agents can proactively leverage stored information to integrate and enforce composite, cross-domain preferences (Section~\ref{sec:method:eval}).

\subsection{Event-Driven Dialogue Reconstruction}
% \subsection{Phase I: Chronological Timeline Generation}
\label{sec:event}
We construct dialogue histories based on interaction summaries from PersonaLens~\cite{zhao2025personalensbenchmarkpersonalizationevaluation}, which are grounded in authentic user profiles derived from the PRISM dataset~\cite{kirk2024prismalignmentdatasetparticipatory}. These profiles span 20 distinct domains, ranging from Movies to Flights. We extend these concise summaries into long-horizon, consistent dialogue histories that capture the gradual emergence and iterative refinement of user preferences.

As illustrated in Figure~\ref{fig:pipeline} (a), we decompose each domain-level interaction summary into a sequence of chronological events, where each event corresponds to a coherent interaction session and remains consistent with the user profile. For each event in the timeline, we generate (1) an event description that defines the conversational scope and the key behavioral cues, (2) an interaction goal that specifies the intended dialogue outcome, and (3) a set of dependencies that determine the generation ordering among events. This event specification constrains dialogue generation to realize concrete behavioral signals, rather than vague conversational templates. During the generation, the agent is provided with user preference categories and the aforementioned context. Guided by event descriptions and objectives, the dialogue generation agent elaborates on the necessary preference details through either explicit user descriptions or implicit feedback.

\subsubsection{Phase I: Timeline and Dialogue Generation}
\label{sec:method:phase1}
Given an interaction history summary $\mathcal{H}_u$ provided by Personalens~\cite{zhao2025personalensbenchmarkpersonalizationevaluation} and a user demographic profile $\mathcal{D}_u$, we employ a capable LLM (e.g., GPT-4o) as a high-level planner to construct an interaction timeline $\mathcal{T}$. Formally, let $f_{\text{plan}}$ denote the LLM timeline planner. For each topic domain $k \in \mathcal{I}_u$, the planner generates a domain-specific timeline conditioned on the corresponding historical summary $\mathcal{H}_u^{k}$:
\begin{equation}
\mathcal{T}_k = f_{\text{plan}}(\mathcal{D}_u, k, \mathcal{H}_u^{k}).
\end{equation}

We first prompt the model to categorize timeline entries into two types: preference emergence and preference supplement as introduced in Definition~\ref{def:type}, thereby modeling iterative preference refinement. Each entry $\hat{e} \in \mathcal{T}_k$ is further associated with a detailed conversation description and cross-domain dependencies to facilitate inter-domain linkage.

% We generate dialogues iteratively for each domain by feeding the domain's historical conversations as well as the current timeline entry to the generation agent. 

Let $\mathcal{T}_k = \{\hat{e}_1, \dots, \hat{e}_M\}$ denote the chronological sequence of $M$ timeline entries associated with user $u$ within the same domain $k$. 
We iteratively generate a multi-turn dialogue $\mathcal{C}_m$ for each timeline entry $\hat{e}_m$ using an LLM $P_g$, conditioned on the user profile $\mathcal{D}_u$, interest categories $\mathcal{I}_u$ and the previously generated dialogue history $\mathcal{G}^k_{u, m-1} = \{\mathcal{C}^k_1, \dots, \mathcal{C}^k_{m-1}\}$.
\begin{equation}
\mathcal{C}^k_m, \Phi^k_m \sim P_g(\cdot \mid \mathcal{D}_u, \mathcal{I}_u, \hat{e}_t, \mathcal{G}^k_{u, m-1}) \label{eq:cm}
\end{equation}
where $\Phi_m$ represents any new preferences that emerged or updated during this session. For the initial preference emergence entry within a domain, where no prior interactions exist, the entry itself serves as the starting point. We validate generated results with human experts and an LLM-based verification pipeline. The final event is constructed by concatenating the timeline entry $\hat{e}$ with the corresponding dialogue $\mathcal{C}^k_m$ and preferences, i.e., $e_m = \textbf{concat}(\hat{e}_m, \mathcal{C}^k_m, \Phi^k_m)$.

After generating all events across domains, we construct a global timeline $\mathcal{T}_{\text{final}}$ by interleaving the domain-specific timelines $\{\mathcal{T}_k\}_{k=1}^K$. 
This aggregation respects explicitly defined cross-domain dependencies and strictly enforces chronological order according to the parsed timestamps, with a unified time index $t$ across all events.

\subsubsection{Phase II: Task Insertion}
\label{sec:Insertion}
To evaluate memory utilization at varying temporal depths, we inject evaluation tasks into $\mathcal{T}_{\text{final}}$ using a timestamp-based insertion strategy that probes the lifecycle of preference development. For a task event $e_t (\tau_t = \textsc{Task})$ associated with relevant domains $dom_t \subset \mathcal{I}_u$, we establish three types of evaluation checkpoints (Definition~\ref{def:type}). As shown in Figure~\ref{fig:pipeline} (b), we place Type 1 (Zero-Memory) tasks at the onset of the timeline, before any relevant events occur, serving as a non-personalized control to separate the parametric bias of the agent from persona-based answers. Type 2 (In-Time) task is placed right after the final events of relevant domains to evaluate the integration of the entire trajectory. 
%%
% Formally, the timestamp index for the probing is defined as,
% \begin{equation}
%     t_{\text{Type2}} = \max \{ t \mid \forall \delta \in dom_t, \exists e_k \in \mathcal{E}_{<t} \text{ s.t. } \text{domain}(e_k) = \delta \} + 1.
% \end{equation}
%%
Additionally, to quantify robustness against forgetting and contextual interference, we construct Type 3 (Post-Intervention) tasks by positioning tasks at the conclusion of the global timeline. This checkpoint follows a series of intervening sessions involving unrelated domains. By comparing the performance delta between Type 2 and Type 3, we can effectively measure the agent's resilience over a long horizon. 

% Following task placement, the final interaction sequence $\mathcal{T}_{\text{final}}$ is constructed by re-indexing all timestamps to preserve strict chronological consistency. This controlled orchestration ensures that observed performance variances are attributed to memory depth and retrieval precision rather than stochastic ordering effects.

% \vspace{-0.4mm}

\subsection{Alignment with Real-World Interactions}
\label{sec:noise}
% 加一个真实数据对齐的例子
% While event-driven reconstruction enables the modeling of persona evolution via user feedback and interaction, real-world interactions can be erratic and contain disfluencies. For example, a user message can be vague or interspersed with irrelevant and contradictory stochastic signals within a single session. While prior research~\cite{zhao2025llmsrecognizepreferencesevaluating, jiang2025knowmerespondme} has introduced preference updates across multiple sessions, we aim to explore further whether intra-session conflicts impede an agent’s ability to reason over user preferences.
In event-driven persona state modeling, the inherent noise and disfluencies of real-world interactions present significant hurdles. For example, user utterances can be vague, and preferences are often fragmented across multiple messages. While prior research~\cite{zhao2025llmsrecognizepreferencesevaluating, jiang2025knowmerespondme} has introduced preference updates post multiple sessions, we aim to incorporate a broader spectrum of text variability observed in human prompts for LLM~\cite{tian2025taxonomypromptdefectsllm}, as well as individual idiolects.

\subsubsection{Text Variability Injection.}
In natural interactions, user inputs frequently contain vague expressions, inconsistent preferences, or context switch, as illustrated in Table~\ref{tab:query_taxonomy}. Such phenomena challenge personalized memory agents not only in retrieving relevant information but also in integrating histories and composing accurate persona-state signals. 

However, synthetic text is often standardized and lacks variability, even when using a high temperature for generation~\cite{west2025base}. To bridge this gap, we introduce a controlled noise injection process during dialogue generation. Specifically, we perturb user turns with contextually grounded in-session noise that preserves conversational plausibility. Building upon the original dialogue-generation prompt and extracted preferences $\Phi$, we augment the generation process by providing a noise specification $\mathcal{Z}$ that describes the intended perturbation type (e.g., omitted references, context switch, or inconsistent preferences). 
We also generate examples conditioned on $\mathcal{Z}$, which are incorporated as one-shot in-context demonstrations~\cite{brown2020languagemodelsfewshotlearners} in noise specification to guide the dialogue generation agent.

% However, synthetic text lacks such variability. To bridge this gap, we introduce a controlled noise injection process during dialogue generation. Therefore, we propose to perturb user turns with contextually grounded distractors that preserve conversational plausibility. Building upon the original dialogue-generation prompt and extracted preferences $\Phi$, we augment the generation process by providing a noise specification $\mathcal{Z}$ that describes the intended perturbation type (e.g., omitted reference, context switching, or inconsistent preference). We then generate examples conditioned on $\mathcal{Z}$, which are incorporated as one-shot in-context learning~\cite{brown2020language} to guide the dialogue generation agent.

Formally, similar to Equation~\ref{eq:cm}, let $\hat{e}_m$ denote the current timeline entry in domain $k$, and $\mathcal{G}^{noise}_{u,m-1} = \{\mathcal{C}^{noise}_1, \dots, \mathcal{C}^{noise}_{m-1}\}$ the previously generated noisy dialogues. Let $\mathcal{Z}_m$ denote the noise specification sampled for event $e_m$. The noisy dialogue generation process can be expressed as:
\begin{equation}
\mathcal{C}_m^{\text{noise}} 
\sim 
P_g \big( \cdot \mid \mathcal{D}_u, \Phi_m, \hat{e}_m, \mathcal{G}^{noise}_{u,m-1}, \mathcal{Z}_m \big).
\end{equation}
% where $\mathcal{E}_{\text{demo}}(\mathcal{Z}_m)$ denotes the one-shot dialogue example generated by a strong model $P_g$ (e.g., GPT-5) conditioned on noise description $\mathcal{Z}_m$.

Compared to the Clean setting, the conditioning variable $\mathcal{Z}_m$ steers the user turn construction toward vague or misleading patterns, all while preserving semantic alignment with the underlying persona state.
By substituting the original turns in $\mathcal{E}$ with $\mathcal{C}^{noise}_m$, we derive the perturbed event sequence $\mathcal{E}^{noise}$, and keep the global timeline $\mathcal{T}_{\text{final}}$ unchanged. Under these conditions, the model is expected to maintain a stable persona state $\mathcal{S}^u_t$, perform context-aware retrieval and resist irrelevant turns.

\subsubsection{Linguistic Alignment.}
\label{sec:Style-Aligned}
Real-world user inputs exhibit individual idiolects. In contrast, synthetic conversations often follow standardized templates, which simplifies retrieval and artificially inflates evaluation, especially when the generator and the agent share the same underlying model. To prevent models from exploiting overt lexical cues, we construct user prompts to mimic the authentic linguistic patterns observed in real-world interaction corpora.

Specifically, we randomly sample real-world conversations from WildChat~\cite{zhao2024wildchat1mchatgptinteraction} for each user prompt in the \textit{Clean} setting, and instruct an LLM to align the linguistic style of the synthetic user prompt with the real data, while preserving the underlying user intent. In addition, we interleave WildChat data as irrelevant distractors within the global timeline, extending the total context up to 128k tokens. This design challenges models to infer persona states from diverse, individualized linguistic patterns across long-context sessions, thereby increasing the tasks' complexity.

Together, text variability and linguistic alignment ensure that our evaluation extends beyond idealized preference recall. Instead, \method probes the model’s capacity to maintain coherent persona-state tracking even in the presence of pervasive ambiguity and authentic linguistic variability.

\subsection{Evaluation Protocol}
\label{sec:method:eval}
To systematically evaluate persona state tracking across temporal depths and contextual complexities, \method adopts a multi-dimensional evaluation protocol. This protocol integrates temporally controlled task insertion to evaluate cross-session evolution, positional probing within long contexts to assess retrieval robustness, and comprehensive metrics that capture both factual correctness and persona consistency.

\subsubsection{Evaluation Checkpoints (Type 1/2/3).}
Checkpoints are strategically inserted into the global timeline to probe distinct stages of persona-state development. Type~1 (Zero-Memory) tasks are placed before any domain-relevant events occur, serving as a zero-memory baseline to disentangle the agent’s parametric bias from its persona-based responses. Type~2 (In-Time) tasks are introduced immediately following the emergence of all relevant domains to assess peak recall. Finally, Type~3 (Post-Intervention) tasks are placed at the end of the global timeline, after multiple intervening sessions. At this stage, the persona state is fully developed but heavily occluded within a substantially longer and noisier history.

\subsubsection{Contextual Position Probing.}
In addition to temporal stage control, we probe evaluation tasks at varying positions within contexts (e.g., 10\%--100\% event segments of the histories). By measuring performance as a function of this relative distance, we analyze how different personalized memory agents balance retrieval efficiency and contextual scanning. This positional probing provides a fine-grained view of how persona state utilization degrades or stabilizes as context length increases.

\subsubsection{Metrics and Assessment Dimensions.}
Evaluation is conducted along three primary dimensions: task performance, memory fidelity and efficiency. 

\paragraph{Task Performance} We evaluate system performance in both one-shot and interactive settings. In the former, we employ Multiple-Choice Question (MCQ) to measure selection accuracy (\texttt{MCQ} \texttt{Acc.}). The MCQ options are generated by ablating three core aspects: (i) Task Completion, indicating the fulfillment of defined goals; (ii) Preference Consistency, ensuring responses are grounded in long-term preferences without hallucinating unsupported inferences; and (iii) Informational Confidence, identifying whether the model maintains a decisive stance without uncertainty (see Figure~\ref{fig:option} for details). By enumerating all binary combinations of three aspects, we yield eight distinct evaluation categories. 

To better approximate real-world usage, we introduce an interactive evaluation using an LLM-based user simulator (see Figure~\ref{fig:interative} for the prompt). The simulator is granted access to the ground-truth dialogue history and relevant preference annotations. It terminates the session by outputting ``TERMINATE'' only when the task is completed while satisfying all user preferences. Otherwise, it provides supplementary feedback, and the dialogue continues. 

We report the success rate at \texttt{Turn=1} and \texttt{Turn$\leq$2}. Compared to the one-shot setting, the interactive evaluation not only assesses open-text responses but also distinguishes suboptimal memory (resolvable with a single feedback round) from low-quality memory (requiring extensive clarification). Interactions are capped at a maximum of 10 turns. We report task \texttt{Completion} representing the percentage of tasks successfully terminated within this interaction limit, serving as a measure of overall system reliability.

\paragraph{Memory Fidelity} We calculate the \texttt{BERT-f1} between the retrieved context and ground-truth dialogues, complemented by a \texttt{Memory} \texttt{Score} (see Figure~\ref{fig:memory_evaluation2}) derived via ``LLM-as-a-judge''. This protocol evaluates task coverage, accuracy, and preference consistency. Specifically, this scoring mechanism penalizes the omission of critical persona facets while rewarding concise, task-relevant retrievals that enhance downstream utility.

\paragraph{Efficiency} We calculate \texttt{Context} \texttt{Token} to measure the average number of memory tokens consumed per query. To quantify memory retrieval efficiency within long-context environments, we also record the \texttt{Search} \texttt{Duration}.  In the interactive setting, \texttt{User} \texttt{Token} tracks the total tokens generated by the user simulator, capturing the communicative overhead and degree of preference clarification required to resolve a task.

% By separating memory fidelity from task-level correctness and combining static multiple-choice assessment with interactive dialogue evaluation, our framework comprehensively assesses whether a model can maintain, retrieve, and operationalize a stable persona state across evolving conversational contexts.

% Integrating contextual information is often more challenging than filtering irrelevant content. To further evaluate an agent’s ability to track user preferences, we design tasks that require composite reasoning through the synthesis of information spanning multiple user traits and, in some cases, disparate domains. Following PersonaLens~\cite{zhao2025personalensbenchmarkpersonalizationevaluation}, we reuse its task descriptions to reconstruct evaluation queries, while explicitly avoiding the inclusion of overt preference statements or preference-eliciting cues in the queries, ensuring that agents cannot rely on explicit signals and must instead ground their responses in stored memories.

%% file: content/04_experiments.tex
\subsection{Experiment Setup}

\subsubsection{Datasets and Evaluation Metrics.}
\label{dataset}
We utilize the datasets detailed in Section~\ref{sec:event} \&~\ref{sec:noise}, which incorporate event-driven preferences, text variability, and linguistic alignment. The evaluation setup is described in Section~\ref{sec:method:eval}. 

\subsubsection{Personalized Memory Approaches.}
We evaluated a range of personalized memory approaches on \method, categorized into two paradigms: 1) standalone LLMs, 2) agents with RAG and memory systems.
\paragraph{Standalone LLMs.}
In this setting, the full dialogue history is provided to the model as input without any retrieval mechanism or memory compression. We evaluate both reasoning models and chat models, investigating their capabilities in synthesizing preferences across multiple domains and fragmented sessions.

\begin{itemize}[leftmargin=*, topsep=0pt, itemsep=0pt]
\item \textbf{MiniMax-M2.5}\footnote{https://www.minimax.io/news/minimax-m25}: A large-scale reasoning model specializing in zero-shot reasoning over extensive contexts.
\item \textbf{GLM-5/GLM-4.7-Flash}~\cite{glm5team2026glm5vibecodingagentic}: GLM-5 is a reasoning model tested by consuming the complete interaction history, and GLM-4.7-Flash provides fast inference while maintaining solid reasoning ability. 
\item \textbf{Kimi-K2.5}~\cite{kimiteam2026kimik25visualagentic}: A reasoning model designed for large input windows, evaluated without external retrieval.
\end{itemize}
\begin{itemize}[leftmargin=*, topsep=0pt, itemsep=0pt]
\item \textbf{Qwen3-32B/Qwen2.5-72B/Qwen2.5-14B-1M}~\cite{yang2025qwen3technicalreport}: Reasoning and chat models from the Qwen family, offering strong general-purpose reasoning and long-context comprehension.
\item \textbf{Llama3.3-70B}~\cite{grattafiori2024llama3herdmodels}: 70B model in the Llama3 known for reasoning and instruction-following capabilities.
\item \textbf{Gemini2.5-Flash}~\cite{comanici2025gemini25pushingfrontier}: An instruct model designed to deliver strong generation performance with low latency.
\item \textbf{GPT-4o-mini}~\cite{openai2024gpt4ocard}: A variant of the GPT-4 tailored for efficient conversational reasoning and general tasks.
\end{itemize}

\paragraph{Agents with RAG and Memory Systems.}
To evaluate retrieval-augmented approaches, we implement several RAG and structured memory systems with GPT-4o-mini as the backbone response generator.

\begin{itemize}[leftmargin=*, topsep=0pt, itemsep=0pt]

\item \textbf{RAG}~\cite{gao2024retrievalaugmentedgenerationlargelanguage}: A standard retrieval-augmented generation baseline using BGE-M3~\cite{bge_m3} as the dense retriever to retrieve top-$k$ relevant dialogue chunks, which are then concatenated as evidence for generation.
\item \textbf{MemOS}~\cite{li2025memosmemoryosai} introduces a memory operating system for LLMs that manages heterogeneous memory types through MemCube and provides lifecycle scheduling, storage, and retrieval mechanisms.
\item \textbf{Mem0}~\cite{chhikara2025mem0buildingproductionreadyai} introduces a scalable memory layer that automatically extracts, compresses, and persistently stores salient user information, enabling dynamic long-term personalization across sessions.
\item \textbf{LightMem}~\cite{fang2025lightmemlightweightefficientmemoryaugmented} proposes a lightweight memory-augmented framework inspired by cognitive memory models, organizing information into short- and long-term memory stages with efficient compression.
\item \textbf{Memobase}\footnote{https://www.memobase.io/} provides a persistent memory system that organizes user preferences across sessions.
\item \textbf{EverMemOS}~\cite{hu2026EverMemOSselforganizingmemoryoperating} proposes a long-term conversational memory architecture that organizes interaction histories into episodic traces and semantic memory structures, enabling reconstructive retrieval for long-horizon reasoning.
\item \textbf{Supermemory}\footnote{https://supermemory.ai/research} uses hierarchical structured memory for retrieval and long-term persona modeling.
\end{itemize}
% \item \textbf{Supermemory}~\cite{supermemory2026} uses a hierarchical memory to organize interactions with temporal and relational metadata, enabling selective retrieval, conflict resolution, and robust long-term persona modeling beyond traditional RAG.

% All memory systems are evaluated under the same event-driven dialogue setting and use identical task queries to ensure fair comparison in memory fidelity, task completion, and efficiency metrics.

\subsubsection{Implementation Details.}
For standalone LLMs, we invoke their APIs. For memory systems, we evaluate both cloud-based and locally deployed frameworks. MemOS (v1), Mem0 (v2), Memobase (v1), and Supermemory (v4) are accessed via online cloud services for \textsc{Add} and \textsc{Search} operations. EverMemOS and Lightmem are deployed locally using MongoDB~\cite{tao2024pastpostevaluatingquery} for information storage and Milvus~\cite{10.1145/3448016.3457550} as the vector database. For RAG and memory systems, we perform semantic retrieval by fetching the top-$10$ chunks as the default setting, where each chunk in RAG is defined as a single user-assistant pair for embedding. 

\subsubsection{Human Validation.}
During event construction, rigorous human validation is performed to ensure that all user preferences are explicitly or implicitly manifested within the dialogue histories (see Section~\ref{sec:event} and Figure~\ref{fig:pipeline}). This process guarantees the foundational integrity of the dataset. Furthermore, we conduct a human evaluation of the MCQ options to verify their quality. Given the ground-truth dialogue history, human experts achieve an impressive 97.75\% accuracy on the tasks, demonstrating high alignment between the task objectives and the provided dialogue data.

Beyond the MCQ tasks, human experts evaluate additional content generation aspects using the criteria protocol detailed in Figure~\ref{apx:candidate_response2}. Specifically, we assess whether the constructed events accurately cover the goals and descriptions specified in the timeline entries (as described in Section~\ref{sec:method:phase1}), yielding a near-perfect score of 1.99/2. Regarding the injection of text variability, experts find that 98\% of the generations successfully adhere to the designated noise types, confirming the efficacy of our text variability injection process.

\subsection{Overall Performance}

The empirical performance of all evaluated models across single- and multi-domain tasks, under both \textit{Clean} and \textit{Noise} scenarios, is detailed in Tables \ref{tab:clean} through \ref{tab:performance_v4}. We analyze the persona state across temporal depth and compare different models and memory systems from multiple perspectives, including MCQ task accuracy (\texttt{MCQ} \texttt{Acc.}), memory fidelity (\texttt{BERT-f1}, \texttt{Memory} \texttt{Score}), Search efficiency (\texttt{Search} \texttt{Tokens} and \texttt{Duration}), and interactive success rates (\texttt{Completion}, \texttt{User} \texttt{Token} usage, \texttt{Turn = 1}, \texttt{Turn$\le$2}).

\begin{table*}[t]
\centering
\caption{Performance comparison in \textbf{Clean} setting. We group metrics into four aspects. \textbf{Bold} and \underline{underlined} values denote the best and second-best results within each category.}
\label{tab:clean}
\resizebox{\textwidth}{!}{
\begin{tabular}{lccccccccc}
\toprule
\multirow{2}{*}{\centering\textbf{Baseline}} &
\multirow{2}{*}{\centering\textbf{MCQ Acc.}} &
\multicolumn{2}{c}{\textbf{Memory Fidelity}} &
\multicolumn{2}{c}{\textbf{Search Efficiency}} &
\multicolumn{4}{c}{\textbf{Interactive Evaluation}} \\
\cmidrule(lr){3-4} \cmidrule(lr){5-6} \cmidrule(lr){7-10}
& &
\textbf{BERT-f1} &
\textbf{Memory Score} &
\textbf{Context Token $\downarrow$} &
\textbf{Duration (ms) $\downarrow$} &
\textbf{Completion} &
\textbf{User Token $\downarrow$} &
\textbf{Turn=1} &
\textbf{Turn$\leq$2} \\
\midrule

\rowcolor{gray!15}
\multicolumn{10}{c}{\textbf{Standalone LLM (Reasoning model)}} \\
MiniMax-M2.5 & 0.797 & - & - & 34078.6 & - & - & - & - & - \\
GLM-5 & 0.811 & - & - & 34078.6 & - & - & - & - & - \\
Kimi-K2.5 & \textbf{0.882} & - & - & 34078.6 & - & - & - & - & - \\

\rowcolor{gray!15}
\multicolumn{10}{c}{\textbf{Standalone LLM (Chat model)}} \\
Qwen3-32B & \underline{0.870} & - & - & 34078.6 & - & - & - & - & - \\
Qwen2.5-72B & 0.790 & - & - & 34078.6 & - & - & - & - & - \\
Qwen2.5-14B-1M & 0.759 & - & - & 34078.6 & - & - & - & - & - \\
Llama3.3-70B & 0.818 & - & - & 34078.6 & - & - & - & - & - \\
Gemini2.5-Flash & \underline{0.870} & - & - & 34078.6 & - & - & - & - & - \\
GLM-4.7-Flash & 0.868 & - & - & 34078.6 & - & - & - & - & - \\
GPT-4o-mini & 0.780 & - & - & 34078.6 & - & - & - & - & - \\

\rowcolor{gray!15}
\multicolumn{10}{c}{\textbf{RAG \& Memory System based on GPT-4o-mini}} \\
RAG (BGE-M3) & 0.702 & \textbf{0.859} & 1.89 & 928.8 & \underline{16.2} & 0.830 & 61.9 & 0.461 & 0.797 \\
MemOS & \textbf{0.811} & \underline{0.830} & \textbf{2.27} & 709.1 & 369.1 & \underline{0.842} & 60.7 & \textbf{0.548} & 0.801 \\
Mem0 & 0.686 & 0.781 & 1.91 & 340.1 & 557.0 & 0.797 & 69.4 & 0.475 & 0.775 \\
Lightmem & 0.657 & 0.792 & 1.83 & \underline{297.3} & \textbf{8.5} & 0.794 & 62.3 & \underline{0.532} & \underline{0.813} \\
Memobase & 0.733 & 0.781 & 1.86 & 1033.3 & 1991.0 & 0.804 & \textbf{59.2} & 0.504 & \textbf{0.830} \\
EverMemOS & \underline{0.728} & 0.827 & \underline{2.08} & 3230.5 & 16666.5 & \textbf{0.846} & \underline{60.0} & 0.508 & 0.790 \\
Supermemory & 0.655 & 0.799 & 1.84 & \textbf{94.3} & 2881.7 & 0.804 & 65.9 & 0.501 & 0.804 \\

\bottomrule
\end{tabular}}
\end{table*}

\subsubsection{Performance on Clean Interactions}
Table~\ref{tab:clean} and Figure~\ref{fig:type23_clen} present a comprehensive comparison of standalone LLMs and personalized memory agents under the \textit{Clean} setting. From these results, we derive the following observations:

\begin{figure*}[!t]
    \centering
    % 左侧子图
    \begin{subfigure}[b]{0.49\textwidth}
        \centering
        \includegraphics[width=\textwidth]{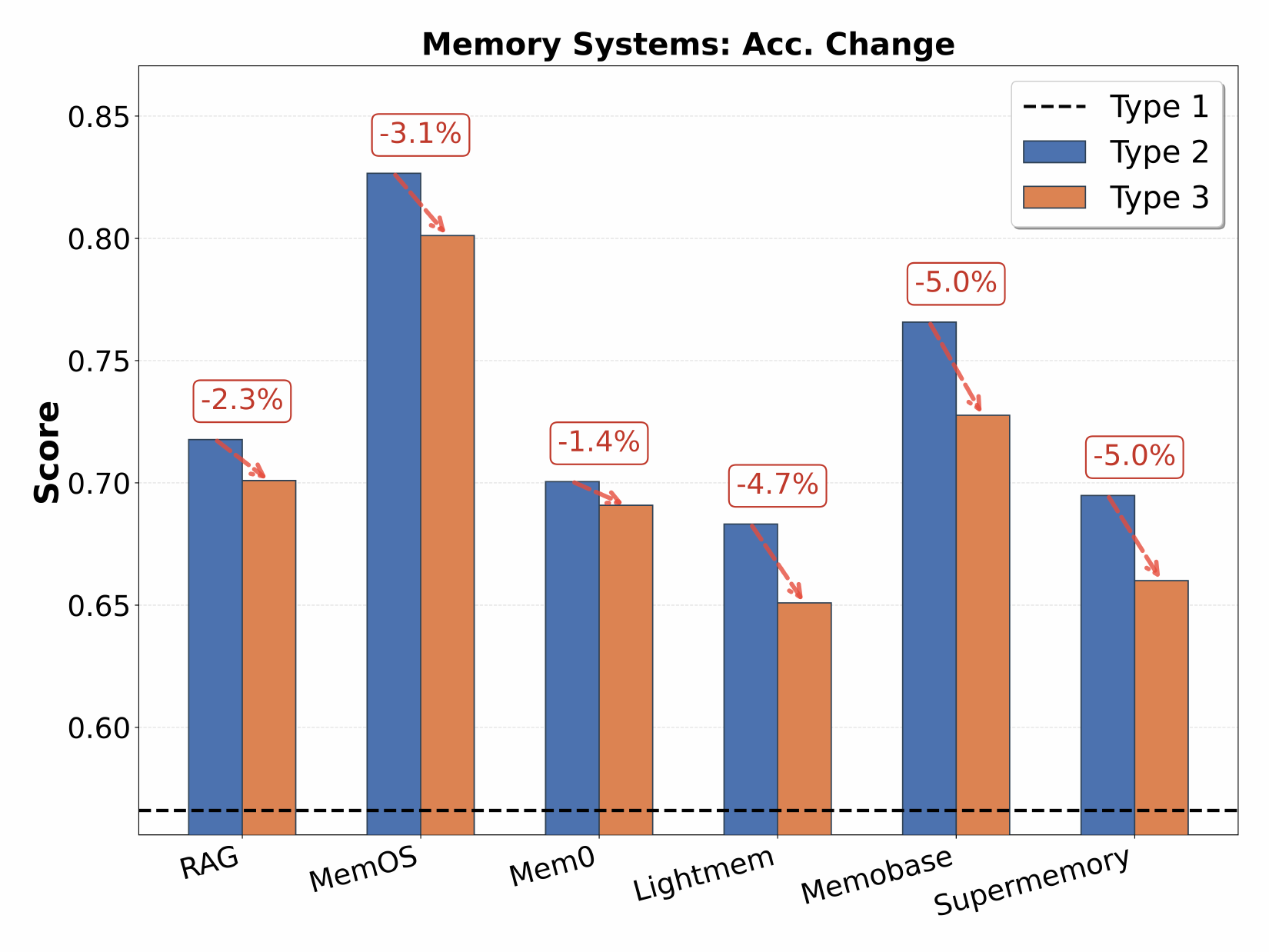}
        % 如果需要给左图加小标题 (a)，可以取消下一行的注释
        % \caption{MCQ Acc. Change}
    \end{subfigure}
    \hfill % 在两个子图之间插入水平填充，使排版更美观
    % 右侧子图
    \begin{subfigure}[b]{0.49\textwidth}
        \centering
        \includegraphics[width=\textwidth]{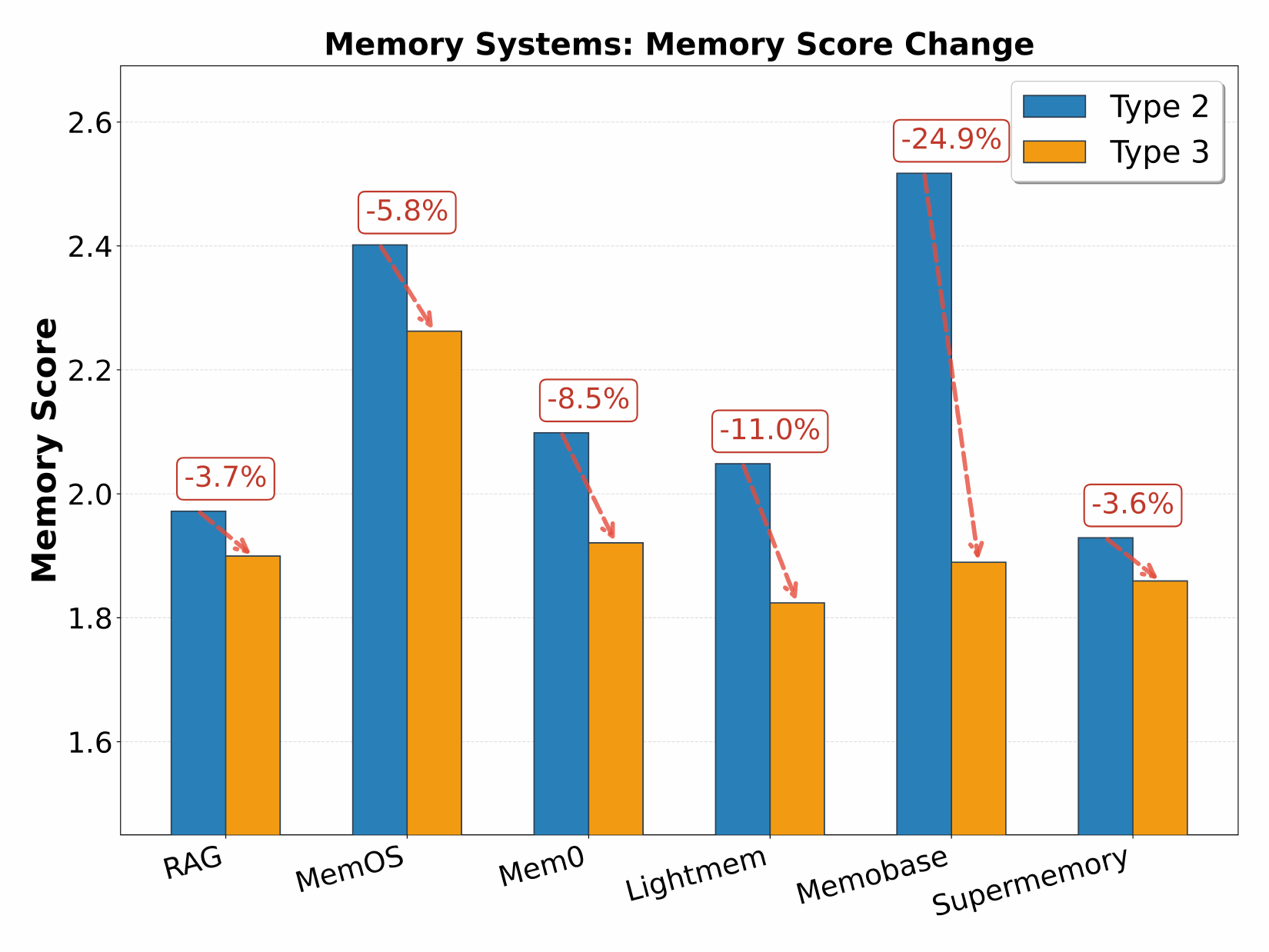}
        % 如果需要给右图加小标题 (b)，可以取消下一行的注释
        % \caption{Memory Score Change}
    \end{subfigure}
    
    % \vspace{-3mm}
    \caption{Performance comparison of memory systems across evaluation checkpoints in the \textbf{Clean} setting (Single). (\textbf{Left}) \texttt{MCQ} \texttt{Acc.} across three checkpoint types. (\textbf{Right}) \texttt{Memory} \texttt{Score} across the checkpoint types. From Type 1 to Type 3, temporal depth and cross-domain interference of the dialogue history increase (see Definition~\ref{def:type} for checkpoint type specification).}
    \label{fig:type23_clen}
\end{figure*}

\begin{enumerate}[leftmargin=*, topsep=0pt, itemsep=0pt]
    \item \textbf{Reasoning Models vs. Chat Models.} Reasoning models demonstrate a clear advantage in maintaining persona state consistency. Specifically, Kimi-K2.5 achieves the highest \texttt{MCQ} \texttt{Acc.} (\textbf{0.882}), followed by Qwen3-32B (\textbf{0.870}) and Gemini2.5-Flash (\textbf{0.870}). It suggests that thinking capabilities are essential for synthesizing user preferences and episodic information from interaction dialogue histories.
    
    \item \textbf{Performance and Efficiency of Memory Systems.}
    Memory systems present a compelling alternative for resource-constrained applications. Notably, Supermemory and Lightmem compress the context tokens to fewer than 300, representing a reduction of over 99\% raw content. However, we find these two systems sacrifice task fulfillment for context compression, underperforming even compared to vanilla RAG. Among the evaluated methods, MemOS achieves an effective balance of the performance-efficiency trade-off, obtaining the highest \texttt{MCQ} \texttt{Acc.} (\textbf{0.811}) and \texttt{Memory} \texttt{Score} (\textbf{2.27}) while consuming fewer \texttt{Context} \texttt{Token} (\textbf{709.1}) than vanilla RAG (\textbf{828.8}).

    Additionally, EverMemOS attains a high \texttt{Completion} rate (\textbf{0.846}), it incurs substantial search latency when deployed locally (\textbf{16s}), which limits its applicability under time-sensitive scenarios. In contrast, locally deployed vanilla RAG spends only \textbf{16ms} on retrieval and obtains competitive success rates (\textbf{0.830}).

    Furthermore, the interactive evaluation—captured by \texttt{Turn=1} and \texttt{Turn$\le$2} metrics—reveals distinct behavioral patterns among the memory systems. MemOS achieves the highest \texttt{Turn=1} success rate (\textbf{0.548}), indicating that its retrieved persona state more effectively enables agents to provide preference-satisfied responses in a one-shot manner. In contrast, while Memobase and Lightmem lag in immediate resolution, they lead in \texttt{Turn$\le$2} performance (Memobase: \textbf{0.830} and Lightmem: \textbf{0.813}). This suggests that while the responses provided by them are initially suboptimal, their persona states possess strong recovery capabilities, allowing agents to reach the desired response with just a single round of user feedback. Conversely, Mem0 exhibits the lowest \texttt{Turn$\le$2} success rate, implying that its persona state often requires extensive user involvement and multiple clarification rounds to resolve the task. 

    \item \textbf{RAG vs.\ Memory Systems.}
    RAG achieves the highest \texttt{BERT-f1} score (\textbf{0.849}), indicating that most key information can be successfully retrieved via dense semantic matching. However, because these retrieved dialogue fragments are often disjointed and unorganized, the downstream model struggles to reconstruct a coherent persona state. In contrast, advanced memory systems (e.g., MemOS) effectively link related information and integrate temporal structures into a unified persona. This structured representation substantially alleviates the cognitive and synthesis burden on the response-generating LLM. Consequently, advanced memory systems significantly outperform RAG, as evidenced by superior \texttt{MCQ Acc.} and \texttt{Completion} rates.
    
    \item \textbf{Performance across Different Temporal Depths.} Figure~\ref{fig:type23_clen} demonstrates the robustness of persona tracking by comparing performance across three types of evaluation checkpoints (Types 1–3). We utilize \texttt{Type} \texttt{1} (Zero-Memory) accuracy as a non-personalized baseline, representing the model's zero-shot capability. Across all evaluated systems, accuracy consistently declines from \texttt{Type} \texttt{2} (In-Time) to \texttt{Type} \texttt{3} (Post-Intervention), suggesting that increased dialogue depth amplifies memory interference and undermines coherent persona maintenance. \texttt{Memory} \texttt{Scores} for Memobase and LightMem exhibit a pronounced drop in the \texttt{Type} \texttt{3}, indicating a strong recency bias where late-stage inputs break long-term persona maintenance. In contrast, MemOS achieves good and stable performance throughout the interaction, demonstrating architectural robustness against a long interaction history.
\end{enumerate}

\subsubsection{Comparative Analysis: Clean vs. Noisy Settings}

\begin{figure*}[!t]
    \centering
    \includegraphics[width=\textwidth]{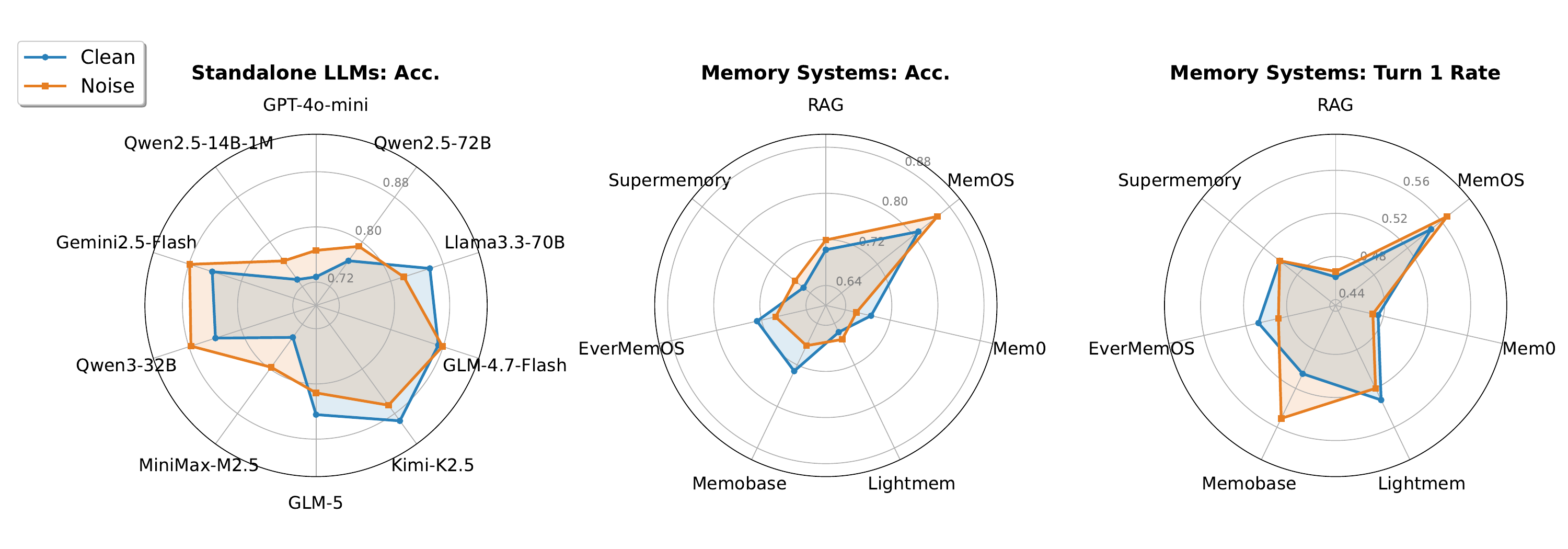}
    % \vspace{-3.3mm}
    \caption{Comprehensive comparison of model and memory system performance across Clean and Noise single-domain scenarios: (\textbf{Left}) \texttt{MCQ} \texttt{Acc.} of standalone LLMs, (\textbf{Center}) \texttt{MCQ} \texttt{Acc.} of memory systems based on GPT-4o-mini, (\textbf{Right}) \texttt{Turn=1} Success Rate of memory systems.}
    \label{fig:Comparison_clean_and_noise}
\end{figure*}

To assess the robustness of different architectures under realistic interaction conditions, we further extend our evaluation to a \textit{Noisy} setting. Specifically, we inject text variability as in-session noise while maintaining the original user intent (see Section~\ref{sec:noise}). By jointly analyzing the results reported in Table~\ref{tab:clean} and Table~\ref{tab:noise}, together with the visualization in Figure~\ref{fig:Comparison_clean_and_noise}, we identify several critical findings.

\begin{table*}[t]
\centering
\caption{Performance comparison in \textbf{Noise} setting. We evaluate baselines from multiple perspectives, consistent with Clean setting. \textbf{Bold} and \underline{underlined} values denote the best and second-best results within each category, respectively.}
\label{tab:noise}
\resizebox{\textwidth}{!}{
\begin{tabular}{lccccccccc}
\toprule
\multirow{2}{*}{\centering\textbf{Baseline}} &
\multirow{2}{*}{\centering\textbf{MCQ Acc.}} &
\multicolumn{2}{c}{\textbf{Memory Fidelity}} &
\multicolumn{2}{c}{\textbf{Search Efficiency}} &
\multicolumn{4}{c}{\textbf{Interactive Evaluation}} \\
\cmidrule(lr){3-4} \cmidrule(lr){5-6} \cmidrule(lr){7-10}
& &
\textbf{BERT-f1} &
\textbf{Memory Score} &
\textbf{Context Token $\downarrow$} &
\textbf{Duration (ms) $\downarrow$} &
\textbf{Completion} &
\textbf{User Token $\downarrow$} &
\textbf{Turn=1} &
\textbf{Turn$\leq$2} \\
\midrule

\rowcolor{gray!15}
\multicolumn{10}{c}{\textbf{Standalone LLM (Reasoning model)}} \\
MiniMax-M2.5 & 0.797 & - & - & 35585.9 & - & - & - & - & - \\
GLM-5 & 0.813 & - & - & 35585.9 & - & - & - & - & - \\
Kimi-K2.5 & 0.865 & - & - & 35585.9 & - & - & - & - & - \\

\rowcolor{gray!15}
\multicolumn{10}{c}{\textbf{Standalone LLM (Chat model)}} \\
Qwen3-32B & \underline{0.877} & - & - & 35585.9 & - & - & - & - & - \\
Qwen2.5-72B & 0.792 & - & - & 35585.9 & - & - & - & - & - \\
Qwen2.5-14B-1M & 0.766 & - & - & 35585.9 & - & - & - & - & - \\
Llama3.3-70B & 0.820 & - & - & 35585.9 & - & - & - & - & - \\
GLM-4.7-Flash & 0.853 & - & - & 35585.9 & - & - & - & - & - \\
Gemini2.5-Flash & \textbf{0.879} & - & - & 35585.9 & - & - & - & - & - \\
GPT-4o-mini & 0.766 & - & - & 35585.9 & - & - & - & - & - \\

\rowcolor{gray!15}
\multicolumn{10}{c}{\textbf{RAG \& Memory System based on GPT-4o-mini}} \\
RAG (BGE-M3) & \underline{0.719} & \textbf{0.852} & 1.92 & 933.4 & \underline{16.9} & 0.811 & 60.9 & 0.466 & 0.787 \\
MemOS & \textbf{0.853} & \underline{0.844} & \textbf{2.38} & 1486.7 & 644.5 & \textbf{0.837} & \textbf{56.9} & \textbf{0.567} & \textbf{0.837} \\
Mem0 & 0.660 & 0.779 & 1.87 & 337.1 & 492.6 & 0.818 & 68.7 & 0.470 & 0.754 \\
Lightmem & 0.671 & 0.791 & 1.88 & \underline{292.9} & \textbf{8.0} & \underline{0.820} & 61.4 & 0.520 & 0.806 \\
Memobase & 0.683 & 0.772 & 1.87 & 1061.0 & 1721.5 & 0.785 & 61.2 & \underline{0.551} & 0.787 \\
EverMemOS & 0.695 & 0.824 & \underline{2.09} & 3177.8 & 19246.9 & \underline{0.811} & \underline{60.4} & 0.489 & 0.773 \\
Supermemory & 0.674 & 0.796 & 1.96 & \textbf{92.6} & 3883.6 & 0.806 & 62.0 & 0.501 & 0.811 \\

\bottomrule
\end{tabular}}
\end{table*}

\begin{enumerate}[leftmargin=*, topsep=0pt, itemsep=0pt] 
\item \textbf{Robustness of Standalone LLMs against Text Variability.} Many standalone LLMs maintain comparable performance when in-session noise is introduced. As shown in Figure~\ref{fig:Comparison_clean_and_noise} (1), Gemini 2.5-Flash demonstrates superior handling of noisy contexts, maintaining high accuracy (\textbf{0.879}). In contrast, while Kimi-K2.5 excels in \textit{Clean} scenarios (\textbf{0.882}), its performance drops to \textbf{0.865}. It is thus more susceptible to such distraction compared to Gemini.

\item \textbf{Noise as a Semantic Catalyst for Memory Systems.} Memory systems exhibit improved performance in the noisy setting. Further investigation shows that in-session noise acts as an internal conflict that emphasizes preference information, thereby facilitating more effective extraction. This is consistent with the findings reported in PrefEval~\cite{zhao2025llmsrecognizepreferencesevaluating}. An example is MemOS, whose retrieval volume nearly doubles from \textbf{709.1} tokens (Clean) to \textbf{1486.7} tokens (Noise). This expanded context provides a more detailed description of user preferences, leading to an increase in \texttt{Memory} \texttt{Score} (\textbf{2.27$\rightarrow$2.38}) and \texttt{MCQ} \texttt{Acc.} (\textbf{0.811$\rightarrow$0.853}).

% \item \textbf{Efficiency constraints and retrieval sensitivity.} The trade-off between efficiency and performance becomes more pronounced under noise, revealing how sensitive different retrieval architectures are to context complexity. As shown in Figure~\ref{fig:Comparison_clean_and_noise} (2), MemOS and Supermemory demonstrate superior noise processing during retrieval, which is reflected in both their efficiency and performance gains. Conversely, Memobase and EverMemOS exhibit degradation in the noisy setting, indicating that their indexing mechanisms struggle to distinguish core persona data from event noise. The remaining methods exhibit consistent search durations and retrieval context lengths irrespective of variations in context, indicating that these retrieval and rerankers are relatively invariant to the semantic content of the input.
\end{enumerate}

\subsubsection{Comparative Analysis: Single Domain vs. Multi-Domain.}
In this section, we extend our evaluation to the multi-domain queries to examine the agents' capability in synthesizing information cross-domains (Table~\ref{tab:performance_multi_clean}). The transition from single to multi-domain query reveals a significant performance gap. Our findings are:

\begin{table*}[t]
\centering
\caption{Performance comparison in the \textbf{Clean} setting (\textbf{Multi-domain}). We use multi-domain user queries to evaluate baselines across three paradigms, consistent with the Clean setting.}
\label{tab:performance_multi_clean}
\resizebox{\textwidth}{!}{
\begin{tabular}{lccccccccc}
\toprule
\multirow{2}{*}{\centering\textbf{Baseline}} &
\multirow{2}{*}{\centering\textbf{MCQ Acc.}} &
\multicolumn{2}{c}{\textbf{Memory Fidelity}} &
\multicolumn{2}{c}{\textbf{Search Efficiency}} &
\multicolumn{4}{c}{\textbf{Interactive Evaluation}} \\
\cmidrule(lr){3-4} \cmidrule(lr){5-6} \cmidrule(lr){7-10}
& &
\textbf{BERT-f1} &
\textbf{Memory Score} &
\textbf{Context Token $\downarrow$} &
\textbf{Duration (ms) $\downarrow$} &
\textbf{Completion} &
\textbf{User Token $\downarrow$} &
\textbf{Turn=1} &
\textbf{Turn$\leq$2} \\
\midrule

\rowcolor{gray!15}
\multicolumn{10}{c}{\textbf{Standalone LLM (Reasoning model)}} \\
MiniMax-M2.5 & 0.860 & - & - & 34078.6 & - & - & - & - & - \\
GLM-5 & 0.885 & - & - & 34078.6 & - & - & - & - & - \\
Kimi-K2.5 & \textbf{0.955} & - & - & 34078.6 & - & - & - & - & - \\

\rowcolor{gray!15}
\multicolumn{10}{c}{\textbf{Standalone LLM (Chat model)}} \\
Qwen3-32B & \underline{0.936} & - & - & 34078.6 & - & - & - & - & - \\
Qwen2.5-72B & 0.815 & - & - & 34078.6 & - & - & - & - & - \\
Qwen2.5-14B-1M & 0.873 & - & - & 34078.6 & - & - & - & - & - \\
Llama3.3-70B & 0.682 & - & - & 34078.6 & - & - & - & - & - \\
GLM-4.7-Flash & 0.828 & - & - & 34078.6 & - & - & - & - & - \\
Gemini2.5-Flash & 0.898 & - & - & 34078.6 & - & - & - & - & - \\
GPT-4o-mini & 0.707 & - & - & 34078.6 & - & - & - & - & - \\

\rowcolor{gray!15}
\multicolumn{10}{c}{\textbf{RAG \& Memory System based on GPT-4o-mini}} \\
RAG (BGE-M3) & 0.682 & \textbf{0.849} & 1.78 & 858.1 & \underline{16.5} & \textbf{0.745} & 122.6 & 0.204 & 0.561 \\
MemOS & \textbf{0.732} & 0.819 & \textbf{2.14} & 664.7 & 364.2 & 0.643 & \underline{113.3} & \underline{0.306} & \underline{0.592} \\
Mem0 & 0.650 & 0.785 & 1.78 & 339.5 & 525.3 & \underline{0.694} & 129.7 & 0.280 & 0.529 \\
Lightmem & 0.605 & 0.795 & 1.78 & \underline{289.9} & \textbf{8.5} & 0.643 & 129.2 & 0.274 & 0.580 \\
Memobase & 0.694 & 0.793 & 1.71 & 1033.2 & 2228.0 & 0.650 & \textbf{102.4} & \textbf{0.331} & \textbf{0.656} \\
EverMemOS & \underline{0.713} & \underline{0.820} & \underline{1.98} & 3134.4 & 15847.0 & 0.688 & 115.2 & 0.268 & 0.573 \\
Supermemory & 0.656 & 0.803 & 1.72 & \textbf{92.4} & 3232.3 & 0.675 & 125.4 & 0.248 & 0.554 \\

\bottomrule
\end{tabular}}
\end{table*}

\begin{figure*}[t]
    \centering
    % 第一个图（左）：单栏设置
    \begin{subfigure}[b]{0.49\textwidth}
        \centering
        \includegraphics[width=\textwidth]{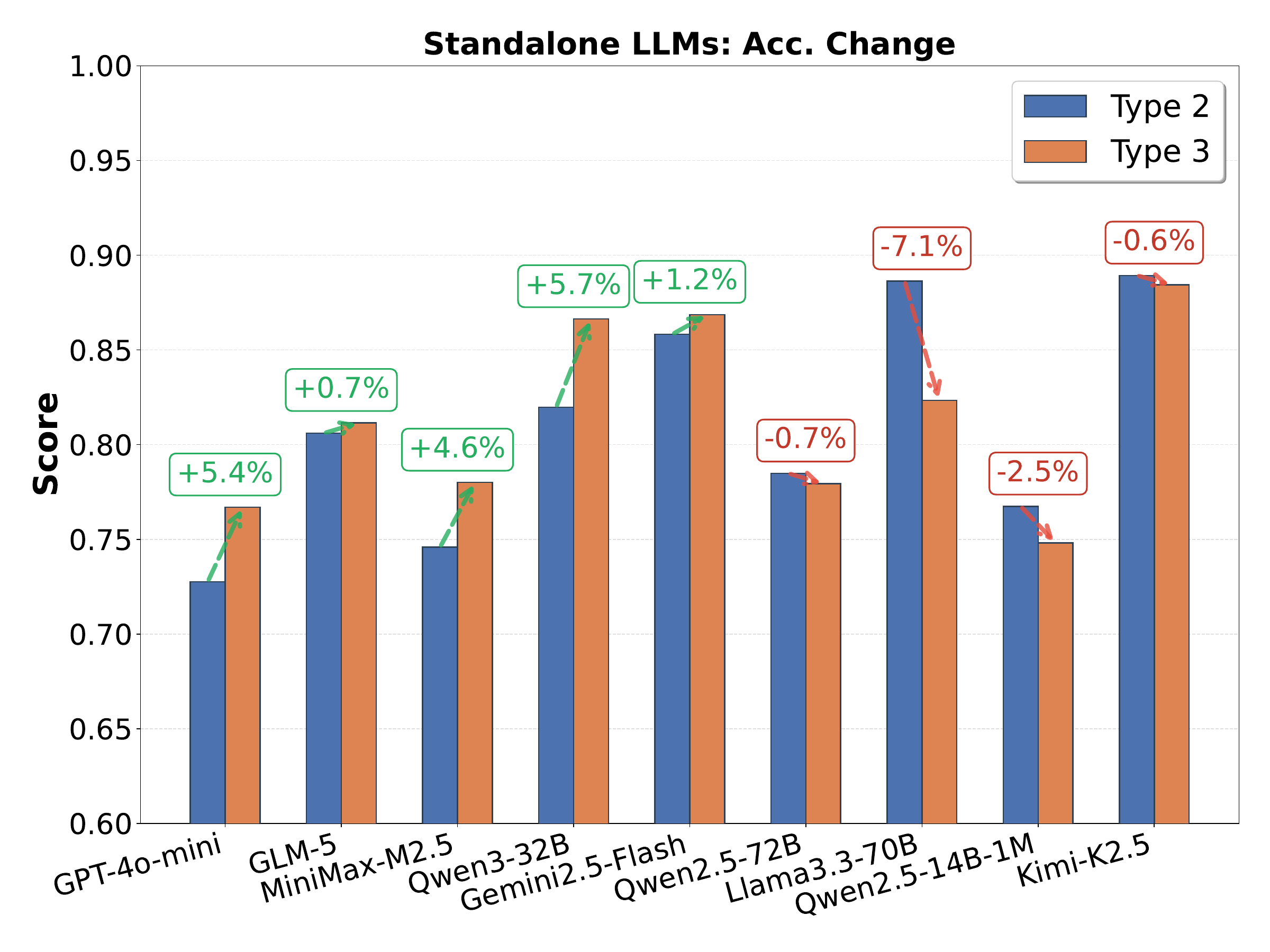}
    \end{subfigure}
    \hfill % 水平填充，使左右子图分别向两端对齐
    % 第一个图（右）：多栏设置
    \begin{subfigure}[b]{0.49\textwidth}
        \centering
        \includegraphics[width=\textwidth]{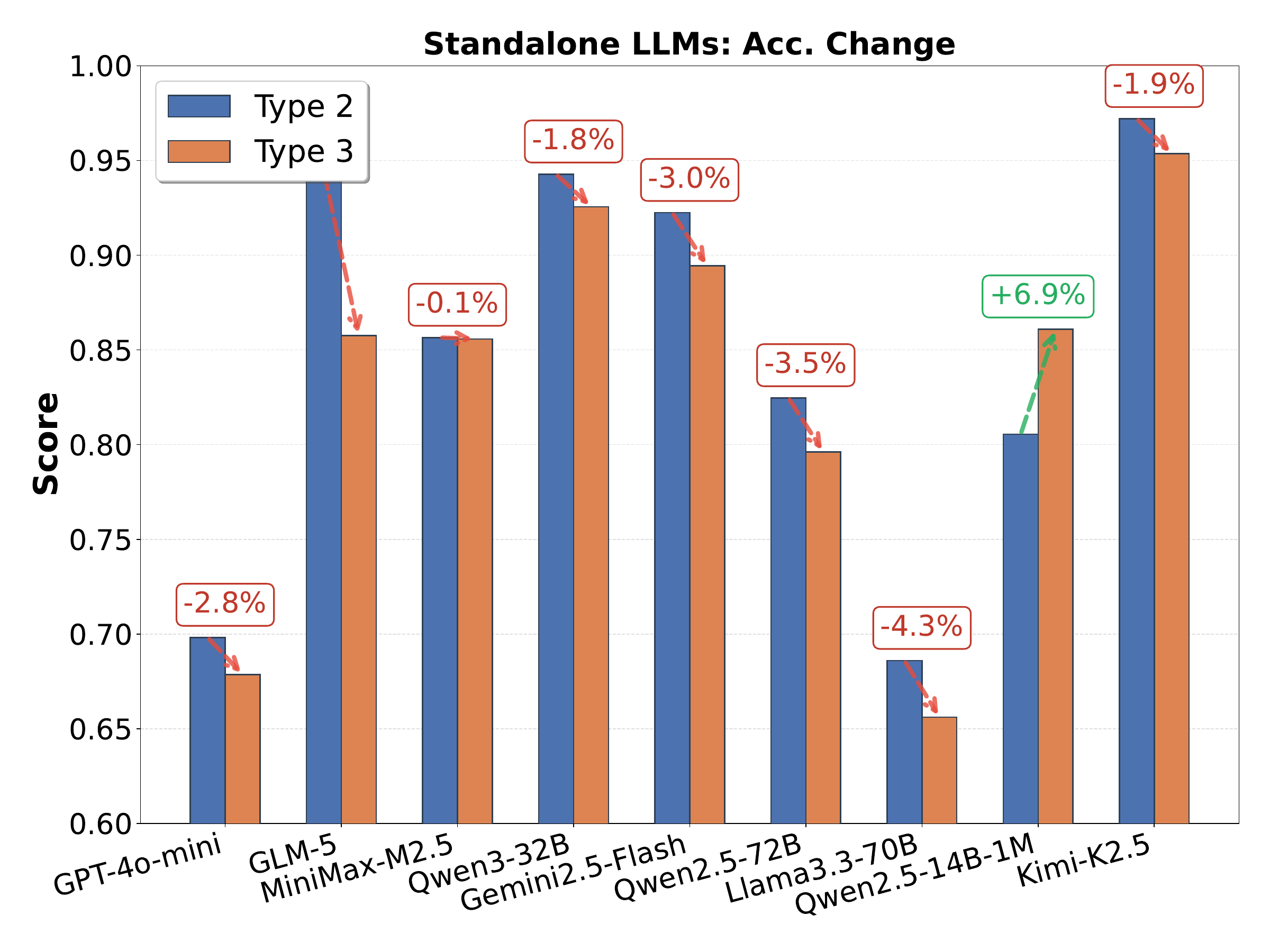}
    \end{subfigure}
    % \vspace{-3mm}
    \caption{\texttt{MCQ} \texttt{Acc.} of standalone LLMs at different evaluation checkpoints. Results are categorized by single-domain (\textbf{Left}) and (\textbf{Right}) multi-domain settings.}
    \label{fig:longcontextacc}
\end{figure*} 

\begin{figure*}[t]
    \centering
    % 第二个图（左）：多域准确率
    \begin{subfigure}[b]{0.49\textwidth}
        \centering
        \includegraphics[width=\textwidth]{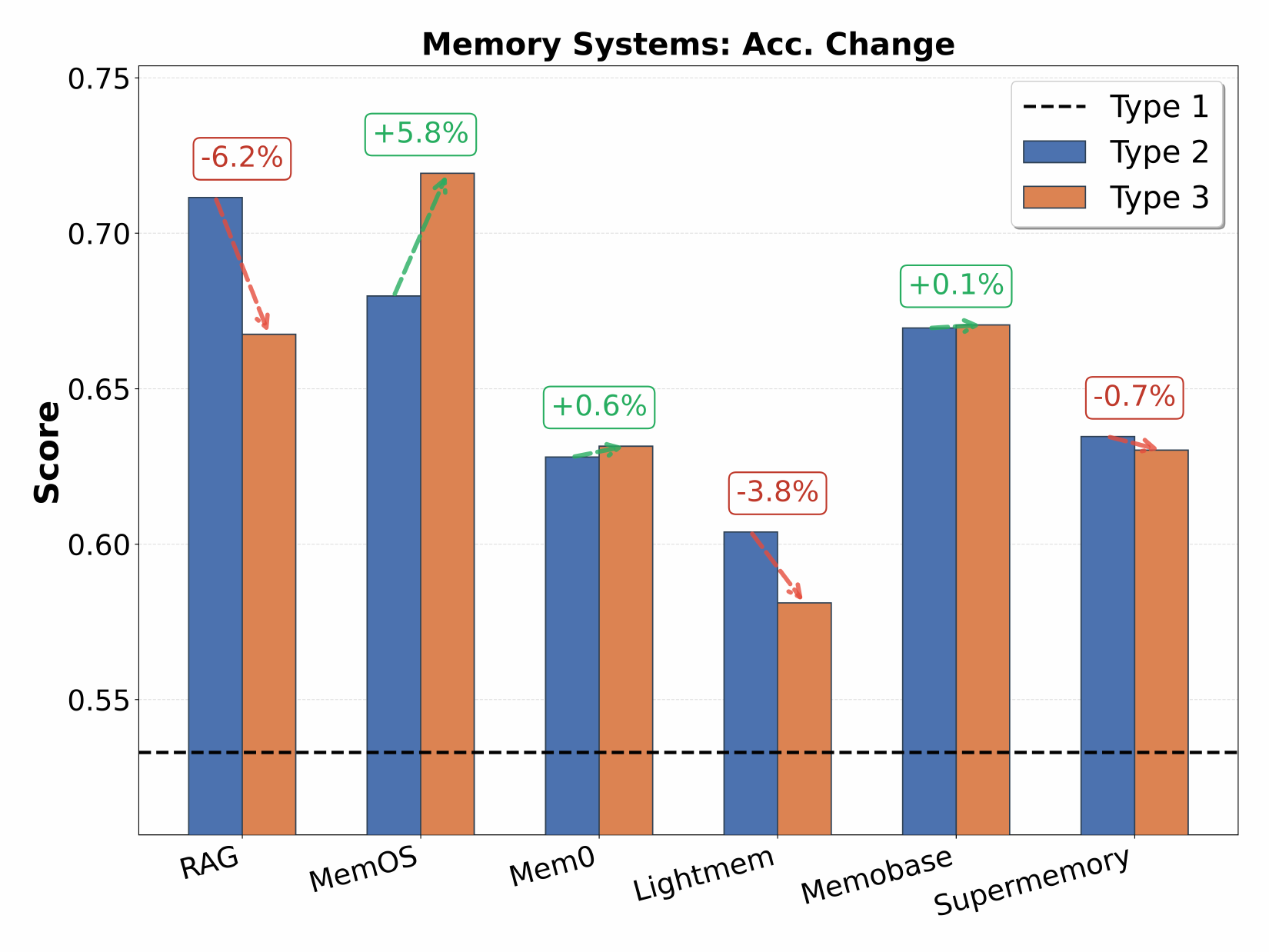}
    \end{subfigure}
    \hfill
    % 第二个图（右）：多域记忆得分
    \begin{subfigure}[b]{0.49\textwidth}
        \centering
        \includegraphics[width=\textwidth]{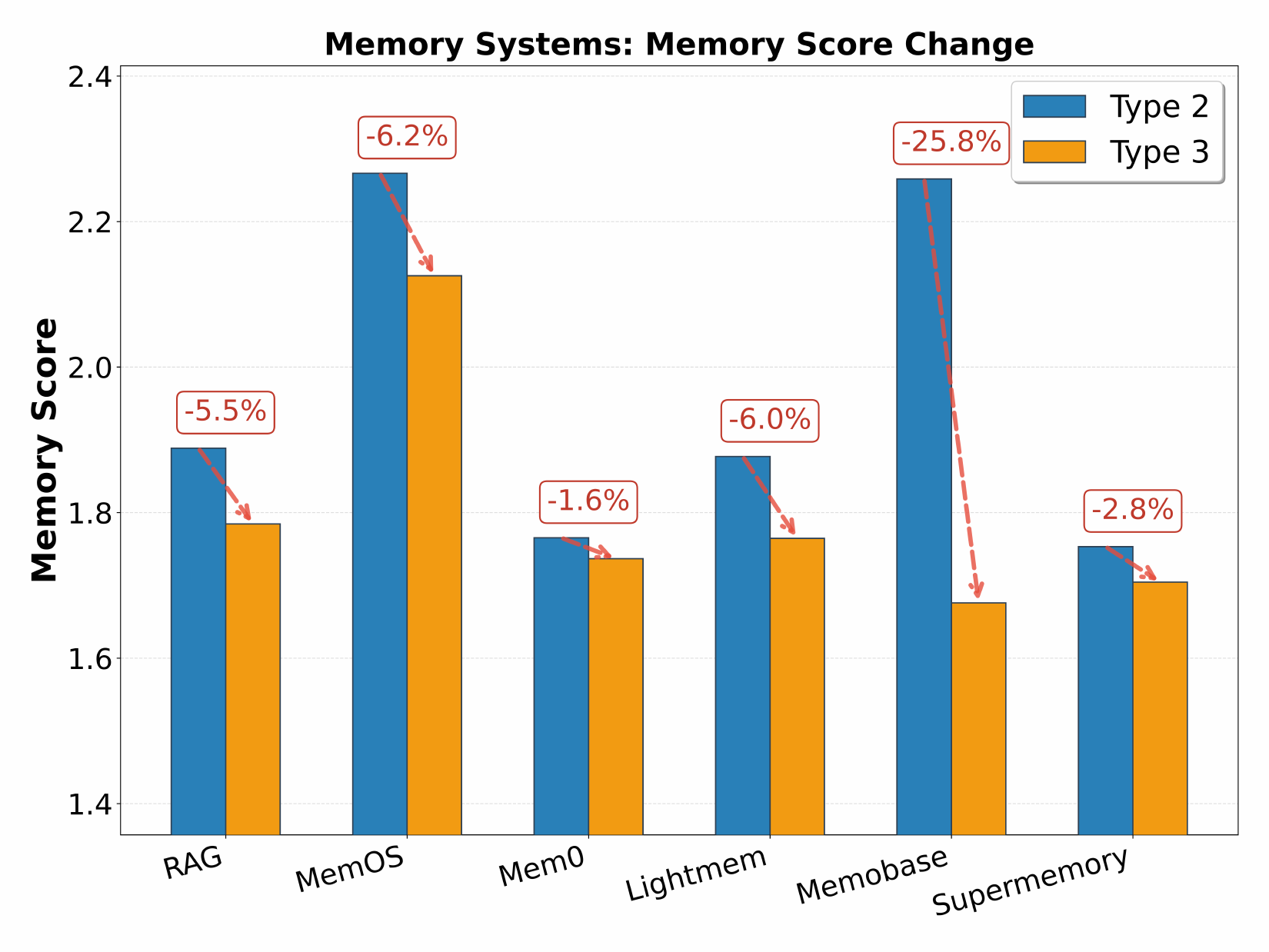}
    \end{subfigure}
    % \vspace{-2.1mm}
    \caption{(\textbf{Left}) Accuracy (\texttt{MCQ} \texttt{Acc.}) across checkpoints in \textbf{Clean} setting (Multi), where the dashed line represents the baseline performance under information omission (Type 1). (\textbf{Right}) Memory Scores across the event types.}
    \label{fig:mem_multi}
\end{figure*} 

\begin{figure*}[t]
    \centering
    % 第三个图（左）：Top-k 选项得分（适当微调比例以保证水平平齐）
    \begin{subfigure}[b]{0.51\textwidth}
        \centering
        \includegraphics[width=\textwidth]{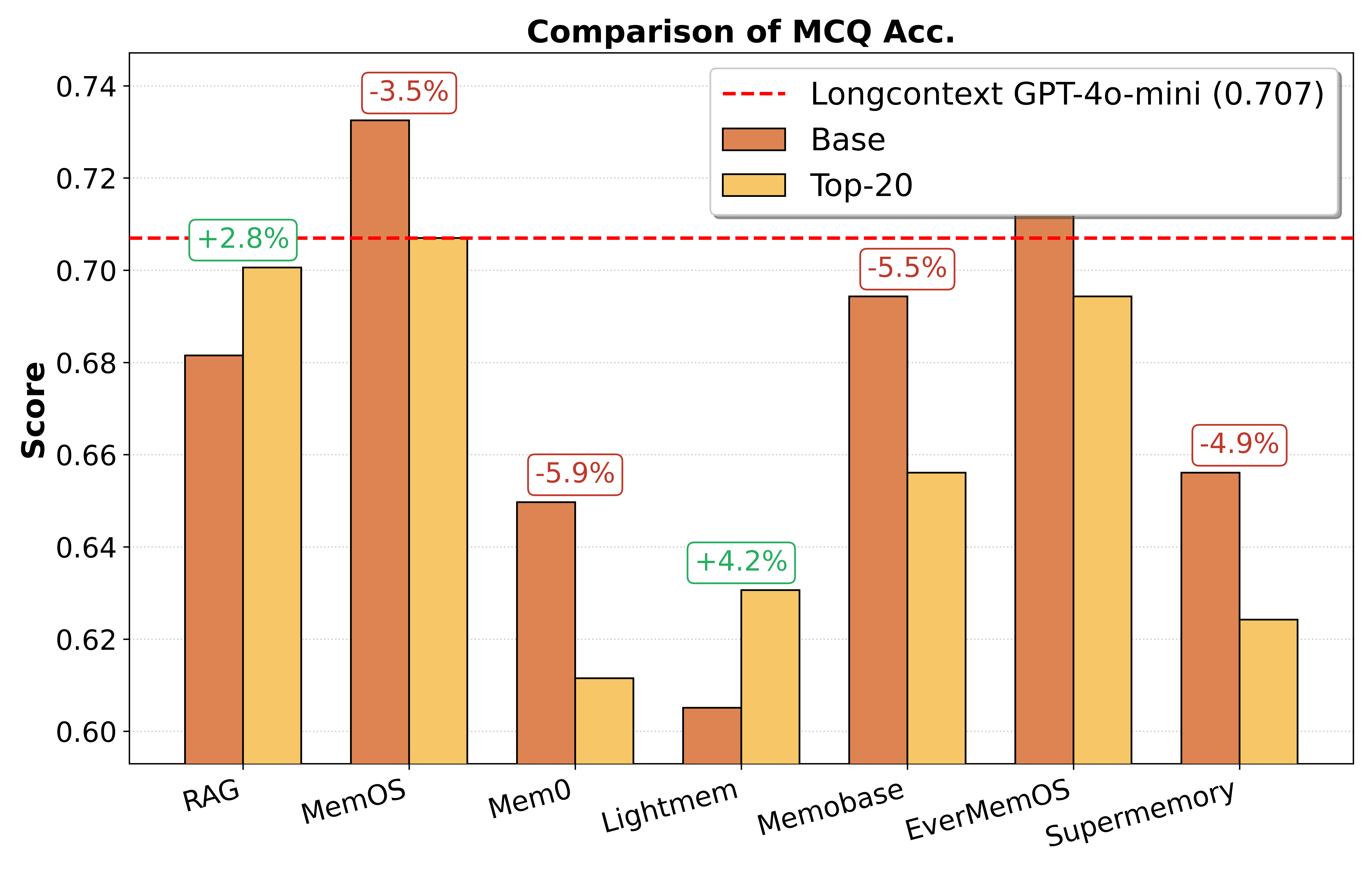}
    \end{subfigure}
    \hfill
    % 第三个图（右）：Turn=1 得分
    \begin{subfigure}[b]{0.45\textwidth}
        \centering
        \includegraphics[width=\textwidth]{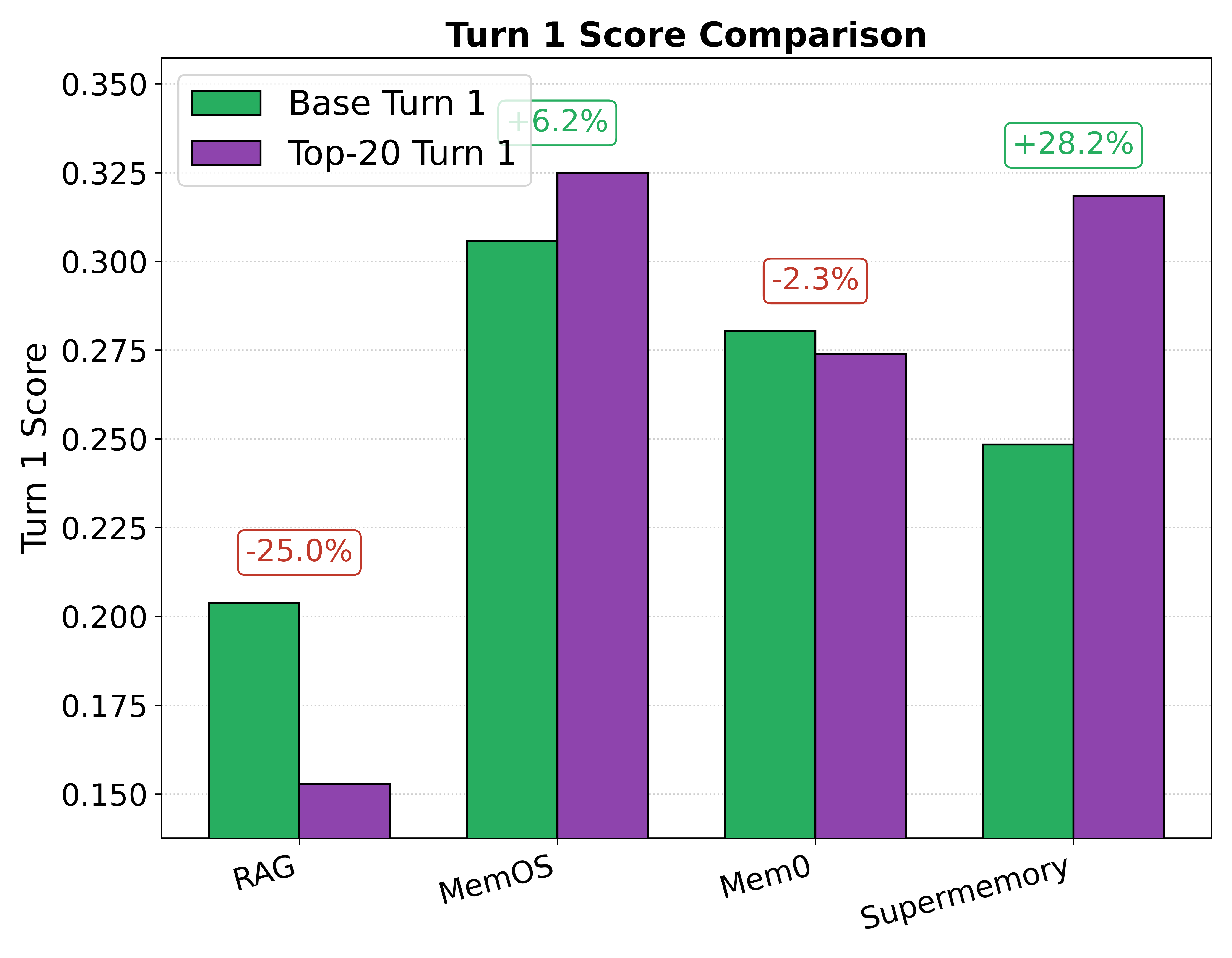}
    \end{subfigure}
    % \vspace{-2.1mm}
    \caption{Analysis of retrieval depth (Top-k) on multi-domain performance. (\textbf{Left}) Comparison of \texttt{MCQ} \texttt{Acc.} between the Top-10 (Base) and Top-20 settings. (\textbf{Right}) \texttt{Turn=1} Score comparison across representative memory systems.}
    \label{fig:top20}
\end{figure*}

\begin{enumerate}[leftmargin=*, topsep=0pt, itemsep=0pt]

\item \textbf{Overall Performance of Multi-Domain Queries.} As shown in Table~\ref{tab:performance_multi_clean}, all memory system-based agents experience a pronounced decline in performance compared to the single-domain tasks. It is particularly evident in interactive metrics, where the success rate at the first attempt (\texttt{Turn=1}) has declined significantly, meanwhile the average \texttt{User} \texttt{Token} count has increased. 
% Interestingly, as cross-domain information is integrated, the overall gap narrows in the multi-domain setting, suggesting structured memory may provide greater stability in complex contexts.

\item \textbf{Cross-Domain Interference and Evaluation Limitations.} As shown in Figure~\ref{fig:longcontextacc} (1), in the single-domain tasks, standalone LLMs may not perform worse at Type 3 checkpoints with longer dialogue histories compared with Type 2 checkpoints. This phenomenon suggests that the presence of irrelevant tokens does not necessarily obstruct the LLM’s ability to retrieve and utilize task-relevant information. Instead, the model appears capable of filtering or tolerating moderate noise when the contextual structure remains simple and domain boundaries are clear.

However, this robustness does not generalize to multi-domain tasks (Figure~\ref{fig:longcontextacc} (2)). As noise increases, performance deteriorates consistently across most LLMs, indicating the difficulty of synthesizing information across domains in a long context. In particular, Llama3.3-70B exhibits a sharp performance drop, reflecting susceptibility to irrelevant information when contexts and dialogues are interleaved.

Beyond this degradation, we observe a clear divergence between option accuracy \texttt{MCQ} \texttt{Acc.} and interactive performance as well as \texttt{Memory} \texttt{Score} in memory systems. While memory system-based agents maintain relatively stable accuracy (Table~\ref{tab:performance_multi_clean}, Figure~\ref{fig:mem_multi} (2)), both their memory scores and response quality decline substantially under the same conditions. This indicates that selecting the correct option from predefined candidates is considerably easier than generating a reasoned answer through multi-step inference. Moreover, since the memory score primarily reflects retrieval precision, successful recall of key preferences can enable downstream reasoning to compensate for irrelevant context, partially hiding degradation in memory quality. These findings highlight a limitation of prior evaluations: option accuracy alone is insufficient to characterize the effectiveness of memory in multi-domain tasks.

\item \textbf{Recall Robustness of Memory Systems vs. RAG.} As shown in Figure~\ref{fig:mem_multi}, the effectiveness of memory systems varies across different evaluation checkpoints. While the introduction of irrelevant information reduces retrieval precision, memory systems maintain high recall of task-relevant preference signals due to their structured persona-state representation. In contrast, RAG relies on semantic similarity-based retrieval, making it susceptible to interference from irrelevant context and leading to a performance decline.

\item \textbf{Retrieval Expansion and Structural Limitations.} To better support multi-domain queries that require broader evidence aggregation, we expand the number of retrieved items from 10 to 20. However, this expansion leads to performance degradation in most memory systems, as illustrated in Figure~\ref{fig:top20}. Since these memory systems rely on compressed persona-state representations, the inclusion of excessive irrelevant information may introduce noise that interferes with downstream reasoning. In contrast, standard RAG benefits significantly from increased retrieval depth. Since RAG retrieves semantic fragments that may lack contextual completeness, retrieving more candidates improves coverage of missing situational details. These findings reveal that current memory systems rely on fixed retrieval memory segments and cannot adapt retrieval precision to varying task difficulty and information demands.

\end{enumerate}

\begin{table*}[t]
\centering
\caption{Performance comparison in the \textbf{Noise} setting (\textbf{Multi-domain}). We use multi-domain user queries to evaluate baselines across three paradigms, consistent with the Noise setting.}
\label{tab:performance_v4}
\resizebox{\textwidth}{!}{
\begin{tabular}{lccccccccc}
\toprule
\multirow{2}{*}{\centering\textbf{Baseline}} &
\multirow{2}{*}{\centering\textbf{MCQ Acc.}} &
\multicolumn{2}{c}{\textbf{Memory Fidelity}} &
\multicolumn{2}{c}{\textbf{Search Efficiency}} &
\multicolumn{4}{c}{\textbf{Interactive Evaluation}} \\
\cmidrule(lr){3-4} \cmidrule(lr){5-6} \cmidrule(lr){7-10}
& &
\textbf{BERT-f1} &
\textbf{Memory Score} &
\textbf{Context Token $\downarrow$} &
\textbf{Duration (ms) $\downarrow$} &
\textbf{Completion} &
\textbf{User Token $\downarrow$} &
\textbf{Turn=1} &
\textbf{Turn$\leq$2} \\
\midrule

\rowcolor{gray!15}
\multicolumn{10}{c}{\textbf{Standalone LLM (Reasoning model)}} \\
MiniMax-M2.5 & 0.866 & - & - & 35585.9 & - & - & - & - & - \\
GLM-5 & \underline{0.905} & - & - & 35585.9 & - & - & - & - & - \\
Kimi-K2.5 & \textbf{0.930} & - & - & 35585.9 & - & - & - & - & - \\

\rowcolor{gray!15}
\multicolumn{10}{c}{\textbf{Standalone LLM (Chat model)}} \\
Qwen3-32B & \textbf{0.930} & - & - & 35585.9 & - & - & - & - & - \\
Qwen2.5-72B & 0.841 & - & - & 35585.9 & - & - & - & - & - \\
Qwen2.5-14B-1M & 0.841 & - & - & 35585.9 & - & - & - & - & - \\
Llama3.3-70B & 0.656 & - & - & 35585.9 & - & - & - & - & - \\
GLM-4.7 Flash & 0.841 & - & - & 35585.9 & - & - & - & - & - \\
Gemini2.5-Flash & \textbf{0.930} & - & - & 35585.9 & - & - & - & - & - \\
GPT-4o-mini & 0.720 & - & - & 35585.9 & - & - & - & - & - \\

\rowcolor{gray!15}
\multicolumn{10}{c}{\textbf{RAG \& Memory System based on GPT-4o-mini}} \\
RAG (BGE-M3) & 0.669 & \textbf{0.845} & 1.85 & 864.8 & \underline{16.8} & \textbf{0.732} & 120.3 & 0.229 & 0.529 \\
MemOS & \textbf{0.752} & 0.816 & \textbf{2.17} & 680.6 & 360.6 & 0.650 & 121.7 & \underline{0.268} & 0.580 \\
Mem0 & 0.662 & 0.785 & 1.77 & 332.4 & 555.9 & 0.650 & 137.3 & 0.210 & 0.522 \\
Lightmem & 0.631 & 0.795 & 1.77 & \underline{287.8} & \textbf{9.1} & 0.656 & \textbf{110.8} & 0.236 & \underline{0.611} \\
Memobase & 0.643 & 0.796 & 1.72 & 1038.0 & 1645.2 & 0.701 & \underline{112.8} & \textbf{0.274} & 0.580 \\
EverMemOS & \underline{0.732} & \underline{0.820} & \underline{1.98} & 3092.9 & 26227.5 & \underline{0.713} & 114.0 & \underline{0.268} & 0.522 \\
Supermemory & 0.637 & 0.803 & 1.75 & \textbf{90.7} & 3606.1 & 0.675 & 122.3 & 0.248 & \textbf{0.612} \\

\bottomrule
\end{tabular}}
\end{table*}

\begin{figure*}[!t]
    \centering
    \includegraphics[width=\textwidth]{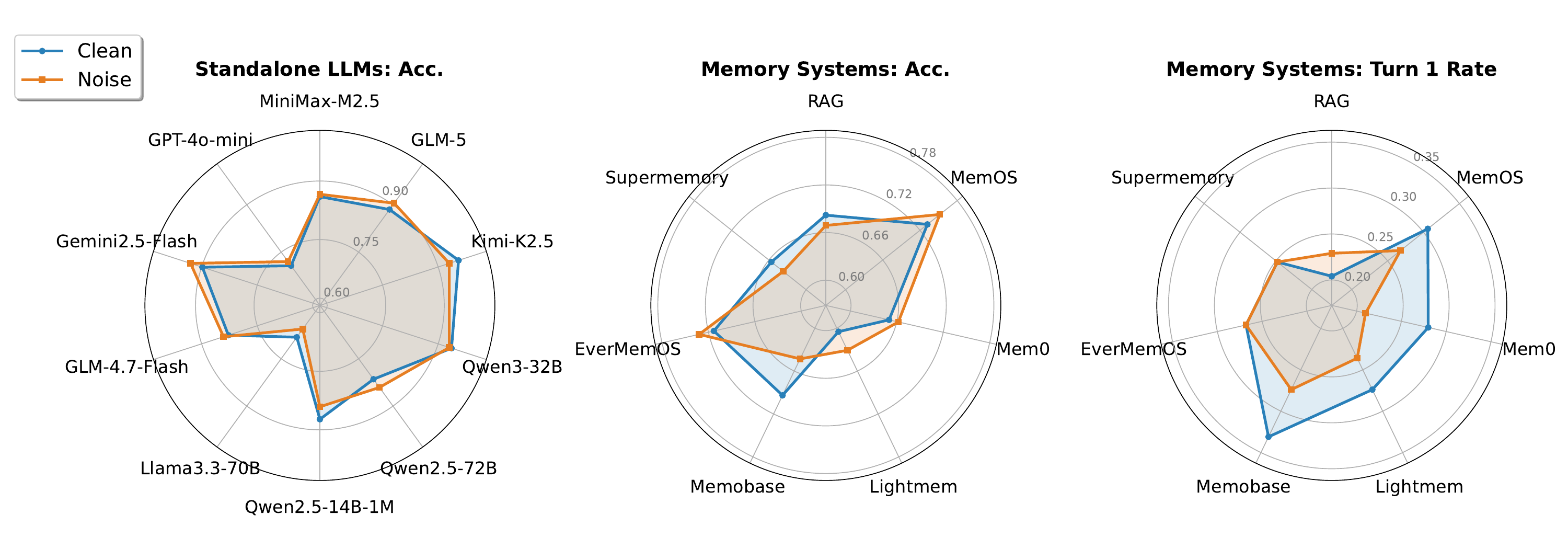}
    \caption{Comprehensive comparison of model performance across Clean and Noise multi-domain scenarios: (\textbf{Left}) \texttt{MCQ} \texttt{Acc.} of standalone LLMs, (\textbf{Center}) \texttt{MCQ} \texttt{Acc.} and (\textbf{Right}) \texttt{Turn=1} Success Rate of memory systems.}
    \label{fig:large_radar_multi}
\end{figure*} 

\begin{figure*}[!t]
    \centering
    \includegraphics[width=\textwidth]{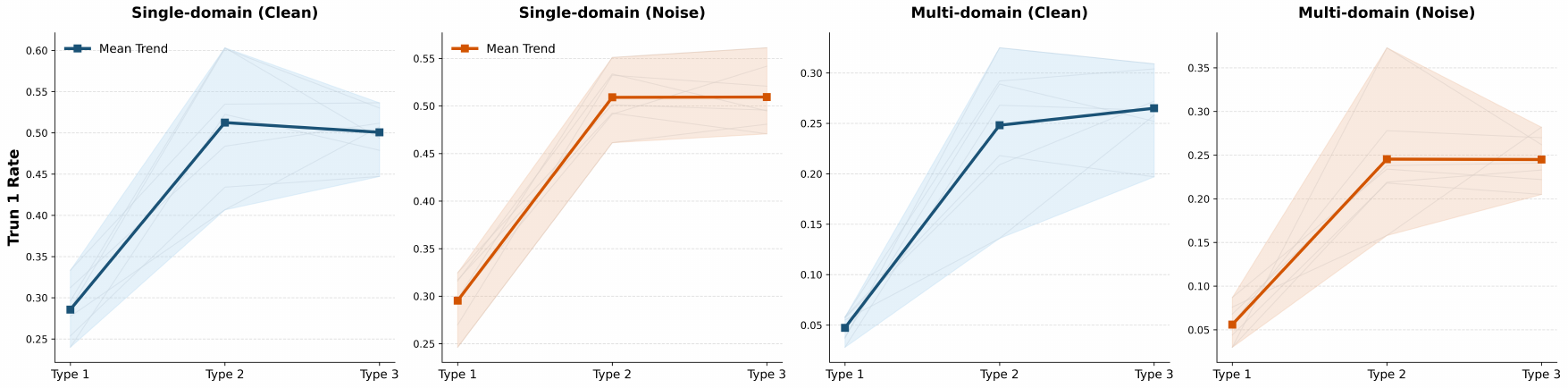}
    \caption{Variation and performance gap of baselines in \texttt{Turn=1} interaction across stages under different settings.}
    \label{fig:type}
\end{figure*} 

\subsubsection{Performance Dynamics in Multi-Domain Noise Scenarios}
In the multi-domain noise environment, the performance of standalone LLMs diverges significantly from memory systems-based agents, revealing critical insights into reasoning capacity and retrieval efficiency. 

\begin{enumerate}[leftmargin=*, topsep=0pt, itemsep=0pt]
\item \textbf{Performance of Standalone LLMs.} As indicated in Table~\ref{tab:performance_v4}, standalone LLMs with enhanced reasoning capability, such as Kimi-K2.5, demonstrate substantial resilience to text variability and the complexity of multi-domain queries, maintaining a robust \texttt{MCQ} \texttt{Acc.} of \textbf{0.930}. It suggests that advanced reasoning is a pivotal driver in stable preference formation, especially navigating environments with high contextual entropy. In contrast, chat models such as Llama3.3-70B suffer a performance degradation, with accuracy falling to \textbf{0.656}. highlighting a significant susceptibility to complex tasks with high contextual entropy.

\item \textbf{Performance Changes in Interactive Scenarios.} 
% The evaluation in Table~\ref{tab:performance_v4} of memory systems highlights the structural advantages of persona state-based persistence over raw context windowing. MemOS, for instance, maintains a high \texttt{Memory} \texttt{Score} of \textbf{2.17} as the complexity of the interaction environment increases. While the success rate for single-turn responses (\texttt{Turn=1}) is relatively low across all memory systems, the significantly higher success rate within two turns (\texttt{Turn$\leq$2}=\textbf{0.580}) indicates a robust capacity for error correction and interactive recovery. 
Figure~\ref{fig:large_radar_multi} shows the performance fluctuations of memory system-based agents compared to standalone LLMs in the presence of noise. Notably, memory systems exhibit higher sensitivity, particularly during interactive tasks where response accuracy drops after noise is introduced. This degradation is more pronounced than that observed in the single-domain setting shown in Figure~\ref{fig:Comparison_clean_and_noise}. These results suggest that while current memory systems decompose queries~\cite{ma2023query} and rerank retrieved memory, they still struggle to integrate and properly order information across domains when dealing with noisy environments.

Finally, taking a holistic view of Figure~\ref{fig:type}, we observe that, when comparing performance between the Type~3 and Type~2 positions, memory systems generally maintain stable \texttt{Turn=1} performance and remain largely robust to token-level noise introduced by irrelevant dialogue. Moreover, relative to the no-memory baseline (Type~1), they achieve consistent and steady improvements. However, compared to the single-domain tasks, the disparity among different memory systems becomes more pronounced in the multi-domain evaluation, particularly at the Type~2 position, suggesting that memory systems differ in their sensitivity to cross-domain complexity.

\item \textbf{Efficiency Changes in Multi-domain Noisy Scenarios.} Efficiency metrics show that the introduction of noise leads to a substantial increase in computational overhead. As shown in Table~\ref{tab:performance_v4}, the \texttt{Search} \texttt{Duration} of EverMemOS rises to \textbf{26}s, significantly higher than in the clean multi-domain setting, suggesting that semantic noise increases the difficulty of retrieval and preference synthesis. In contrast, Lightmem maintains a consistently low latency of \textbf{9.1}ms, indicating better scalability under noisy conditions.

\end{enumerate}

\begin{table}[t]
\centering
\caption{Performance under the style-aligned long-context setting across both single- and multi-domain tasks. For single-domain evaluation, values in parentheses denote results obtained using linguistically aligned queries without incorporating real-data padding into the context. \texttt{Acc.} denotes MCQ accuracy. \texttt{Search} \texttt{Duration} is measured in milliseconds (ms).}
\label{tab:combined_all_turns}
% 增加行高，因为去掉了 booktabs，需要手动撑开一点间距防止文字贴线
\renewcommand{\arraystretch}{1.6} 
\resizebox{\textwidth}{!}{
\begin{tabular}{lccccc|ccccc}
\hline % 替换 \toprule
\multirow{2}{*}{\textbf{Baseline}} & \multicolumn{5}{c|}{\textbf{Single-Domain}} & \multicolumn{5}{c}{\textbf{Multi-Domain}} \\
\cline{2-6} \cline{7-11} % 替换 \cmidrule，原生命令不会打断竖线
& \textbf{Acc.} & \textbf{Context Token \textdownarrow} & \textbf{Search Duration \textdownarrow} & \textbf{Turn=1} & \textbf{Turn$\leq$2} & \textbf{Acc.} & \textbf{Context Token \textdownarrow} & \textbf{Search Duration \textdownarrow} & \textbf{Turn=1} & \textbf{Turn$\leq$2} \\ 
\bottomrule % 替换 \midrule
\rowcolor{gray!15} 

\multicolumn{11}{c}{\textbf{Standalone LLMs}} \\ 
Qwen2.5-1M & 0.716 (0.766) & 116541.5 & - & - & - & 0.726 & 116541.5 & - & - & - \\
GPT-4o-mini & 0.0 (0.775) & 116541.5 & - & - & - & 0.0 & 116541.5 & - & - & - \\
\rowcolor{gray!15} 
\multicolumn{11}{c}{\textbf{RAG \& Memory System based on GPT-4o-mini}} \\ 
RAG (BGE-M3) & 0.697 (0.707) & 962.3 & \underline{17.5} & 0.440 & 0.757 & 0.650 & 857.6 & \underline{20.1} & 0.217 & 0.522 \\
MemOS & \textbf{0.813} (\textbf{0.809}) & 647.7 & 276.3 & \textbf{0.563} & \textbf{0.839} & \textbf{0.764} & 644.1 & 370.0 & \textbf{0.331} & \textbf{0.637} \\
Mem0 & 0.681 (0.669) & 299.4 & 526.2 & 0.492 & 0.790 & 0.637 & 333.3 & 588.2 & 0.210 & 0.465 \\
Lightmem & 0.645 (0.638) & \underline{293.7} & \textbf{8.4} & 0.497 & 0.809 & 0.592 & \underline{291.1} & \textbf{8.2} & 0.255 & \underline{0.580} \\
Memobase & 0.700 (\underline{0.719}) & 1031.0 & 1479.9 & \underline{0.551} & \underline{0.816} & 0.707 & 1034.7 & 1607.2 & \underline{0.274} & 0.573 \\
EverMemOS & \underline{0.740} (0.707) & 3307.2 & 13258.1 & 0.501 & 0.785 & \underline{0.720} & 3185.7 & 10419.0 & \underline{0.274} & 0.561 \\
Supermemory & 0.671 (0.662) & \textbf{119.6} & 3412.8 & 0.527 & 0.801 & 0.586 & \textbf{116.1} & 3812.5 & 0.255 & 0.541 \\
\hline % 替换 \bottomrule
% \multicolumn{11}{p{\textwidth}}{\footnotesize \textit{Note: Except for Qwen2.5-14B-1M, the other models do not support ultra-long context window; therefore, no comparison is provided under this setting.}} \\
\end{tabular}}
\end{table}

\begin{figure*}[!t]
    \centering
    \includegraphics[width=\textwidth]{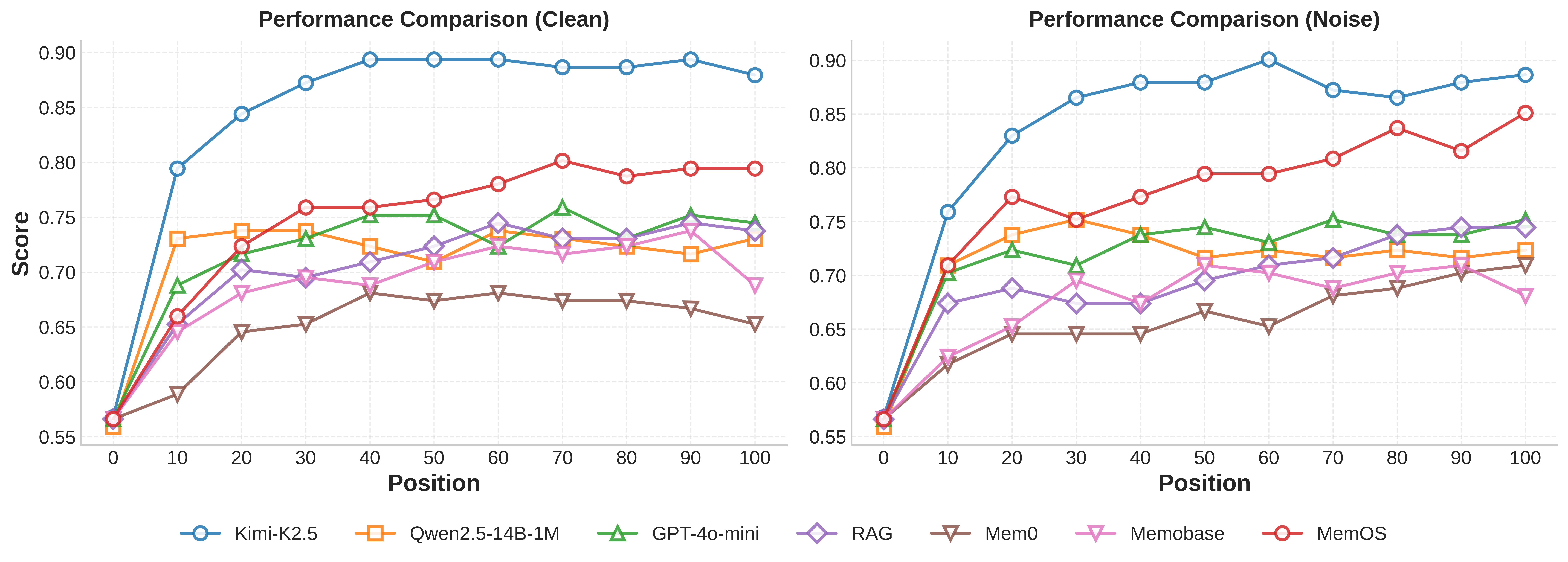}
    % \vspace{-2.5mm}
    \caption{\texttt{MCQ} \texttt{Acc.} performance trends of approaches across different segment positions in single-domain tasks under Clean and Noisy settings.}
    \label{fig:Comparison2}
\end{figure*}

\begin{figure*}[!t]
    \centering
    % 左侧子图：用户热力图
    \begin{subfigure}[b]{0.50\textwidth}
        \centering
        \includegraphics[width=\textwidth]{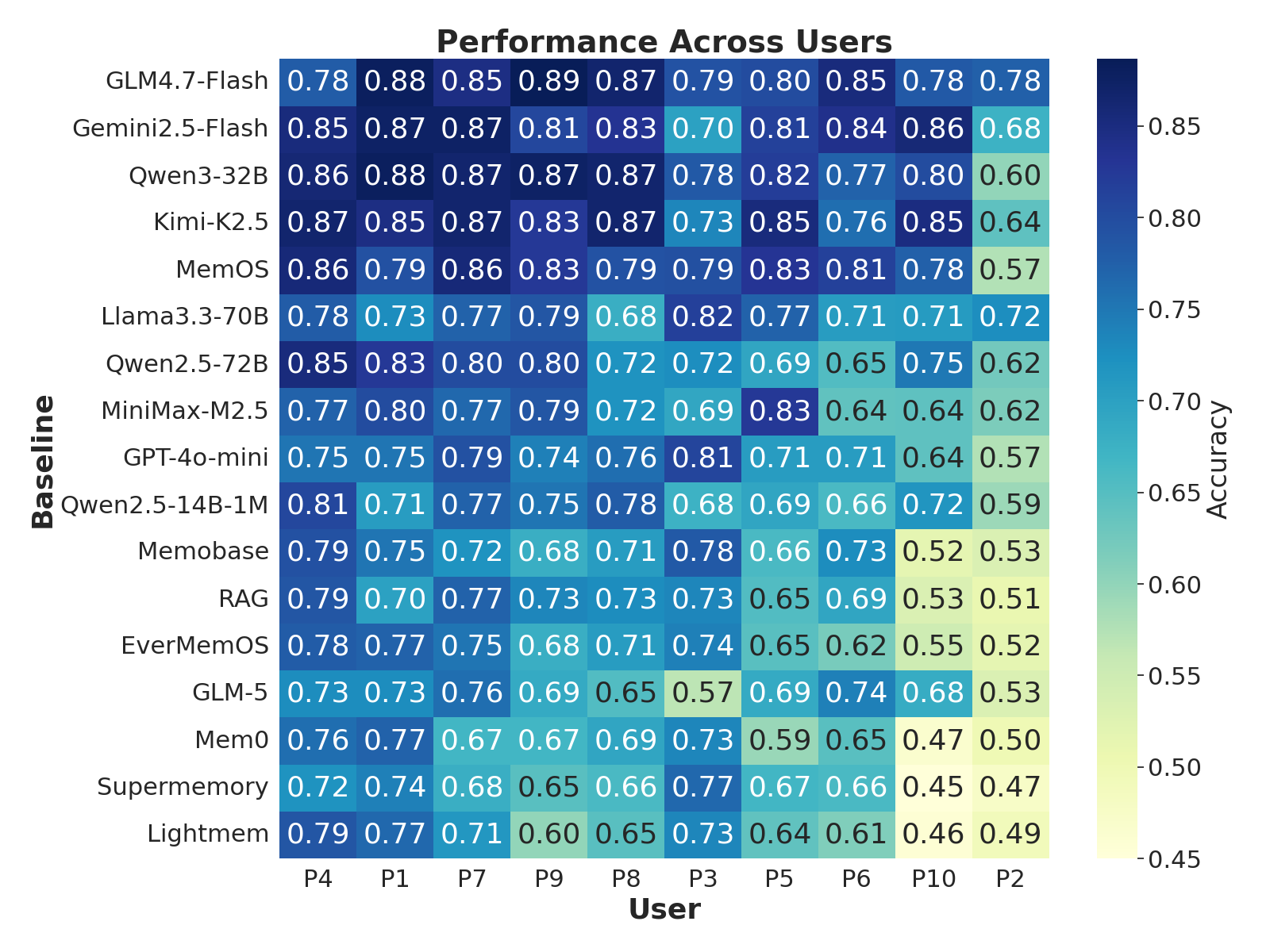}
    \end{subfigure}
    \hfill % 水平弹性间距，使左右子图完美向两端对齐
    % 右侧子图：风格对齐趋势图
    \begin{subfigure}[b]{0.48\textwidth}
        \centering
        \includegraphics[width=\textwidth]{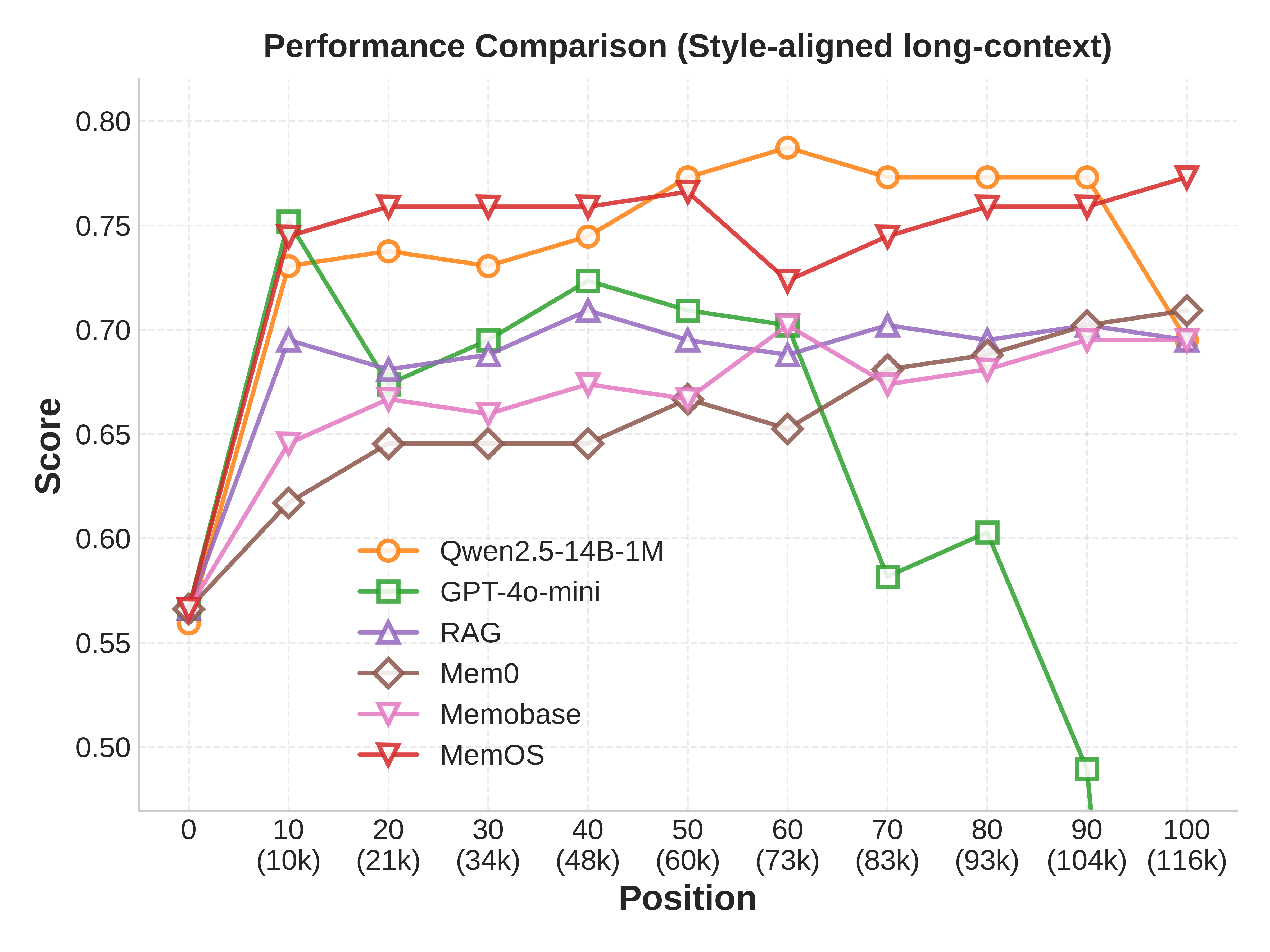}
    \end{subfigure}
    
    % \vspace{-2.5mm}
    \caption{(\textbf{Left}) Heatmap of \texttt{MCQ} \texttt{Acc.} across diverse users, highlighting how persona uniqueness influences system success regardless of context length. (\textbf{Right}) \texttt{MCQ} \texttt{Acc.} trends across positions in style-aligned long context settings.}
    \label{fig:heat}
\end{figure*}

% \begin{wrapfigure}{r}{0.4\textwidth}
%     \centering
%     \includegraphics[width=0.38\textwidth]{Figures/personascope/performance_comparison_(long_&_style-aligned).png} % 图片宽度略小于环境宽度
%     \caption{test}
% \end{wrapfigure}

\begin{table}[t]
\centering
\small
\caption{\texttt{MCQ} \texttt{Acc.} across Positions. \#Ques. denotes the number of questions whose corresponding domain events have already appeared. \textbf{Bold} and \underline{underlined} values denote the best and second-best results within segment position.}
\label{tab:combined_results_refined}
\renewcommand{\arraystretch}{0.97} % 将行间距缩小为原来的 0.8 倍
\resizebox{\textwidth}{!}{
\begin{tabular}{lcccccccccc} % 修正为11列
\toprule
\textbf{Pos.\%} 
& \textbf{Qwen2.5-1M} 
& \textbf{GPT-4o-mini} 
& \textbf{RAG} 
& \textbf{EverMemOS} 
& \textbf{Lightmem} 
& \textbf{Mem0} 
& \textbf{Memobase} % 修正拼写
& \textbf{Supermemory} 
& \textbf{MemOS} 
& \textbf{\#Ques.} \\
\midrule
\rowcolor{gray!15} % 设置本行背景为浅灰色
\multicolumn{11}{c}{\textbf{Performance (Clean)}} \\ % 修正跨列数为11
% \midrule
10  & 0.703 & \textbf{0.757} & 0.676 & \underline{0.730} & 0.676 & 0.595 & \underline{0.730} & 0.649 & 0.676 & 37 \\ % 补充了MemOS的数据占位
20  & \textbf{0.765} & \underline{0.728} & \underline{0.728} & \underline{0.728} & 0.679 & 0.679 & 0.716 & 0.654 & \textbf{0.765} & 81  \\
30  & \underline{0.757} & 0.729 & 0.730 & 0.720 & 0.645 & 0.664 & 0.738 & 0.645 & \textbf{0.785} & 107  \\
40  & 0.727 & \underline{0.758} & 0.727 & 0.695 & 0.617 & 0.695 & 0.695 & 0.680 & \textbf{0.766} & 128  \\
50  & 0.708 & \underline{0.752} & 0.723 & 0.715 & 0.613 & 0.672 & 0.708 & 0.664 & \textbf{0.766} & 137  \\
\midrule
60  & 0.736 & 0.721 & \underline{0.743} & 0.729 & 0.643 & 0.679 & 0.721 & 0.629 & \textbf{0.779} & 141  \\
70  & 0.731 & \underline{0.759} & 0.731 & 0.731 & 0.653 & 0.674 & 0.716 & 0.638 & \textbf{0.801} & 141  \\
80  & 0.723 & \underline{0.731} & \underline{0.731} & \underline{0.731} & 0.645 & 0.674 & 0.723 & 0.660 & \textbf{0.787} & 141  \\
90  & 0.716 & \underline{0.752} & 0.745 & 0.716 & 0.645 & 0.667 & 0.738 & 0.653 & \textbf{0.794} & 141  \\
100 & 0.731 & \underline{0.745} & 0.738 & 0.723 & 0.631 & 0.653 & 0.688 & 0.653 & \textbf{0.794} & 141  \\
% \midrule
\rowcolor{gray!15} % 设置本行背景为浅灰色
\multicolumn{11}{c}{\textbf{Performance (Noise)}} \\
% \midrule
10  & 0.703 & \underline{0.730} & \underline{0.730} & \underline{0.730} & 0.649 & 0.703 & \underline{0.730} & 0.676 & \textbf{0.811} & 37 \\
20  & \underline{0.778} & 0.741 & 0.704 & 0.691 & 0.654 & 0.728 & 0.691 & 0.704 & \textbf{0.803} & 81  \\
30  & \underline{0.776} & 0.729 & 0.701 & 0.710 & 0.654 & 0.710 & 0.738 & 0.636 & \textbf{0.794} & 107  \\
40  & \underline{0.742} & 0.734 & 0.680 & 0.680 & 0.609 & 0.695 & 0.680 & 0.648 & \textbf{0.789} & 128  \\
50  & 0.715 & \underline{0.745} & 0.693 & 0.723 & 0.628 & 0.657 & 0.708 & 0.599 & \textbf{0.796} & 137  \\
\midrule
60  & 0.721 & 0.729 & 0.707 & \underline{0.743} & 0.629 & 0.671 & 0.700 & 0.629 & \textbf{0.793} & 141  \\
70  & 0.716 & \underline{0.752} & 0.716 & 0.702 & 0.667 & 0.653 & 0.688 & 0.638 & \textbf{0.809} & 141  \\
80  & 0.723 & \underline{0.738} & \underline{0.738} & 0.709 & 0.653 & 0.681 & 0.702 & 0.653 & \textbf{0.837} & 141  \\
90  & 0.716 & 0.738 & 0.745 & \underline{0.752} & 0.631 & 0.667 & 0.709 & 0.617 & \textbf{0.816} & 141  \\
100 & 0.723 & \underline{0.752} & 0.745 & 0.709 & 0.631 & 0.674 & 0.681 & 0.653 & \textbf{0.851} & 141  \\
% \midrule
\rowcolor{gray!15} % 设置本行背景为浅灰色
\multicolumn{11}{c}{\textbf{Performance (Long-context Style Alignment)}} \\
% \midrule
10  & 0.756 & \underline{0.829} & 0.707 & 0.707 & 0.659 & 0.683 & 0.707 & 0.683 & \textbf{0.854} & 41 \\
20  & \textbf{0.774} & 0.691 & 0.691 & 0.726 & 0.631 & 0.691 & 0.679 & 0.643 & \underline{0.762} &  84  \\
30  & \textbf{0.766} & 0.720 & 0.682 & \underline{0.729} & 0.636 & 0.664 & 0.664 & 0.645 & \textbf{0.776} &  107  \\
40  & \underline{0.764} & 0.732 & 0.701 & 0.732 & 0.606 & 0.654 & 0.677 & 0.622 & \textbf{0.772} &  127  \\
50  & \textbf{0.774} & 0.708 & 0.693 & 0.737 & 0.620 & 0.664 & 0.664 & 0.620 & \underline{0.766} &  137   \\
60  & \textbf{0.786} & 0.700 & 0.686 & \underline{0.729} & 0.636 & 0.650 & 0.700 & 0.636 & 0.721 &  140   \\
\midrule
70  & \textbf{0.771} & 0.579 & 0.700 & 0.707 & 0.650 & 0.679 & 0.671 & 0.643 & \underline{0.743} &  141   \\
80  & \textbf{0.773} & 0.603 & 0.695 & 0.738 & 0.645 & 0.688 & 0.681 & 0.660 & \underline{0.759} &  141   \\
90  & \textbf{0.773} & 0.489 & 0.702 & 0.745 & 0.617 & 0.702 & 0.695 & 0.653 & \underline{0.759} &  141   \\
100 & 0.695 & 0.000 & 0.695 & \underline{0.731} & 0.638 & 0.709 & 0.695 & 0.653 & \textbf{0.773} &  141   \\
\bottomrule
% \addlinespace
% \multicolumn{11}{p{\textwidth}}{\footnotesize \textit{Note: Qwen2.5-1M refers to Qwen2.5-14B-1M. RAG, as well as the other Memory Systems, are all based on GPT-4o-mini as the inference model. Due to the insertion of real user dialogues at different positions in the Long-context setting, the number of answerable questions varies slightly across positions.}} \\
\end{tabular}
}
\end{table}

\subsection{Further Analysis}

\subsubsection{Performance under Linguistic Style-aligned Long Contexts}
% sec:Style-Aligned
To evaluate the limits of standalone LLMs and memory systems, we extend our experiments to a long context setting with an average length of 116k tokens. This setup simulates large-scale interaction histories and stress-tests models’ ability to maintain effective memory under extreme context lengths.  Following the pipeline described in Section~\ref{sec:Style-Aligned}, we align the user queries of the \textit{Clean} setting with real-world conversational styles and populate the context with authentic dialogue segments sampled from WildChat~\cite{zhao2024wildchat1mchatgptinteraction}.

\begin{enumerate}[leftmargin=*, topsep=0pt, itemsep=0pt]
    \item \textbf{The Impact of Stylistic Interference and Token-level Noise.} In the single-domain task performance shown in Figure~\ref{tab:combined_all_turns}, comparing the base metrics (Clean setting in Table~\ref{tab:clean}) to the style-aligned results shows that the performance of most memory systems remains largely unchanged. This suggests that, as long as preference information and task requirements are preserved, variations in linguistic style have a limited impact on long-context understanding. 
    
    In contrast, the introduction of substantial irrelevant token noise reveals divergent behaviors between standalone LLMs and memory systems. For GPT-4o-mini, the model fails to respond under long dialogue inputs. Similarly, Qwen2.5-14B-1M, despite being fine-tuned on ultra-long contexts, exhibits a performance decline (from \textbf{0.766} to \textbf{0.716}). Memory systems, however, maintain stable performance, highlighting their robustness to token-level noise that is less relevant to personal preference.
    
    \item \textbf{Persona State Consistency in Long-Context Multi-Domain Settings.} The multi-domain setting exposes the fragility of most memory systems. While MemOS is the top-performing system with an \texttt{MCQ} \texttt{Acc.} of \textbf{0.764}, and its interaction performance also shows improvement compared to the Clean version (\texttt{Turn=1}: \textbf{0.306 $\rightarrow$ 0.331}; \texttt{Turn$\leq$2}: \textbf{0.592 $\rightarrow$ 0.637}). Systems like Supermemory and Mem0 witness a performance collapse (falling below \textbf{0.60}). This divergence suggests that maintaining persona consistency over 100k+ tokens requires resilient state tracking rather than simple semantic matching. Further analysis of memory content in Appendix~\ref{apx:case} shows that, although these systems improve token efficiency relative to MemOS, they primarily record episodic memories and fail to abstract and consolidate specific facts or preferences.

    \item \textbf{User Variability and Long-Context Degradation.} Figure~\ref{fig:heat} reveals that performance is non-uniform across different user profiles, highlighting the diversity present in the dialogues. As the context length increases, Figure~\ref{fig:heat}(2) illustrates a catastrophic performance degradation beyond the 60\% position (approximately 73k tokens), where the base model (GPT-4o-mini) reaches the limits of its effective context processing capability and ultimately fails at the 100\% mark. In contrast, memory systems such as MemOS exhibit remarkable stability. By transforming the global long-context reasoning problem into a persona-state matching task, MemOS effectively bypasses the physical context window limitations of the underlying LLM, thereby sustaining high accuracy even at extreme context length.

    \item \textbf{Robustness under Positional Probing.} Compared to prior work that evaluates preference at a fixed position, we assess the model's ability to maintain multiple preferences across diverse positions through event-driven persona state evolution. In contrast to the global query evaluation in Figure~\ref{fig:heat} (2), the results in Table~\ref{tab:combined_results_refined} specifically focus on queries associated with previously emerged domain events, measuring the variation in \texttt{MCQ} \texttt{Acc.} across positional checkpoints from 10\% to 100\% depth. While some memory systems experience slight performance variance within the median depth range, MemOS maintains superior and stable accuracy, surpassing even the Qwen2.5-14B-1M. This stability suggests that its indexing mechanism successfully handles semantic overlap and retrieval interference in dense context segments.
    
\end{enumerate}

\begin{table}[t]
\caption{Comparison of \texttt{MCQ} accuracy between LLMs and memory system-based agents across 20 domains, highlighting the challenges of maintaining consistent performance on heterogeneous tasks.}
\label{tab:model_performance_merged}
\resizebox{\textwidth}{!}{%
\begin{tabular}{lccccccccccccc}
\toprule
\textbf{Baseline} & \textbf{Alarm} & \textbf{Book} & \textbf{Calendar} & \textbf{Event} & \textbf{Finance} & \textbf{Media} & \textbf{Messaging} & \textbf{Movie} & \textbf{Music} & \textbf{Restaurant} & \textbf{Shopping} & \textbf{Travel} & \textbf{Others} \\
\midrule
\rowcolor{gray!15} % 设置本行背景为浅灰色
\multicolumn{14}{c}{\textbf{Longcontext}} \\ 
MiniMax-M2.5 & 0.722 & 0.680 & 0.744 & 0.611 & 0.829 & 0.689 & 0.890 & 0.700 & 0.800 & 0.690 & 0.800 & 0.800 & 0.678 \\
GLM-5 & 0.744 & 0.550 & 0.700 & 0.689 & 0.857 & 0.600 & 0.860 & 0.600 & 0.660 & 0.590 & 0.567 & 0.750 & 0.691 \\
Kimi-K2.5 & \textbf{0.856} & 0.780 & 0.867 & 0.833 & 0.914 & 0.733 & \textbf{0.980} & 0.757 & 0.810 & 0.630 & 0.833 & \textbf{0.870} & 0.797 \\
Qwen3-32B & 0.744 & 0.770 & 0.844 & 0.833 & \textbf{0.957} & 0.700 & 0.890 & 0.743 & \underline{0.880} & 0.790 & 0.833 & 0.820 & \textbf{0.822} \\
Qwen2.5-72B & 0.811 & 0.770 & 0.667 & \underline{0.878} & 0.900 & 0.522 & 0.820 & 0.629 & 0.800 & 0.630 & 0.811 & 0.810 & 0.725 \\
Qwen2.5-1M & 0.811 & 0.690 & 0.644 & 0.833 & 0.886 & 0.489 & 0.850 & 0.514 & 0.830 & 0.550 & 0.744 & 0.800 & 0.709 \\
Llama3.3-70B & 0.589 & 0.780 & 0.733 & 0.700 & 0.843 & \underline{0.767} & 0.870 & \underline{0.800} & 0.830 & 0.740 & 0.700 & 0.760 & 0.709 \\
GLM-4.7-Flash & \underline{0.833} & \textbf{0.830} & \textbf{0.889} & 0.767 & \underline{0.929} & 0.733 & 0.810 & \textbf{0.814} & 0.840 & \textbf{0.840} & 0.856 & \underline{0.850} & \textbf{0.822} \\
Gemini2.5-Flash & 0.800 & 0.750 & 0.800 & \textbf{0.911} & 0.900 & \textbf{0.778} & \underline{0.890} & 0.700 & \textbf{0.930} & 0.630 & \textbf{0.922} & 0.820 & \underline{0.809} \\
GPT-4o-mini & 0.700 & 0.770 & \underline{0.800} & 0.733 & 0.786 & 0.578 & 0.960 & 0.671 & 0.830 & 0.560 & \underline{0.678} & 0.760 & 0.691 \\
\midrule
\rowcolor{gray!15} 
\multicolumn{14}{c}{\textbf{RAG \& Memory System}} \\ 
Rag & 0.689 & 0.690 & 0.700 & 0.667 & 0.771 & 0.533 & \textbf{0.950} & 0.657 & 0.800 & 0.490 & 0.644 & 0.710 & \underline{0.681} \\
EverMemOS & 0.644 & 0.670 & \underline{0.767} & 0.689 & 0.743 & 0.567 & \textbf{0.950} & \underline{0.671} & \underline{0.820} & \underline{0.540} & 0.611 & 0.720 & 0.616 \\
Lightmem & 0.567 & \underline{0.740} & 0.744 & 0.589 & \underline{0.814} & 0.556 & 0.930 & 0.429 & 0.750 & 0.410 & 0.633 & 0.710 & 0.619 \\
Mem0 & 0.567 & 0.660 & 0.700 & 0.667 & 0.757 & \underline{0.589} & 0.860 & 0.529 & 0.740 & 0.470 & \underline{0.678} & 0.690 & 0.638 \\
Memobase & \underline{0.667} & 0.760 & 0.756 & \underline{0.678} & 0.800 & 0.567 & 0.900 & 0.571 & 0.790 & 0.520 & 0.689 & \underline{0.760} & 0.641 \\
MemOS & \textbf{0.732} & \textbf{0.810} & \textbf{0.820} & \textbf{0.821} & \textbf{0.859} & \textbf{0.657} & \textbf{0.950} & \textbf{0.694} & \textbf{0.848} & \textbf{0.719} & \textbf{0.889} & \textbf{0.856} & \textbf{0.777} \\
Supermemory & \underline{0.667} & 0.700 & 0.722 & 0.611 & 0.757 & 0.500 & 0.900 & 0.400 & 0.760 & 0.390 & 0.744 & 0.720 & 0.622 \\
\bottomrule
% \addlinespace
% \multicolumn{14}{p{\textwidth}}{\footnotesize \textit{Note: Qwen2.5-1M refers to the model Qwen2.5-14B-1M. RAG, as well as the other Memory Systems are all based on GPT-4o-mini as the inference model.}} \\
\end{tabular}
}
\end{table}

\begin{figure*}[!t]
    \centering
    % 左侧子图：单域实验结果
    \begin{subfigure}[b]{0.48\textwidth}
        \centering
        \includegraphics[width=\textwidth]{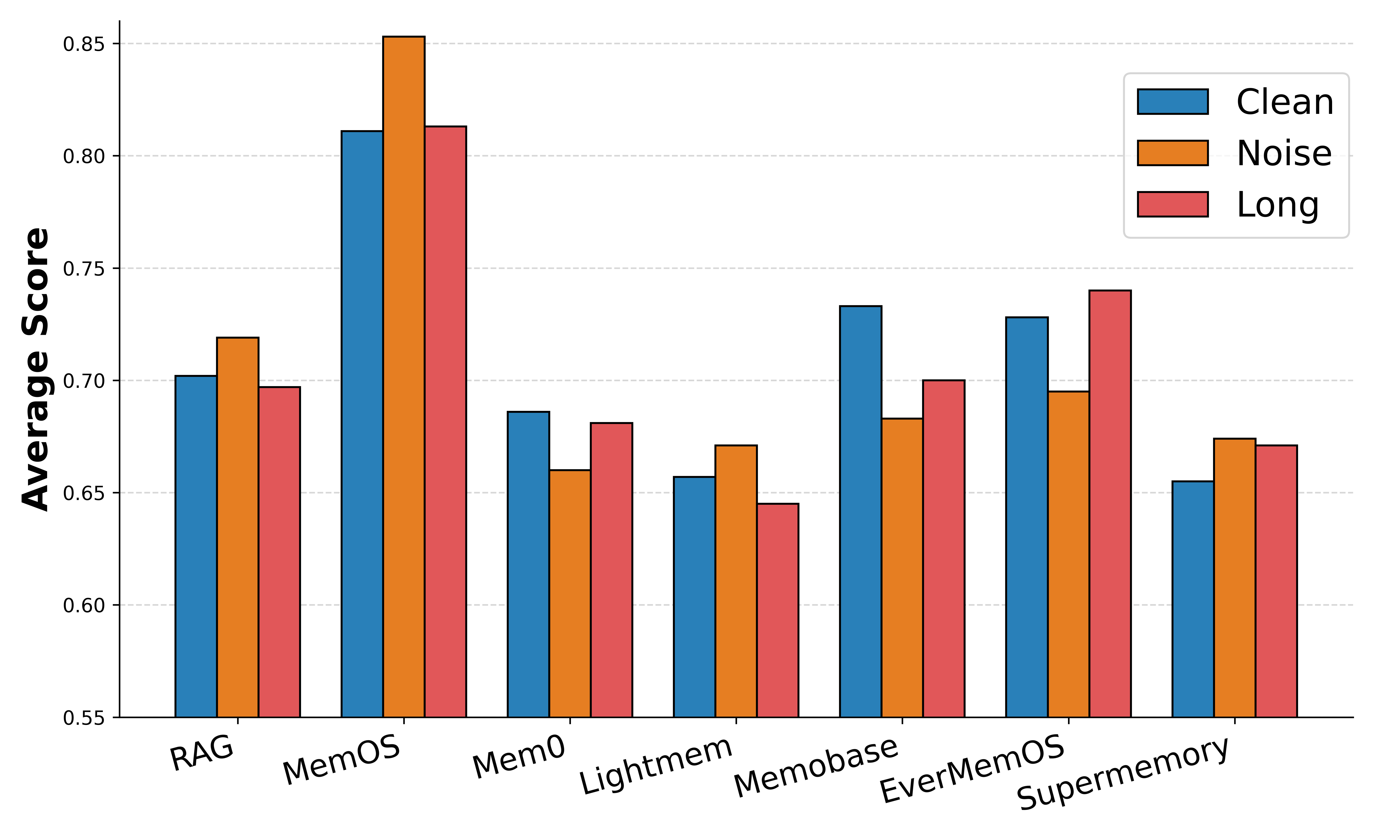}
    \end{subfigure}
    \hfill % 水平弹性间距，使左右子图完美向两端对齐
    % 右侧子图：多域实验结果
    \begin{subfigure}[b]{0.48\textwidth}
        \centering
        \includegraphics[width=\textwidth]{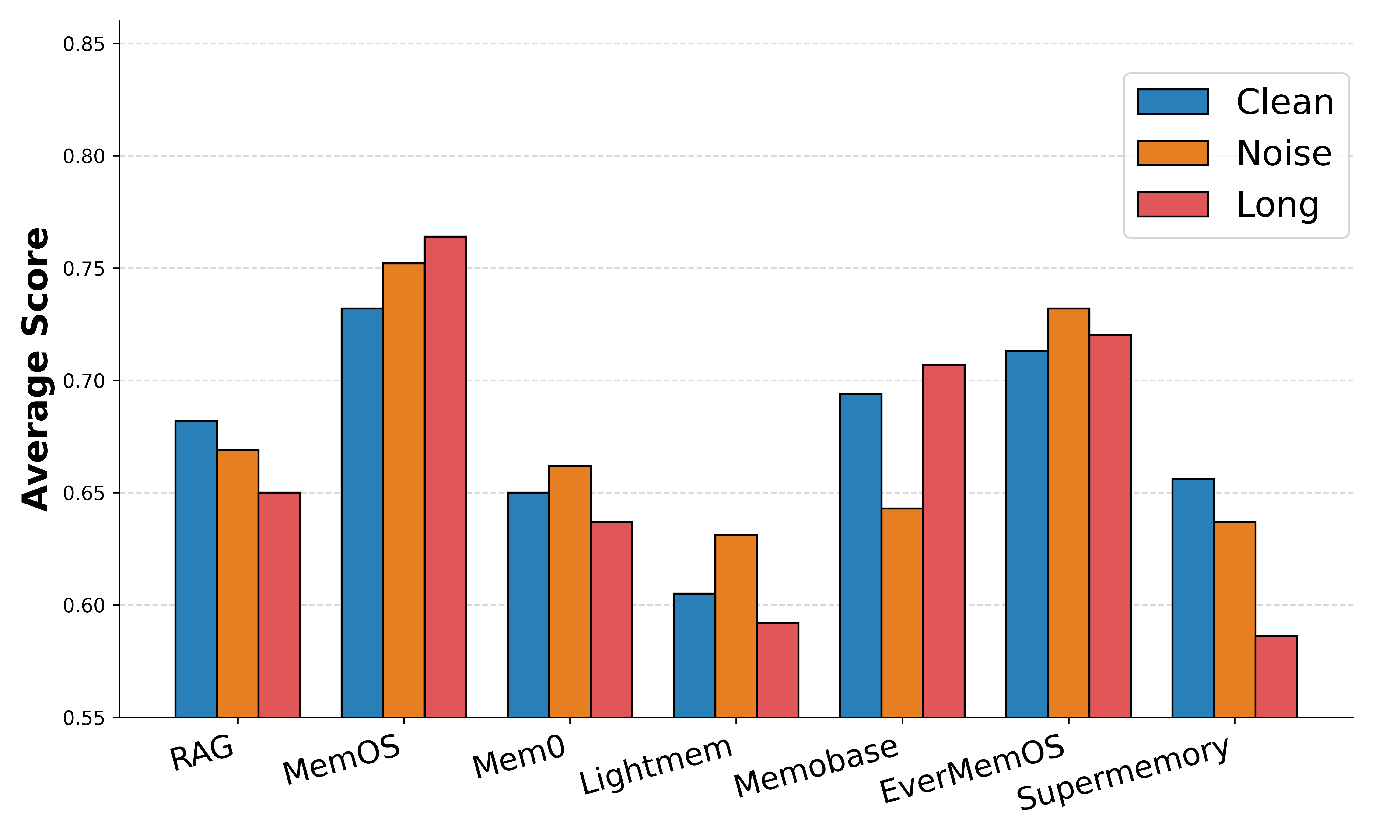}
    \end{subfigure}
    
    % \vspace{-2.1mm}
    \caption{Overall \texttt{MCQ} \texttt{Acc.} across all experimental configurations (Clean, In-session Noise, and Style-aligned Long-context): single-domain (\textbf{Left}) and multi-domain (\textbf{Right}) tasks.}
    \label{fig:postion}
\end{figure*}
% \subsubsection{Positional Variance and Noise Mitigation Strategies}

% As detailed in Table~\ref{tab:combined_results_refined}, the interaction between retrieval position (Pos. \%) and data integrity reveals a stark contrast between native long-context models and memory-augmented systems. Qwen2.5-14B-1M maintains a highly consistent performance profile ($\approx 0.77$) across the entire context window, showcasing its inherent capability for long-context style alignment. Conversely, GPT4o-mini encounters a failure at the $100\%$ retrieval depth, suggesting a threshold where its context window limits are reached.

% In the Noise setting, MemOS consistently achieves scores above $0.80$, frequently surpassing its own Clean baseline, which is consistent with our analysis in Figure~\ref{fig:Comparison_clean_and_noise}. This phenomenon suggests that the system’s indexing and filtering mechanisms are highly optimized for high-entropy environments, effectively mitigating distracting tokens that would otherwise degrade the performance of standard LLMs.

\subsubsection{Single-domain vs. Multi-domain Robustness}

% Figure~\ref{fig:postion} provides a comprehensive comparison across different settings, highlighting a consistent performance divergence between single-domain and multi-domain scenarios. MemOS achieves the best overall performance in both tasks. Notably, a distinct performance shift occurs in the style-aligned long-context setting relative to the Clean baseline.In single-domain tasks (Figure~\ref{fig:postion} Left), the style-aligned long-context typically yields the lowest accuracy. The significantly extended sequence length introduces irrelevant memory segments, which correlates with a performance decline in these focused tasks. Conversely, in multi-domain scenarios (Figure~\ref{fig:postion} Right), this same setting consistently outperforms the Clean configuration. This contrast indicates that while the total sequence length increases, the authentic data distribution of the style-aligned context provides more effective support for resolving cross-domain dependencies than the baseline or synthetic noise settings. RAG and other baselines remain sensitive to domain complexity and noise. By effectively isolating relevant persona states from high-entropy backgrounds, MemOS maintains superior architectural resilience when transitioning from focused interactions to integrated, multi-domain environments.

Figure~\ref{fig:postion} provides a comprehensive comparison across different settings, highlighting a performance divergence between single-domain and multi-domain tasks. In single-domain tasks (Figure~\ref{fig:postion} (left)), the style-aligned long-context (Long) setting generally acts as a performance bottleneck for most memory systems, such as Memobase, where accuracy falls below the Clean setting. This indicates that extended sequences introduce interference during retrieval in focused tasks. In multi-domain tasks (Figure~\ref{fig:postion} (righ)), the trends diverge: while RAG and Supermemory exhibit performance degradation under the Long setting, MemOS and Memobase instead show improvements over their Clean settings. This suggests that certain architectures benefit from the structured data distribution in style-aligned contexts, which helps resolve cross-domain dependencies.

% Figure~\ref{fig:postion} provides a comprehensive comparison across different settings, highlighting a performance divergence between single-domain and multi-domain tasks. MemOS achieves the best overall performance across both settings, while RAG and other systems remain sensitive to domain complexity and noise. Notably, a distinct performance shift occurs in the style-aligned long-context setting relative to the Clean baseline. In single-domain tasks (Figure~\ref{fig:postion} (1)), the style-aligned long-context setting typically yields the lowest accuracy. The significantly extended sequence length introduces additional context that may not directly contribute to the task, which is associated with a performance decline in these focused scenarios. Conversely, in multi-domain scenarios (Figure~\ref{fig:postion} (2)), this same setting consistently outperforms the Clean configuration. This contrast indicates that, despite the increased sequence length, the authentic data distribution in the style-aligned context appears to provide more useful signals for integrating information across domains compared to Clean and Noise settings.

\subsubsection{Topic-Specific Adaptation and Domain Expertise}

% Table~\ref{tab:model_performance_merged} underscores the proficiency of different architectures. Large-scale models such as Kimi-K2.5 exhibit peak performance in domains like Finance (\textbf{0.957}) and Messaging (\textbf{0.980}). Within the Memory System category, MemOS effectively bridges the gap between plug-in architectures and native models, particularly in the Shopping (\textbf{0.889}) and Travel (\textbf{0.856}) domains. However, a universal performance trough is observed in the Media and Restaurant sectors across all tested models. This is especially evident in scenarios requiring models to extract from historical memories to new preference queries. This indicates that memory systems still fail to track frequent updates. Rather than just archiving the past, memory should serve as a predictive tool for future events.

Table~\ref{tab:model_performance_merged} reveals clear domain-specific performance variations across different approaches. Large-scale models such as Kimi-K2.5 achieve the strongest results in structured domains like Finance (\textbf{0.914}) and Messaging (\textbf{0.980}), where task requirements are relatively stable. Within the memory systems, MemOS narrows the gap between plug-in memory-based agents and vanilla models, particularly in the Shopping (\textbf{0.889}) and Travel (\textbf{0.856}) domains, demonstrating improved utilization of historical preferences. In contrast, a consistent performance drop is observed in the Media and Restaurant domains across all evaluated models. These domains typically involve more dynamic and frequently updated user preferences, making them inherently more challenging. This suggests that memory should not function solely as a storage of historical interactions and preferences, but should be actively leveraged to guide future responses by transforming accumulated preference signals into informative priors for inference.

%% file: content/05_conclusion.tex
In this work, we introduce a benchmark that shifts the evaluation of personalized memory agents from static preference recall to persona consistency. By constructing temporally ordered interaction events, we propose an event-driven framework, \method, for assessing LLMs and memory systems under text variability and real-world linguistic alignment. Our experiments in single- and multi-domain tasks reveal that most memory-based systems outperform RAG by linking related interactions into a persistent persona state. This advantage becomes more pronounced in high-entropy settings, where memory systems demonstrate greater robustness and achieve 300$\times$ higher token efficiency than vanilla long-context models. Notably, while standalone LLMs exhibit severe performance degradation when context exceeds their window sizes, memory systems remain stable by reducing long-range reasoning to structured persona state matching. Despite these gains, performance degradation under multi-domain transitions and conflicting signals reveals a sensitivity to context complexity, highlighting the necessity of \method. Overall, our findings suggest that, while memory enhances robustness, achieving personalized agents requires moving beyond static retrieval toward task-aware memory management capable of filtering noise and maintaining long-term coherence, enabling lifelong companionship.

%% file: content/06_appendix.tex
\section{Prompt}
\label{apx:prompt}
To establish a realistic foundation for evaluation, the framework utilizes data construction prompts that simulate the evolution of user preferences. By leveraging TIMELINE\_GENERATION in Figure~\ref{fig:timeline}, the system generates interaction timelines, modeling how user interests emerge and deepen. These timelines are then transformed into naturalistic dialogues via the DIALOGUE\_GENERATION prompts in Figure~\ref{fig:generation1}, which guide the LLMs to reveal preferences through feedback or corrections, thereby mirroring the ambiguity of real-world interactions.

Building upon this data, the framework transitions to a dynamic interactive simulation driven by the User simulator. As the interaction unfolds, the USER\_FEEDBACK\_PROMPT in Figure~\ref{fig:interative} provides the user's responses, enabling the agent to provide adaptive feedback. This mechanism allows the simulated user to clarify intent or correct the assistant based on history, creating a test of the assistant's ability to maintain consistency and adapt to user guidance in real-time.

Finally, to quantify agent performance, the module implements a standardized ``LLM-as-a-Judge'' evaluation. We score the assistant's responses on quality and personalized memory (EVAL\_MEMORY\_SCORE in Figure~\ref{fig:memory_evaluation2}). Crucially, the system assesses memory precision, verifying the retrieval of details against ground truth. OPTION\_GENERATION\_PROMPT in Figure~\ref{fig:option} synthesizes challenging multiple-choice questions with options that are plausible but intentionally misaligned with user preferences or facts, thereby testing the agent's discriminative ability in noisy environments.

\section{Case Study}
\label{apx:case}
In this section, we present a case study comparing the retrieved memory of different systems to provide a clearer horizontal evaluation. As illustrated in Table~\ref{tab:case_study_1}--\ref{tab:case_study_2}, we can intuitively observe the retrieved memory results of various systems for the same query: MemOS categorizes memory into three distinct dimensions—episodic facts, explicit preferences, and implicit preferences—utilizing a parallel retrieval mechanism that effectively addresses the three categories of challenges inherent in preference-driven scenarios; meanwhile, Mem0, Memobase, and EverMemOS prioritize the concatenation of episodic memories, which makes them better suited as personal assistants for querying historical records but leaves them relatively weak in preference extraction and processing; conversely, Supermemory leans toward highly abstracted preference extraction but fails to reconstruct the specific details of relevant events along the timeline.

% \section{Manual Evaluation of Data Quality}
% To ensure the rigor and reliability of the synthetic dataset, we conducted a comprehensive manual evaluation focusing on the following core dimensions. \textbf{Preference Alignment Accuracy (Human Baseline)} evaluates the accuracy of responses in strictly adhering to the user’s predefined profiles and historical preferences within a given session. Manual verification reveals a $97.75\%$ accuracy rate for answers provided, given the established preferences and dialogue context, confirming the reliability of the ground truth labels in the dataset. \textbf{Task Completion} assessed whether the assistant provides executable solutions tailored to the user’s goals. The results indicate that the assistant consistently meets user intent by providing specific, actionable courses of action. \textbf{Content Coverage} measures the extent to which the dialogue covers key points outlined in the task description. The synthetic data achieved a coverage score of 1.99/2.0, signifying that nearly all required contextual details and functional requirements were addressed within the conversations. \textbf{Noise Robustness} assesses whether the pre-selected noise is correctly integrated and reflected within the dialogue. A noise usage score of 0.98/1.0 validates that the specific noise interference defined in the task is accurately applied, maintaining logical coherence while fulfilling the benchmark's requirements for complexity.

% ------------------prompt框架---------------------
\lstdefinestyle{prompt}{
    basicstyle=\ttfamily\fontsize{7pt}{8pt}\selectfont,
    frame=none,
    breaklines=true,
    backgroundcolor=\color{white},
    breakatwhitespace=true,
    breakindent=0pt,
    escapeinside={(*@}{@*)},
    numbers=none,
    numbersep=5pt,
    xleftmargin=5pt,
    aboveskip=2pt,
    belowskip=2pt,
}

\tcbset{
  aibox/.style={
    top=10pt,
    colback=white,
    % colframe=black,
    % colbacktitle=black,
    enhanced,
    center,
    % attach boxed title to top left={yshift=-0.1in,xshift=0.15in},
    % boxed title style={boxrule=0pt,colframe=white,},
  }
}
\newtcolorbox{AIbox}[2][]{aibox, title=#2,#1}

% Slightly darker frame colors (recommended for visibility)
\colorlet{PBRedFrame}{red!60}
\colorlet{PBPurpleFrame}{purple!60}
\colorlet{PBOrangeFrame}{orange!60}
\colorlet{PBBlueFrame}{blue!50}

% ========== Frame-only box templates (white background) ==========
\tcbset{
  aiboxFrameOnly/.style={
    aibox,
    colback=white,      % keep background pure white
    colframe=black,     % default (overridden in variants)
    boxrule=0.8pt,
    arc=2pt,
    left=6pt,right=6pt,top=6pt,bottom=6pt,
  },
}

\newtcolorbox{AIboxRedFrame}[2][]{aiboxFrameOnly, colframe=PBRedFrame, title=#2,#1}
\newtcolorbox{AIboxPurpleFrame}[2][]{aiboxFrameOnly, colframe=PBPurpleFrame, title=#2,#1}
\newtcolorbox{AIboxOrangeFrame}[2][]{aiboxFrameOnly, colframe=PBOrangeFrame, title=#2,#1}
\newtcolorbox{AIboxBlueFrame}[2][]{aiboxFrameOnly, colframe=PBBlueFrame, title=#2,#1}
% ------------------prompt框架---------------------

\begin{figure*}[h]
\begin{AIboxRedFrame}{TIMELINE\_GENERATION}
{
\footnotesize
\setlength{\parindent}{0pt}
\setlength{\parskip}{2pt}

You are a professional expert in user behavior analysis. Your task is to convert summarized user-assistant interactions into a coherent timeline of preference development events. Please make sure to consider user demographics and extend the content appropriately based on user preferences.

\textbf{\#\# User Profile Information}

- User Demographic Profile: \textcolor{gray}{\{demographic\_profile\}}

\textbf{\#\# Summarized Interaction History:} \textcolor{gray}{\{topic\_interactions\}}

- Relevant Domain: \textcolor{gray}{\{selected\_topics\}}

("Relevant Domain" refers to the topic domain that appeared in the current interactions, indicating the area in which the user shows interest or preference.)

The order of interactions represents their chronological sequence, and each interaction includes a description of the user's preferences and possible events related to that domain.

\textbf{\#\# Timeline Generation Requirements}

1. Date Span: Generate events over a 1-6 month period **in chronological order**, showing natural preference development. Each event should have a specific time of day: Morning, Afternoon, or Evening, based on the interaction context.

2. Event Types: Only include two types of events — "preference\_emergence" and "preference\_supplement".

  - preference\_emergence: The first interaction and initial interest in a domain.
  
  - preference\_supplement: A supplement, deepening, or change in the user's existing preferences **after they have already discovered the domain**.

3. Coherence and Progression: Ensure that events are logically connected and reflect realistic user behavior patterns. Show how the user first discovers a domain (emergence) and later supplements their preferences (supplement).

4. Domains: Each event must correspond to the same domain as in the selected user interactions. 

5. If an event explicitly refers to content from other domains, you need to include those domains in the dependencies, e.g., `he has asked about setting multiple alarms with different sounds or integrating his **calendar events** to adjust alarm times accordingly.` Otherwise, simply set it to `None`. The other domains include \textcolor{gray}{\{other\_domains\}}.

6. **Every interaction (overall \textcolor{gray}{\{topic\_interactions\_number\}}) in the interaction history needs to be used to generate events.**

\textbf{\#\# Output Format}

The task\_id field is composed of the event\_index and event\_type, for example: event\_1\_preference\_emergence, event\_2\_preference\_supplement. 

Each timeline must start with an event of type preference\_emergence, with all subsequent events labeled as supplement, and **the event\_index should be initialized at 1**.

Generate the interaction timeline in JSON format

(**Include \textcolor{gray}{\{topic\_interactions\_number\}} events**, using English):

}
\end{AIboxRedFrame}
\caption{TIMELINE\_GENERATION}
\label{fig:timeline}
\end{figure*}

% \verb|```|json

% \{\{

% \quad"events": [
  
% \quad\quad\{\{
    
% \quad\quad\quad"date": "YYYY-MM-DD Morning/Afternoon/Evening",
      
% \quad\quad\quad"event\_type": "preference\_emergence",
      
% \quad\quad\quad"description": "Detailed description of the user's first encounter with the domain, based on a single interaction history",
      
% \quad\quad\quad"task\_goal": "Task-specific goal or action",
      
% \quad\quad\quad"relevant\_domain": \textcolor{gray}{\{selected\_topics\}},  
      
% \quad\quad\quad"checkpoint": false,
      
% \quad\quad\quad"task\_id": "event\_(event\_index)\_(event\_type)",
      
% \quad\quad\quad"dependencies": (other domain name) or "None"
      
% \quad\quad\}\},
    
% \quad\quad...
    
% \quad]
  
% \}\}

% \verb|```|

\begin{figure*}[!t]
\begin{AIboxRedFrame}{ANSWER\_OPTION\_PROMPT}
{
\footnotesize
\setlength{\parindent}{0pt}
\setlength{\parskip}{2pt}

You are an assistant specialized in answering multiple-choice questions.

\textbf{\#\# Your Memory}

\textcolor{gray}{\{context\}}

\textbf{\#\# User Task Query}

\textcolor{gray}{\{question\}}

\textbf{\#\# Options:}

\textcolor{gray}{\{options\}}

Your goal is to choose **the most appropriate answer option for the User Task Query** from the Options based on your memory. The output should be **ONLY the option key** without any additional explanation, e.g., `A`, etc.

Your response:
}
\end{AIboxRedFrame}
\caption{ANSWER\_OPTION\_PROMPT}
\label{fig:_OPTION}
\end{figure*}

\begin{figure*}[!t]
\begin{AIboxRedFrame}{DIALOGUE\_GENERATION}
{
\footnotesize
\setlength{\parindent}{0pt}
\setlength{\parskip}{2pt}

You are a professional expert in user behavior analysis. Your task is to generate a natural, realistic multi-turn conversation between a User and a virtual Assistant that reveals user preferences within the domain specified by \{relevant\_domains\} based on the given context.

**Preference Supplement Moment:**

This is a conversation where the user may supplement additional information or provide new preferences in the domain of \{relevant\_domains\}. The user's preference may emerge implicitly (through feedback, comparisons, corrections or reactions to Assistant suggestions, e.g., "Maybe something lighter would suit me better" instead of directly stating "I prefer X"). 

- If the Dependency Domains are not set to None, you also need to take into account the information in User Preferences for those domains.

\textbf{\#\# User Profile Information}

- User Demographic Profile:\textcolor{gray}{\{demographic\_profile\}}

- User Preferences:\textcolor{gray}{\{preference\_str\}}

The user must not introduce new preferences or options that are not included in the provided User Preferences. All preferences discussed, confirmed, or supplemented by the User within the conversation must strictly adhere to and originate from the provided User Preferences.

\textbf{\#\# User-Assistant Past Interaction:} \textcolor{gray}{\{message\_history\}}

Next, you need to generate the conversation for the following task by considering the information comprehensively.

\textbf{\#\# Task Information:}

- Task Description: \textcolor{gray}{\{task\_description\}}

- Relevant Domains: \textcolor{gray}{\{relevant\_domains\}}

- Task Goal: \textcolor{gray}{\{task\_goal\}}

- Dependency Domains: \textcolor{gray}{\{dependency\_domains\}}

\textbf{\#\# Conversation Guidelines:}

1. Generate a realistic multi-turn (5-10 turn) User-Assistant conversation for the above task where the user continues to interact with the assistant in the topic of \{relevant\_domains\}.

2. You must consider the user's demographic information. If the task requires specific content from the past interactions, you must also take into account the past interactions in user's profile to inform the preference evolution. All dialogue turns must be goal-oriented, focusing strictly on task completion.

3. The assistant in the conversation must identify the user's preferences and requirements through feedback and inquiry. Furthermore, the assistant must diligently and thoroughly complete every user query during interaction until the entire task is finalized, ensuring each response is substantial and informative (aiming for 3-5 sentences).

4. The conversation must integrate and fulfill **all key components and concrete facts specified in the Task Description**! (e.g., if the Task Description mentions 'requesting portfolio recommendations tailored to risk tolerance,' the user's query must contain those specific elements.)

5. The conversation starts with the user posing a **task-oriented query** and aims to conclude with the **assistant's successful task completion and user satisfaction**.

6. Summarize the User's preferences that were discussed, confirmed, or supplemented within the dialogue. **Only include preferences that were used by the User during the conversation.**

\textbf{\#\# Requirements for Generating the Dialogue:}

1. **User Self-Clarification:** After receiving the assistant's reply, the user may realize their previous query was not clearly expressed and actively clarifies or reformulates it.

2. **Progressive Interaction:** The user continuously follows up to further specify or narrow down a previous topic, leading to a gradual supplement of preferences or requirements.

3. **Full Task Completion Requirement**: When the user requests a task (e.g., email writing, recommendation, comparison, booking assistance, planning, summarization, message crafting, etc.), the assistant MUST provide a real, fully actionable output, not just a template or placeholder.

\textbf{\#\# Output Format:}

\verb|```|json

\{\{

  \quad"conversation": "User: [Initial user query about changing needs or dissatisfaction]\verb|\n|Assistant: [Assistant response]\verb|\n|User: [User feedback]\verb|\n|Assistant: [Assistant follow-up helping supplement and complete the task as thoroughly as possible]\verb|\n|...",
  
  \quad"preferences": "The user preferences mentioned in the conversation follow the format: domain-preference\_name-content, e.g., Books-Favourite Authors-Michael Crichton, ..."
  
\}\}

\verb|```|

Generate the conversation immediately in the specified JSON format.

}
\end{AIboxRedFrame}
\caption{DIALOGUE\_GENERATION}
\label{fig:generation1}
\end{figure*}

\begin{figure*}[!t]
\begin{AIboxRedFrame}{USER\_FEEDBACK\_PROMPT}
{
\footnotesize
\setlength{\parindent}{0pt}
\setlength{\parskip}{2pt}

You are tasked with generating realistic user responses in a conversation with a personalized assistant.

**Remember**: In this task, you are not an assistant; you are **the user** seeking help. Maintain this perspective throughout the conversation. Here is your profile:

\textbf{\#\# Profile:}

- Your Profile Information: \textcolor{gray}{\{demographic\_profile\}}

- Your Preferences: \textcolor{gray}{\{preference\_str\}}

This is the relevant interaction with the assistant:

\textbf{\#\# Past Interaction History:} \textcolor{gray}{\{user\_topic\_dialog\}}

Next, you need to provide feedback based on the assistant's last reply in the Current Task Interaction by considering the following information of the current task comprehensively.

- Task Description: \textcolor{gray}{\{task\_description\}}

- Task Goal: \textcolor{gray}{\{task\_goal\}}

- Task Question: \textcolor{gray}{\{task\_question\}}

- Current Task Interaction with Assistant: \textcolor{gray}{\{history\}}

Your responses should follow these guidelines:

1. Be natural and conversational, avoiding artificial or robotic language.

2. Reflect your profile and preferences provided, and take Past Interaction History into account. Staying consistent with the user's personality throughout the conversation.

3. Keep each response focused and concise (1-3 sentences maximum)

4. As soon as BOTH are true:

   \quad- The assistant's latest reply provides the information you requested, successfully addressing the Task Question and achieving the Task Goal.
   
   \quad- The response also aligns with your stated preferences regarding the task.
   
   \quad you must output **TERMINATE** immediately. NEVER add new requests if the task is already satisfied.
   
5. If the assistant's latest reply does not meet the above conditions, provide a focused, concise correction or supplement to drive the task directly toward completion.

Your response:
}
\end{AIboxRedFrame}
\caption{USER\_FEEDBACK\_PROMPT}
\label{fig:interative}
\end{figure*}

\begin{figure*}[!t]
\begin{AIboxRedFrame}{ANSWER\_INTERACTION\_PROMPT}
{
\footnotesize
\setlength{\parindent}{0pt}
\setlength{\parskip}{2pt}
You are a conversational AI assistant focused on creating natural, thorough, and personalized interactions to complete the user query.

Below is the memory accumulated from your past interactions with this user 

\textbf{\#\# Your Memory}

\textcolor{gray}{\{context\}}

\textbf{\#\# User Task Query}

\textcolor{gray}{\{question\}}

\textbf{\#\# Current Task Conversation History:}
\textcolor{gray}{\{history\}}

You need to provide a reply based on the user's last query in the Current Task Interaction.

\textbf{\#\# Guidelines}
1. Your goal is to provide targeted, complete responses by actively integrating **the user's preferences from your memory**, ensuring the response is tailored and moves the task toward completion.

2. You need to supplement your answer according to the most recent user feedback in the Conversation History, aiming to immediately initiate the task and address the User Task Query as fully as possible.

3. Do NOT introduce unrelated topics or unnecessary follow-up questions.

4. Full Task Completion Requirement: When the user requests a task (e.g., email writing, recommendation, comparison, booking assistance, planning, summarization, message crafting, etc.), the assistant MUST provide a real, fully actionable output, not just a template or placeholder.

Your response:
}
\end{AIboxRedFrame}
\caption{ANSWER\_INTERACTION\_PROMPT}
\label{fig:_INTERACTION}
\end{figure*}

\begin{figure*}[t]
\begin{AIboxRedFrame}{OPTION\_GENERATION\_PROMPT}
{
\footnotesize
\setlength{\parindent}{0pt}
\setlength{\parskip}{2pt}

You are an AI behavior analysis expert. Your task is to produce eight distinct response options that an AI assistant might generate for the given user query. Each option (generated response) must be fluent and consistent with the assistant's tone.

1. \textbf{INPUT CONTEXT}

\textbf{\#\#\# Task Information}

- Task Description: \textcolor{gray}{\{task\_description\}}

- Task Goal: \textcolor{gray}{\{task\_goal\}}

- User Query for the Task: \textcolor{gray}{\{user\_query\}}

\textbf{\#\#\# Interaction History and Preference}

- Interaction History: \textcolor{gray}{\{user\_topic\_dialog\}}

- User Long-Term Preference Category: \textcolor{gray}{\{preferences\}}

2. \textbf{GENERATION OBJECTIVES}

Generate eight options (A-H) that form a permutation over the following binary dimensions:

**Task Completion (T):**

- T=1 (success): The response accurately captures the information sought in the **User Query** and completely satisfies the defined **Task Goal**.

- T=0 (failure): The response appears relevant to the **Task Description** and includes recall-oriented content, but does not fully accomplish the task objective.

**Preference Consistency (P):**

- P=1 (success): The content of the response consistently aligns with the **user's long-term preference categories**. These types of options require you to carefully read the **Interaction History** and use concrete facts explicitly present in the Interaction History to ensure preference consistency.

- P=0 (failure): The response operates within the correct preference category but introduces inferred, exaggerated, or unsupported preference details that are **not grounded in the Interaction History**.

**Information Confidence (I):**

- I=1 (success): The task is completed thoroughly and definitively, with no expressions of uncertainty anywhere in the response.

- I=0 (failure): The response completes the task but concludes by expressing uncertainty about the final answer, implying that a more satisfactory answer may exist.

3. \textbf{RESPONSE OPTION GUIDELINES}

Option (A) [T=1 / P=1 / I=1] — Optimal Resolution

- Provide a clear response that fully satisfies the task goal. (T=1)

- Correctly prioritize the user's long-term preference categories. Ground the response explicitly in **concrete details from the Interaction History (facts, constraints, preferences)**. (P=1)

- Use historical information confidently and efficiently, without uncertainty and additional requests throughout the response. (I=1)

Option (B) [T=1 / P=1 / I=0] — Procedural Overreach and Uncertainty

- Reach the same correct response as (A). (T=1/P=1)

- Correctly apply long-term preferences using dialogue-grounded details. (P=1)

- Express unconfidence and uncertainty stemming from a lack of relevant knowledge. (I=0)

...

Option (H) [T=0 / P=0 / I=0] — Systemic Misalignment

- Fails across task interpretation, preference fidelity, and information confidence. (T=0 / P=0 / I=0)

- Densely references interaction history without improving relevance.

4. \textbf{OUTPUT FORMAT}

User Query for the Task: \textcolor{gray}{\{user\_query\}}

Output options must be valid JSON with exactly eight fields: "A" through "H".
}
\end{AIboxRedFrame}
\caption{OPTION\_GENERATION\_PROMPT}
\label{fig:option}
\end{figure*}

\begin{figure*}[t]
\begin{AIboxRedFrame}{EVAL\_MEMORY\_SCORE}
{
\footnotesize
\setlength{\parindent}{0pt}
\setlength{\parskip}{2pt}

\textbf{<instructions>}

Evaluate the assistant’s preference memory using the following criteria:

1. **Coverage (Recall):**

- Compare the `Assistant Retrieved Memory' against `Preferences to be Mastered'.

- Does the retrieved memory include all critical preference points?

- Missing a core preference is a major failure; missing minor details is a minor failure.

2. **Accuracy \& Consistency (Precision):**

- Is the retrieved information consistent with the specific preference details expressed in the conversation?

- **Crucial:** Check for hallucinations (invented preferences) or outdated information (preferences that have since changed).

- Contradictory information without resolution is considered a failure.

3. **Relevance \& Noise:**

- Raw fragments are acceptable, but the memory should not be overwhelmed by irrelevant noise that obscures the true preferences.

- Memory should be concise, focused, and directly usable to respond to the user query.

\textbf{<Scoring Guidelines>}

Score of 1: POOR (Memory Failure / Hallucination)

- Core preferences are missing entirely.  

- Contains significant incorrect information (hallucinations) or direct contradictions to `Preferences to be Mastered'.  

- Retrieval is completely unrelated to the user’s preferences.  

- Excessive repetition of dialogue details obscures key information.

Score of 2: BASIC (Incomplete / Fragmented)

- Some relevant preferences retrieved, but key specific details are missing.  
- Correct information is present but buried under excessive noise.  

- May include outdated or imprecise details.  

- Minor redundancy present; core information is partially retrievable.

Score of 3: STRONG (Accurate but Unrefined)

- All major preferences are retrieved.  

- No hallucinations or contradictions.  

- May miss subtle implicit nuances or contain slightly cluttered raw chunks, but overall core truth is intact.  

- Information is comprehensive and relevant, though not fully polished.

Score of 4: EXCEPTIONAL (Perfect Recall)

- All `Preferences to be Mastered' are identified and retrieved.  

- Memory is precise, accurate, and entirely relevant to the context.  

- Clear and concise presentation; fully supports the user query.  

- Information is complete, distilled, and polished for usability.

\textbf{<response\_format>}

Memory Score: [1-4]

Key Observations:

- **Coverage Check:** [List which preferences from 'Mastered' were retrieved vs. missing]  

- **Accuracy Check:** [Note any hallucinations, contradictions, or outdated info]  

- **Format / Noise:** [Comment briefly if format or noise affected clarity, though coverage is the priority]  

- **Information Quality:** Memory should avoid excessive repetition and should condense dialogue information relevant to the query.

\textbf{</response\_format>}

Insert data below for evaluation:

\textbf{<conversation>}\{conversation\}\textbf{</conversation>}

\textbf{<preferences\_to\_be\_mastered>}\{preferences\_to\_be\_mastered\}\textbf{</preferences\_to\_be\_mastered>}

\textbf{<user\_query>}\{query\}\textbf{</user\_query>}

Next, you need to evaluate the "assistant\_retrieved\_memory" according to the given context and the guidelines mentioned above.

\textbf{<assistant\_retrieved\_memory>}\{assistant\_retrieved\_memory\}\textbf{</assistant\_retrieved\_memory>}

\textbf{<response>}Provide your evaluation enclosed within <response></response> tags following the response format above.\textbf{</response>}

}
\end{AIboxRedFrame}
\caption{EVAL\_MEMORY\_SCORE}
\label{fig:memory_evaluation2}
\end{figure*}

%--------------------------------Case Study-------------------------
% \begin{table}[htbp]
%     \centering
%     \caption{Hyperparameter Settings for Dataset Construction and Evaluation}
%     \label{tab:hyperparameters}
%     \renewcommand{\arraystretch}{1.1} % 调整行高
%     \begin{tabularx}{\textwidth}{@{}lXlXlX@{}} % 使用 tabularx 自动调整列宽
%         \toprule
%         \multicolumn{2}{c}{\textbf{Dataset Construction}} & \multicolumn{2}{c}{\textbf{User Simulator}} & \multicolumn{2}{c}{\textbf{Evaluation Setup}} \\
%         \cmidrule(r){1-2} \cmidrule(lr){3-4} \cmidrule(l){5-6}
%         Param & Value & Param & Value & Param & Value \\
%         \midrule
%         Model & GPT-4o & Model & GPT-4o & Emb. Model & BGE-M3 \\
%         Temp. & 0.5 & Temp. & 0.5 & Top-$k$ & 20 \\
%         Max Tokens & 16,384 & Max Tokens & 2,048 & Users & 10 \\
%         $K_{single}$ & 3 & Max Turns & 10 & & \\
%         $K_{multi}$ & 6 & & & & \\
%         Max Run & 2 & & & & \\
%         Switch Prob. & 0.3 & & & & \\
%         \bottomrule
%     \end{tabularx}
% \end{table}

\begin{table}[ht]
    \centering
    \small % 适当缩小字号
    \caption{Comparison of retrieval memory context among baselines in the single-domain setting.}
    \label{tab:case_study_1}
    \renewcommand{\arraystretch}{1.3} % 适度增加行距
    
    % 第一列(模型名)左对齐且自动加粗，第二列(生成内容)自动折行铺满剩余空间
    \begin{tabularx}{\textwidth}{@{} >{\bfseries}l X @{}}
        \toprule
        Baseline & \textbf{Domain: Hotel Query}: Can you help me find a hotel in Rome, Italy? I'm looking for some good options for my stay. \\
        \midrule
        RAG  & \textbf{User}: Hi there! I'm planning a trip to Italy and would love some help crafting an itinerary that immerses me in the country's rich art and history. I'm particularly interested in unique cultural activities like cooking classes and visiting lesser-known towns. Could you assist me with that? \newline \textbf{Assistant}: Absolutely, I can help with that. Are there any specific hotel chains you prefer, or do you have a preferred location or star rating in mind for your stay abroad?\newline \textbf{User}: I prefer Radisson, Hyatt, or Intercontinental hotels. I'm looking for something in the city center or near cultural landmarks, ideally a 4 or 5-star rating.\newline \textbf{Assistant}: Great choices! Let me search for available options that match your preferences. I'll focus on Radisson, Hyatt, and Intercontinental hotels located in city centers or near cultural landmarks, offering the amenities you mentioned. Would you like information on room types as well? \newline \textbf{User}: Yes, please. I'm considering either a double room or a suite... \textbf{following 957 tokens.} \\
        \addlinespace % 在不同模型的结果之间添加少许空白，提升可读性
        
        LightMem  & 2023-08-27T15:00:00.000 Sun User is planning a trip abroad and needs help finding a hotel.2023-08-27T15:00:00.000 Sun User is looking for hotels in the city center or near cultural landmarks.2023-09-23T15:00:00.000 Sat User is interested in hotels in the city center near cultural sites with amenities such as WiFi, free breakfast, and a pool.2023-09-24T15:00:00.000 Sun User is planning a trip to Italy.2023-09-22T06:00:00.000 Fri User is looking for a luxurious hotel near the Sultanahmet district and Hagia Sophia.2023-09-23T15:00:00.000 Sat User is planning a trip and is looking for unique boutique hotels with rich cultural heritage.2023-09-21T19:00:00.000 Thu User is interested in hotels within walking distance of Bauhaus architecture and the city's nightlife scene.2023-08-27T15:00:00.000 Sun User wants to focus on hotel amenities like complimentary WiFi, free breakfast, pool, gym facilities, and an on-site bar or lounge area.2023-09-24T15:00:00.000 Sun User is interested in immersing themselves in Italy's rich art and history. 2023-07-29T06:00:00.000 Sat User is considering hotels like Radisson or Hyatt.\\

        \addlinespace
        MemOS  & User[2023-07-29T06:00:00]: Oh, I should have mentioned, I'm interested in staying close to the city center or near cultural landmarks. Also, amenities like WiFi, free breakfast, and a pool would be great. Assistant[2023-07-29T06:00:00]: Got it! Let me narrow down some options for you. Based on your preferences, I found a few 4 and 5-star hotels in the city center and near cultural attractions. Both Radisson and Hyatt offer free WiFi, breakfast, and have pools. Would you like me to check availability or provide more details on these options? User[2023-08-27T15:00:00]: Yes, please. I'm considering either a double room or a suite, as I'm traveling with my spouse... \textbf{following 890 tokens}.\newline  \textbf{Explicit Preference}: 1. The user is looking for unique boutique hotels or properties with rich cultural heritage that offer an exceptional experience with comfort, convenience, and an authentic sense of place. 2. The user is interested in hotels located in the city center or near cultural sites and values amenities such as WiFi, free breakfast, and a pool. User prefers Radisson, Hyatt, or Intercontinental hotels with amenities like a rooftop pool, complimentary WiFi, free breakfast, and gym facilities. They are considering either a double room or a suite. The user prefers hotels within walking distance of Bauhaus architecture and the city's nightlife scene in Tel Aviv, specifically from the Radisson, Hyatt, and Intercontinental chains. They want a hotel with amenities like complimentary WiFi, free breakfast, a pool, gym facilities, and an on-site bar or lounge area, preferring Radisson, Hyatt, or Intercontinental hotels with a 4 or 5-star rating.\newline \textbf{Implicit Preference}:1. Preference for hotels with reputable brands. 2. Preference for accommodations that offer unique cultural experiences.3. Preference for personalized and immersive travel experiences that offer a combination of historical exploration, local culture, scenic beauty, and attending festivals related to hobbies or unique experiences...\textbf{following 200 tokens}.\\

        \addlinespace

        Supermemory & User needs help finding a hotel. User is looking for a hotel near Hagia Sophia in Istanbul. User is looking for a luxurious hotel in Istanbul. User is looking for a hotel near Sultanahmet district in Istanbul. User is planning a trip to Italy. User wants off-the-beaten-path experiences for Italy trip. User wants help crafting an itinerary for Italy trip. User is interested in unique cultural activities for Italy trip. User wants hotels near the city's nightlife scene. User wants Italy trip to immerse in art and history. \\

        \bottomrule
    \end{tabularx}
\end{table}

\begin{table}[ht]
    \centering
    \small % 适当缩小字号
    \caption{Comparison of retrieval memory context among baselines in the single-domain setting.}
    \label{tab:case_study_2}
    \renewcommand{\arraystretch}{1.3} % 适度增加行距
    
    % 第一列(模型名)左对齐且自动加粗，第二列(生成内容)自动折行铺满剩余空间
    \begin{tabularx}{\textwidth}{@{} >{\bfseries}l X @{}}
        \toprule
        Baseline & \textbf{Domain: Hotel Query}: Can you help me find a hotel in Rome, Italy? I'm looking for some good options for my stay. \\
        \midrule
        Memobase  & ---\# Memory Unless the user has relevant queries, do not actively mention those memories in the conversation.\#\# User Current Profile:- event::rav\_kav\_application: User is applying for a Rav-Kav card. [mention 2023-10-22]- interest::travel: User is a frequent traveler; they are looking for experiences that are off-the-beaten-path, particularly in lesser-known towns; they are considering visiting Santorini and Delphi, exploring ancient ruins, and experiencing local cuisine and culture; User is interested in sustainable tourism practices, including eco-friendly accommodations and activities. [mention 2023-10-22, 2023-10-07]- event::pricing\_inquiries: User is interested in the pricing and additional services for the trip, including potential discounts for frequent travelers; they requested help with understanding how to get a Rav-Kav card and the discounts it offers. [mention 2023-10-22]- event::bus\_travel\_enhancement: User is interested in enhancing their bus travel experience, particularly regarding entertainment options available on Egged and Dan buses. [mention 2023-10-22]- event::bus\_preferences: User is interested in both Egged and Dan buses for their trips between Tel Aviv and Jerusalem [mention 2023-10-22]- schedule::notifications: User enabled real-time updates for their bus trip to be notified of any schedule changes or delays... \textbf{following 746 tokens}.\\
        \addlinespace
        
        Mem0 & `2023-09-21T15:00:00-07:00: User is planning a trip to Istanbul and is looking for a luxurious hotel near the Sultanahmet district and Hagia Sophia', `2023-09-23T00:00:00-07:00: User wants hotel recommendations located in the city center or near cultural sites with WiFi, free breakfast, and a pool', "2023-09-24T00:00:00-07:00: User wants an itinerary that immerses in Italy's art and history", `2023-10-14T04:00:00-07:00: User is planning a multi-city trip to Europe including Amsterdam, Berlin, and Prague', `2023-09-21T15:00:00-07:00: User requests a curated list of Radisson, and Intercontinental hotels in Tel Aviv, focusing on room rates for suites at Radisson Blu Pera, amenities, and proximity to Bauhaus architecture and nightlife', `2023-08-27T00:00:00-07:00: User is planning a trip abroad and wants assistance finding a hotel with complimentary WiFi, free breakfast, a pool, gym facilities, and an on-site bar or lounge area', `2023-08-27T00:00:00-07:00: User wants to book a suite at Hyatt near cultural landmarks with amenities', `2023-09-24T00:00:00-07:00: User wants information about events or festivals during the fall in Lucca and Umbria', `2023-09-21T15:00:00-07:00: User wants to book the suite at Radisson Blu Pera, which has a rooftop pool and central location.' \\
        \addlinespace

        EverMemOS  & On August 27, 2023 (Sunday) at 12:00 AM UTC, the user initiated a conversation requesting assistance to find a hotel for an upcoming trip abroad. The user specified a preference for hotels offering amenities such as complimentary WiFi, free breakfast, a pool, gym facilities, and an on-site bar or lounge area. The assistant responded promptly, asking if the user had any preferred hotel chains, locations, or star ratings. The user indicated a preference for Radisson, Hyatt, or Intercontinental hotels, ideally located in city centers or near cultural landmarks, with a 4 or 5-star rating. The assistant acknowledged these preferences and offered to search for suitable options, also inquiring if the user wanted information on room types. The user confirmed interest in either a double room or a suite, mentioning travel with a spouse. By 12:02:30 AM UTC, the assistant presented several options meeting the criteria: a 5-star Radisson hotel in the city center and a 4-star Hyatt hotel near cultural landmarks, both offering the requested amenities and room types. The user expressed a preference for the suite at the Hyatt near cultural landmarks, highlighting the appeal of the amenities. The assistant confirmed the booking of the suite at the Hyatt and assured the user that a confirmation including check-in times and amenity details would be sent shortly. Throughout the conversation, the assistant maintained a helpful and accommodating tone, and the user demonstrated clear decision-making and satisfaction with the options provided. The entire interaction took place within approximately four minutes, starting at 12:00 AM UTC on August 27, 2023.The conversation began on September 24, 2023 (Sunday) at 12:00 AM UTC with the user initiating a request for travel planning assistance. The user expressed a desire to plan a trip to Italy focused on immersing themselves in the country's rich art and history. They specifically requested help in crafting an itinerary that included unique cultural activities such as cooking classes and visits to lesser-known towns. This indicated the user's interest in an authentic and culturally rich travel experience beyond typical tourist destinations. Prior to this, ... \textbf{following 2632 tokens}.\\
        \bottomrule
    \end{tabularx}
\end{table}

\begin{figure*}[!t]
    \centering
    \includegraphics[width=0.95\textwidth]{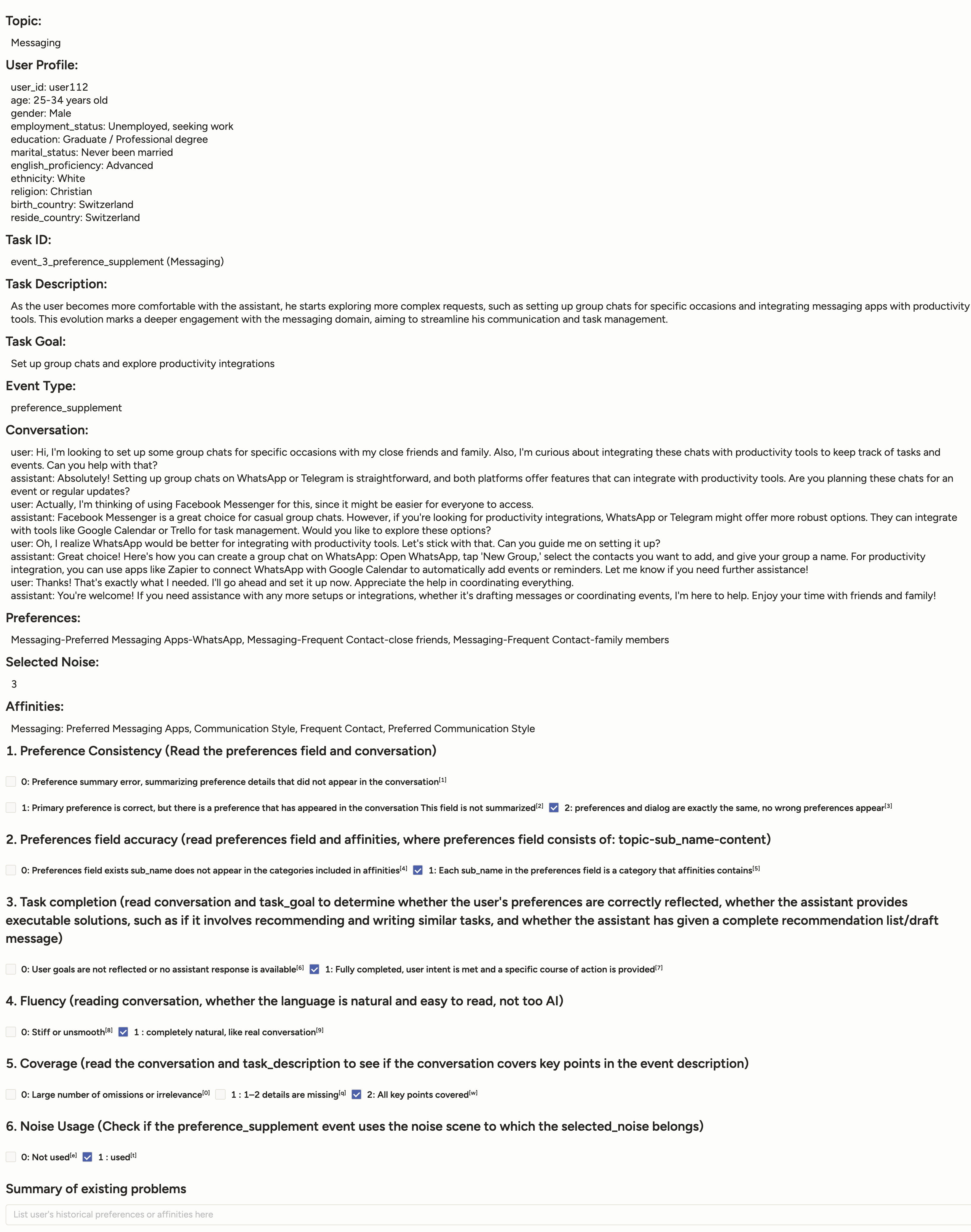}
    \caption{Detailed annotation in Label Studio for data quality assessment. The protocol evaluates 6 criteria, ranging from preference consistency to linguistic naturalness.}
    \label{apx:candidate_response2}
\end{figure*}